%% file: main.tex
\documentclass{article}

\PassOptionsToPackage{hyphens}{url}

\usepackage[main,final]{neurips_2026}

\input{commands}

\input{globe_macros}
\input{preamble}

\title{{\globe}: Accurate Surrogates for Boundary-Driven {PDE}s via Domain-Inspired Architectures and Equivariance}

\author{%
  Peter Sharpe \\
  NVIDIA \\
  \texttt{psharpe@nvidia.com} \\
}

\begin{document}

\maketitle

\begin{abstract}
    We introduce \globe{}, a nonlinear neural surrogate for boundary-driven, homogeneous, weakly nonlinear PDEs that draws inductive bias from boundary-element methods and equivariant ML, representing solutions as superpositions of learnable Green's-function-like kernels evaluated from boundary faces to targets. The architecture is translation-, rotation-, and parity-equivariant; discretization-invariant; and units-invariant via rigorous nondimensionalization. On AirFRANS (steady incompressible RANS over NACA airfoils), a compact ${\sim}\SI{117}{k}$-parameter \globe{} model has lower mean-squared error than published values for the strongest reported ML surrogates on both the interpolation and low-data splits, in single-run comparisons under the released AirFRANS normalization. The model evaluates fields at arbitrary points, handles non-watertight meshes, and generalizes under Reynolds number and angle-of-attack distribution shifts. We further introduce a hierarchical Barnes-Hut acceleration that exploits the kernel's far-field decay to reduce evaluation cost from quadratic to near-linear in the number of boundary elements, with a tunable opening angle that recovers exact dense evaluation in the zero-angle limit. Quantitative validation focuses on 2D AirFRANS; on 3D DrivAerML cars, \globe{} predicts surface and volume fields on the full meshes, and we compare it with a matched Transolver baseline. Together, these results show that boundary-element-inspired inductive biases can substantially improve the accuracy of ML surrogates for boundary-driven PDEs while scaling to large 3D meshes.
\end{abstract}

\section{Introduction}

\subsection{Motivation: The Computational Challenge of CFD and the Need for Surrogates}

Computational fluid dynamics (CFD) is essential to modern engineering, from aerospace and automotive design to energy systems and climate modeling. In aerospace and automotive applications, the engineering quantities of interest (e.g., time-averaged lift, drag, surface pressure) are typically steady-state, so flows are modeled with either the Reynolds-averaged Navier-Stokes (RANS) equations or higher-fidelity (and inherently unsteady) large-eddy simulation (LES) methods, from which a time-averaged solution is extracted. Either way, industrial simulations routinely consume days to weeks of supercomputer time. Several structural features of the underlying PDEs drive this cost. First, the PDE's nonlinearity makes iterative methods most practical,\footnote{Referring to outer-loop iterations for PDE linearization, which wrap any inner-loop iterations that are used to solve the linearized system itself.} yet these introduce the possibility of convergence failures and unfavorable speed-stability tradeoffs. Second, the PDE's elliptic character (discussed in \cref{sec:necessary-attributes}) demands global information propagation and solution relaxation; this directly drives high computational cost, which acceleration techniques such as multigrid or tailored initialization \citep{sharpeAcceleratingTransientCFD2025} only partially mitigate. Third, at the high Reynolds numbers typical of industrial interest, thin boundary layers develop on no-slip walls; these high-gradient regions dominate force predictions and hence must be resolved, yet doing so requires many degrees of freedom. This expense has motivated rapid growth in machine-learning surrogate models for PDE solution operators. However, many existing ML surrogates do not structurally enforce fundamental physical requirements: they may break under coordinate transformations (lacking translation, rotation, or parity equivariance), depend on arbitrary discretization choices, lack unit consistency, or employ local receptive fields that limit generalization for elliptic PDEs. Architecturally enforcing these requirements can improve generalization beyond narrow training distributions, which is critical for industrial deployment.

\subsection{Inspiration from Classical Boundary-Element Methods for PDEs}
\label{sec:ibl}

Advanced numerical methods for aerodynamics provide instructive architectural guidance. Integral boundary layer (IBL) methods such as XFOIL by \citet{drelaXFOILAnalysisDesign1989} achieve 100- to 1,000-fold computational speedups over volumetric RANS by discretizing only boundaries via boundary-element methods (BEM), exploiting the linear, homogeneous structure of potential flow. This boundary-centric formulation with global information transfer through boundary-only integrals encodes powerful inductive biases for fluid dynamics, and indeed more broadly to many PDEs as shown by \citet{martinssonFastDirectSolvers2020}. In this work, we combine such domain-inspired architectural ideas with equivariant ML and the flexibility of learned representations. The long-term goal is to extend the accuracy and speed benefits of IBL methods to complex 3D cases where analytical IBL extensions have long remained elusive. \cref{sec:ibl-detailed} provides additional background on IBL methods and their relevance to the \globe{} architecture introduced in \cref{sec:model}.

\subsection{Design Desiderata for Physics Models}
\label{sec:necessary-attributes}

To motivate the architectural choices in this work, we identify three categories of design desiderata for physics models: (1) \emph{physical conservation laws and symmetries}, namely translation-, rotation-, and parity-equivariance following from Noether's theorem; (2) \emph{representation invariances}, namely discretization-invariance (numerical consistency) and units-invariance (dimensional consistency); and (3) \emph{problem-specific constraints}, such as a receptive field matching PDE information flow (global for elliptic PDEs) and explicit boundary-condition encoding. By architecturally enforcing these exact constraints rather than hoping to learn them from data, we reduce the hypothesis space to physically plausible functions, improving data efficiency and out-of-distribution generalization. A comprehensive discussion with detailed justifications and architectural implications is provided in \cref{sec:necessary-attributes-appendix}.

\subsection{Related Work: ML Surrogate Models for PDEs}

ML surrogate approaches differ in their tradeoffs between unstructured-mesh handling, discretization invariance, and geometric symmetries (see \cref{sec:related-work-appendix} for a structured comparison). CNNs need structured grids; GNNs handle irregular domains but lack discretization-invariance and require many layers for global coupling on elliptic PDEs; neural operators achieve discretization invariance but often still use localized kernels; neural fields \citep{serranoOperatorLearningNeural2023,catalaniScalableSurrogateModels2025} evaluate at arbitrary points but typically compress boundary and problem information into fixed-size global latents. Closer to \globe{} are neural surrogates that use boundary integral or Green's-function structure: BI-GreenNet \citep{linBIGreenNetLearningGreens2023}, BENO \citep{wangBENOBoundaryEmbedded2024}, Neural Green's Functions \citep{yooNeuralGreensFunction2025}, BINO \citep{wuBINOBoundaryIntegral2026}, PIBNet \citep{marsalPIBNetPhysicsInspired2025}, BI-DeepONet \citep{mengBIDeepONet2024}, and Neural FMM \citep{fogniniNeuralFMM2025}. Separately, equivariance-aware PDE surrogates include Lie-symmetry data augmentation \citep{brandstetterLiePointSymmetry2022}, G-FNO \citep{helwigGFNOGroupEquivariant2023}, and the synergistic geometric/physical-constraint work of \citet{huangGeometricPhysicalConstraints2025}. \globe{} differs by simultaneously enforcing Euclidean equivariance (translation, rotation, parity), units-invariance, and discretization-invariance as structural constraints, while learning kernels end-to-end without analytical fundamental solutions or eigendecompositions, and by maintaining geometry-attached per-face latents rather than a single global latent representation. \cref{sec:related-work-appendix} provides a detailed comparison.

\subsection{Contributions and Paper Organization}

This paper's primary contributions are:

\begin{itemize}[noitemsep]
    \item A first-principles review of design desiderata for physics models, and guidance for constructing architectures that respect these (\cref{sec:necessary-attributes}).
    \item A new domain-inspired ML architecture for boundary-driven PDE surrogate modeling that respects these attributes, \textbf{\globe} (\textbf{G}reen's-function-\textbf{L}ike \textbf{O}perator for \textbf{B}oundary \textbf{E}lement PDEs; \cref{sec:model}).
    \item Detailed evaluation on AirFRANS (2D RANS over NACA airfoils): comprehensive accuracy comparisons against state-of-the-art surrogates on the official Full and Scarce splits, plus Reynolds and angle-of-attack extrapolation tasks and a non-watertight-mesh robustness study (\cref{sec:task,sec:results,sec:extended-results}).
    \item A 3D evaluation on DrivAerML car aerodynamics, in which surface and volume \globe{} models train on full-size cars and predict on the full meshes via the dual-tree Barnes-Hut acceleration, with integrated-drag accuracy and a comparison against a matched Transolver baseline (\cref{sec:task-drivaer,sec:drivaer-appendix}).
    \item A dual-tree Barnes-Hut acceleration scheme that exploits the kernel's structurally enforced far-field decay to break the $\mathcal{O}(\nsrc^2)$ scaling barrier while recovering exact dense evaluation, and its equivariance guarantees, as $\theta \to 0$ (\cref{sec:hierarchical-acceleration-appendix}).
    \item Ablation studies of the Pad\'{e}-approximant kernel and the dual-tree Barnes-Hut acceleration on AirFRANS (\cref{sec:ablation-appendix}).
    \item An open-source reference implementation of both the model and case studies, released as part of NVIDIA PhysicsNeMo (\url{https://github.com/NVIDIA/physicsnemo}, under \texttt{examples/cfd/external\_aerodynamics/globe/}).
\end{itemize}

Novel technical contributions include:
\begin{itemize}[noitemsep]
    \item \textbf{Equivariant kernel functions} maintaining exact translation, rotation, and parity symmetries via invariant feature engineering and local basis reprojection.
    \item \textbf{\pade{}-approximant MLPs} whose bounded rational-function structure is motivated by Green's-function behavior.
    \item \textbf{Multiscale kernel composition} encoding multiple characteristic length scales (viscous, geometric, and learned combinations).
    \item \textbf{Communication hyperlayers} that propagate boundary-condition (BC) information between partitions, replacing the traditional BEM linear solve with learned long-range message passing.
    \item \textbf{Rigorous nondimensionalization} throughout the data pipeline, ensuring units-invariance and physical consistency.
\end{itemize}

\section{\globe{} Model Architecture}
\label{sec:model}

We now introduce \textbf{\globe}: \textbf{G}reen's-function-\textbf{L}ike \textbf{O}perator for \textbf{B}oundary \textbf{E}lement PDEs. The architecture synthesizes three key insights: (1) for linear homogeneous PDEs, solutions can be exactly represented as superpositions of Green's functions; for nonlinear PDEs, learnable kernel superpositions can approximate solutions when the PDE operator is weakly nonlinear or exhibits approximately linear behavior in significant domain regions; (2) for boundary-driven problems, discretizing only boundaries (not interiors) dramatically reduces degrees of freedom while maintaining global information flow; and (3) neural networks can learn kernel shapes that generalize Green's function concepts to settings where analytical forms are unknown or do not exist.

While the theoretical framework is general, in this work we develop and evaluate \globe{} specifically for aerodynamic flows, where the vorticity-free regions of subsonic external flows are well-approximated by linear potential flow theory (\cref{sec:ibl-detailed}). \globe{} itself is nonlinear: the linear BEM analogy provides an inductive bias rather than assuming a linear RANS operator, and only the final aggregation over sources is linear. More broadly, the architecture targets PDEs with the following characteristics:

\begin{itemize}[noitemsep]
    \item \textbf{Homogeneous governing equations:} no spatially varying forcing terms or material properties
    \item \textbf{Boundary-driven physics and focus:} quantities of interest are primarily driven by and/or measured at boundaries (e.g., boundary layers, stress concentrations)
    \item \textbf{Elliptic or mixed character:} requiring global information propagation for a solver to be well-posed
    \item \textbf{Weak or regional nonlinearity:} PDEs where the governing operator is linear or exhibits approximately linear behavior in significant portions of the domain
    \item \textbf{Multiscale structure:} PDEs exhibiting diverse phenomena at multiple characteristic length scales
\end{itemize}

This profile encompasses several important problem classes in computational engineering. Our quantitative validation here focuses on 2D external aerodynamics (AirFRANS), together with a 3D evaluation on DrivAerML. While the same architectural principles apply to other boundary-driven elliptic problems (e.g., static structural analysis, heat conduction, and electrostatics), evaluating them in those domains remains future work.

\globe{} maps from boundary meshes (partitioned by boundary condition type), global parameters, and reference length scales to scalar and vector fields at arbitrary query points (see \cref{sec:architecture-diagrams-appendix} for a diagram). While similar to neural-field architectures, \globe{} differs by maintaining per-face latents rather than compressing to a global vector, enabling Euclidean equivariance and avoiding global bottlenecks. More generally, \globe{} has more in common with boundary-element-method solvers than existing ML surrogates, drawing heavy inspiration from aerodynamic panel methods (a classical 2D/3D potential-flow technique that discretizes only the body surface) and integral boundary layer methods \citep{katzLowSpeedAerodynamicsSecond2004,drelaXFOILAnalysisDesign1989,drelaFlightVehicleAerodynamics2013}.

\subsection{Architecture Overview}

\globe{} predicts fields on arbitrary query points in space, given one or more boundary meshes partitioned by BC type. The architecture is organized in three levels, from bottom to top:

\begin{itemize}[leftmargin=*,noitemsep]
    \item At the lowest level, the centroid of each face of a boundary mesh acts as a source point that influences every target point in an all-to-all manner via a learned \textbf{kernel function} $\kernel$. This kernel is shared across all faces. The model's physical symmetries and invariances are enforced within this kernel, described in \cref{sec:kernel-functions}.

    \item At the middle level, \globe{} implements a \textbf{multiscale composition} of multiple kernel functions, whose influences are linearly superimposed. Each individual kernel is still shared across boundary faces, and is associated with a different reference length scale $\ell_i$. These length scales are derived from combinations of dimensional constants within the PDE governing equations, and should be given as model inputs. This multiscale composition is described in \cref{sec:multiscale}.

    \item At the top level, \globe{} implements a stack of \textbf{communication hyperlayers}, which is a sequence of these multiscale kernel compositions. These hyperlayers pass latent scalar and vector fields from all boundaries to all other boundaries, prior to the final evaluation at user-chosen query points. These hyperlayers are also used to compute the \textbf{source strength} values associated with each face, which are used to linearly weight the contributions of each source face to each target point in the subsequent hyperlayer. This is described in \cref{sec:hyperlayers}.
\end{itemize}

\cref{fig:globe_arch} depicts the overall architecture. Boundary meshes are merged by BC type; within each hyperlayer, the kernel is evaluated from every source BC to targets taken from either (a) other boundary face centers (to propagate latent variables) or (b) user query points (final layer). Outputs across source BC types are summed. After the final prediction, a learnable calibration (affine for scalar fields, linear for vector fields\footnote{Required to preserve rotation-equivariance.}) is applied.

\begin{figure}[!tb]
    \centering
    \resizebox{\linewidth}{!}{\input{figures/globe_arch}}%
    \caption{\textbf{GLOBE architecture.} The model operates at three hierarchical levels. \emph{Top:} Communication hyperlayers propagate latent information between boundary partitions before final evaluation. \emph{Middle:} Each hyperlayer contains multiscale kernels operating at different reference length scales, with outputs summed. \emph{Bottom:} Individual kernels evaluate all-to-all source-to-target influences through invariant feature engineering, \pade-approximant networks, far-field decay envelopes, and equivariant vector reprojection, then aggregate over sources weighted by strengths and face areas.}
    \label{fig:globe_arch}
\end{figure}

\paragraph{Mathematical properties.} The architecture satisfies the design desiderata of \cref{sec:necessary-attributes}: translation-, rotation-, and parity-equivariance through relative positions and local reprojection onto carefully constructed vector bases; discretization-invariance via area-weighted boundary integrals; and units-invariance through rigorous nondimensionalization. More detailed discussion of these properties is provided in \cref{sec:mathematical-proofs}. Additionally, \globe{}'s all-to-all boundary-to-target evaluation (\cref{eq:aggregation-over-sources}) provides a true global receptive field: every query point receives information from \emph{every} boundary face, directly fulfilling elliptic PDE requirements. For parabolic and hyperbolic PDEs, global receptive fields remain beneficial (allowing the model to learn long-range couplings if present) without imposing structural limitations. Communication hyperlayers (\cref{sec:hyperlayers}) further enable boundary-to-boundary information exchange, mediating the kind of global boundary coupling that traditional boundary-element methods achieve through dense linear solves.

\subsection{Kernel Functions}
\label{sec:kernel-functions}

A \globe{} model represents PDE solutions as superpositions of learnable \emph{kernel functions} $\kernel$, inspired by Green's functions. Each kernel computes influence values from source boundary faces to target points. Visual illustrations are provided in \cref{sec:kernel-illustrations}; detailed mathematical formulations are in \cref{sec:kernel-appendix}. The kernel evaluation proceeds through the following steps:

\begin{enumerate}[leftmargin=*,noitemsep]
    \item \textbf{Inputs.} Gather scalars $\vect{\xi}$ and vectors $\mat{V}$ for each source-target pair, including the nondimensionalized relative position $\vect{r}_{ts} = (\vect{x}_t - \vect{x}_s)/\ell$ and source normal $\vect{\hat{n}}_s$. Using only relative positions ensures translation-equivariance.
    \item \textbf{Invariant feature engineering} (\cref{sec:kernel-engineered-features-appendix}). Transform vectors into rotationally invariant scalars using $\smoothlog(\norm{\vect{v}_i})$ and magnitude-weighted Legendre polynomial features on pairwise angles, preserving $C^\infty$ continuity. Here, $\smoothlog(x) = (1 - e^{-x}) \cdot \ln(1 + x)$ is a smooth nonlinear compression that preserves $C^\infty$ continuity at zero.
    
    \item \textbf{\pade{}-approximant MLP processing} (\cref{sec:pade-approx-mlp-appendix}). Process invariant features through a rational-function-structured network of the form $\operatorname{sgn}(\phi_n)\,|\phi_n|^N / (1+|\phi_d|^D)$, where $\phi_n,\phi_d$ are independent MLPs (\cref{eq:pade-approx-def}). This rational form is motivated by Green's-function structure, and the denominator ensures bounded Lipschitz constants; the algebraic far-field decay itself is imposed by the envelope of the next step, not by the \pade{} form.
    
    \item \textbf{Far-field decay constraint} (\cref{sec:far-field-decay-appendix}). Scale outputs by a Lamb-Oseen-inspired envelope (\cref{eq:far-field-decay-full}) that drives self-influence to zero at $\vect{r}=\vect{0}$ and enforces $\sim 1/r^{\nsp-1}$ decay at large distances, with $\nsp\in\{2,3\}$. This enables stable spatial extrapolation and justifies the far-field cluster approximation used in \cref{sec:hierarchical-acceleration-appendix}.
    
    \item \textbf{Vector reprojection} (\cref{sec:vector-reprojection-appendix}). Reconstruct physical vector outputs by projecting latent channels onto local basis vectors derived from input directions, restoring rotational equivariance.
    
    \item \textbf{Aggregation over sources.} Sum contributions from all $\nsrc$ source faces, weighted by source strengths $w_s$ and face areas $a_s$:
    \begin{equation}
        \label{eq:aggregation-over-sources}
        \kernel_t \,=\, \sum_{s=1}^{\nsrc} w_s\, a_s\, \kernel_{ts}.
    \end{equation}
    Area weighting ensures discretization-invariance as the boundary mesh is refined (converging to a boundary integral).
\end{enumerate}

\subsection{Multiscale Kernel Functions}
\label{sec:multiscale}

Many physical systems exhibit multiple characteristic lengths. In \globe{}, each user-provided reference length $\ell_i$ defines a parallel \emph{kernel branch} $\kernel^{(i)}$ that shares the kernel architecture of \cref{sec:kernel-functions} and its inputs but has independent learnable parameters and an \emph{effective} reference length $\ell^\text{eff}_i = \ell_i\,\exp(\alpha_i)$, where $\alpha_i$ is a learned scalar. Because the relative position feature is normalized by $\ell^\text{eff}_i$, each branch effectively ``zooms'' the same geometric interaction to a different scale. The branches' outputs are linearly superposed (\cref{eq:multiscale-sum-appendix}), extending \cref{eq:aggregation-over-sources} to a multiscale kernel that respects discretization-invariance and units-invariance.
To expose scale relationships to the network, every pair $(\ell_i,\ell_j)$ contributes a unitless feature $\ln(\ell_i/\ell_j)$ appended to the global scalar input. (Encoding individual length values directly would break units-invariance.) This introduces an inductive bias toward scale similarity: the network behaves equivariantly under absolute scale changes so long as all nondimensional parameters (e.g., Reynolds number, Mach number) are fixed.\footnote{Strictly, this scale-equivariance requires that all relevant length scales (i.e., those identified by Buckingham $\pi$ analysis of the governing-equation parameters) are encoded as reference lengths.} Detailed equations and guidance on choosing $\{\ell_i\}$ for a given PDE are deferred to \cref{sec:multiscale-appendix}.

\subsection{Communication Hyperlayers}
\label{sec:hyperlayers}

The top level stacks $H$ \emph{communication hyperlayers} that exchange latent information between boundary-condition (BC) partitions in an all-to-all manner before the final interior evaluation. Each hyperlayer contains one multiscale kernel per source BC type (e.g., one for all ``no slip'' faces, another for all ``free slip'' faces), enabling \globe{} to learn BC-type-specific behavior, such as boundary-layer presence or absence. The hyperlayers also produce the per-branch source strengths $w^{(k)}_s$ that weight \cref{eq:aggregation-over-sources}.

For each of $H$ hyperlayers, the multiscale kernel of every source BC is evaluated at all boundary face centers (grouped by target BC), passing latent scalars and vectors plus updated source strengths to the next hyperlayer. The final hyperlayer evaluates at the user-specified query points to produce the physical output fields, followed by a learned per-field calibration (an affine transformation for scalars and a scalar multiplier for vectors, preserving rotational equivariance). The first hyperlayer initializes all source strengths to one and the per-face latent channels to empty sets. Per-hyperlayer pseudocode and notation are deferred to \cref{sec:hyperlayer-rationale}.

\paragraph{Inductive bias.} The hyperlayer architecture is inspired by the linear solve $\mat{A}\vect{w}=\vect{b}$ in traditional BEM. We replace this solve with learned message passing because (a) direct solves are $\mathcal{O}(\nsrc^{3})$ and iterative solves explode autodiff memory; (b) BC-residual evaluation requires symbolic PDE knowledge; and (c) hyperlayers enable direct boundary-to-boundary coupling -- e.g., letting freestream information propagate directly to a body boundary without requiring high-magnitude long-range kernels. \cref{sec:hyperlayer-rationale} provides additional rationale.

\paragraph{Hierarchical acceleration.} The dense $\mathcal{O}(\nsrc^2 + \nsrc \ntrg)$ kernel evaluation is accelerated by a dual-tree Barnes-Hut scheme that exploits the kernel's guaranteed far-field decay (\cref{eq:far-field-decay-full}). The same kernel evaluates both individual source-target pairs and cluster-centroid pairs, dropping asymptotic cost to $\mathcal{O}(n \log n)$ in $n_{\max}=\max(\nsrc, \ntrg)$ while recovering the dense result as $\theta \to 0$. Its axis-aligned cluster bounding boxes break exact rotational equivariance at finite $\theta$, by a small amount controlled by $\theta$. The full algorithm, including the opening criterion, area-weighted monopole aggregates, and the target-side expansion option, is described in \cref{sec:hierarchical-acceleration-appendix}.

\section{Experiments}
\label{sec:task}

\subsection{AirFRANS (2D RANS over NACA airfoils)}
\label{sec:task-airfrans}

We evaluate \globe{} on AirFRANS \citep{bonnetAirfRANSHighFidelity2023}, comprising 1,000 steady incompressible RANS solutions over NACA airfoils with Reynolds numbers in $[2,6]\times10^6$ and angles of attack from $-5^\circ$ to $15^\circ$. We adopt the four official evaluation regimes: \textbf{Full} (800 train / 200 test, i.i.d.), \textbf{Scarce} (200 train / 200 test), \textbf{Reynolds extrapolation} (train on $\Rey\in[3,5]\times10^6$), and \textbf{Angle-of-attack (AoA) extrapolation} (train on $\alpha\in[-2.5^\circ,12.5^\circ]$).

Inputs to \globe{} consist of the airfoil boundary mesh (labeled \texttt{no\_slip}), the freestream direction $\bm{U}_\infty/|\bm{U}_\infty|$, and reference lengths (chord $c_\text{ref}$ and viscous scale $\delta_{FS}$, defined in \cref{sec:preprocessing-details}). The model predicts nondimensional velocity, pressure coefficient, total pressure coefficient, turbulent viscosity ratio, and wall-shear stress at arbitrary query points. All fields are rigorously nondimensionalized; detailed preprocessing equations and methodological choices (including exclusion of signed distance fields and treatment of redundant output channels) are provided in \cref{sec:preprocessing-details}. Training details and hyperparameters are in \cref{sec:hyperparameters}.

\subsection{DrivAerML: 3D Car Aerodynamics}
\label{sec:task-drivaer}

\begin{figure}[t]
    \centering
    \includegraphics[width=\textwidth]{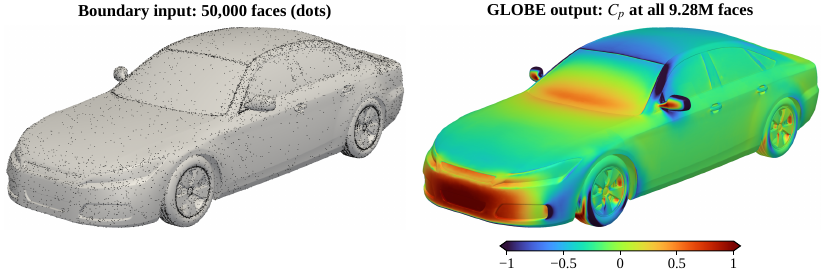}\\[2pt]
    \includegraphics[width=\textwidth]{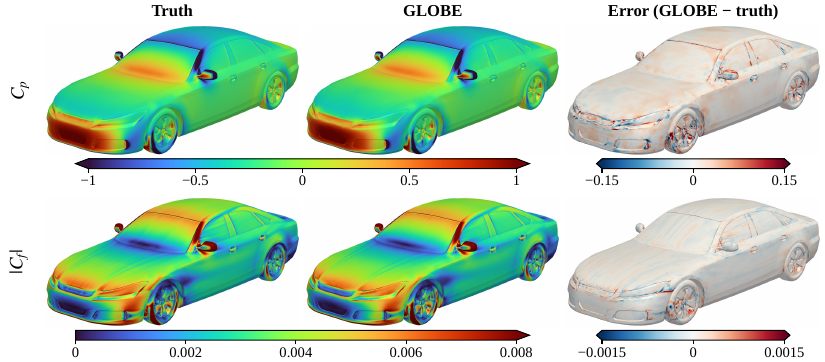}
    \caption{\globe{} surface model on the DrivAerML validation car with the median surface-pressure error (relative $L_2$ error 8.7\% for pressure and 9.4\% for wall-shear stress). Top: the boundary input is a sample of \num{50000} faces of the car surface (dots), plus the floor patches (not drawn); \globe{} predicts at all \SI{9.28}{M} surface faces. Bottom: pressure coefficient $C_p$ and wall-shear-stress coefficient magnitude $|C_f|$ from CFD (truth), \globe{}, and their difference. Truth and \globe{} share one color range per field. The largest errors are at sharp features: the wheels, the grille and front-bumper edges, and the mirrors. \Cref{fig:drivaer-views} adds a rear view.}
    \label{fig:drivaer}
\end{figure}

We also apply \globe{} to DrivAerML \citep{ashtonDrivAerMLHighFidelityComputational2024}, a dataset of 3D CFD solutions over parametrically varied DrivAer cars in a wind tunnel, with ${\sim}\SI{8.8}{M}$ surface cells and ${\sim}\SI{150}{M}$ volume cells per car. We train on 400 cars and evaluate on 34 held-out validation cars. A surface model predicts pressure and wall-shear stress at the surface cells, and a volume model predicts velocity and pressure at the volume cells; both take the car and the wind-tunnel floor as boundaries and train on samples of \num{50000} prediction points per step with the dual-tree Barnes-Hut acceleration of \cref{sec:hierarchical-acceleration-appendix}. Single models reach mean relative $L_2$ errors of 8.9\% for surface pressure, 9.7\% for wall-shear stress, 8.3\% for velocity, and 6.7\% for volume pressure, with $R^2 = 0.976$ for the drag coefficient integrated from the predicted surface fields (\cref{fig:drivaer}). A Transolver \citep{wuTransolverFastTransformer2024} trained and evaluated the same way reaches 4.1\%, 6.2\%, 5.3\%, and 3.8\%, with $R^2 = 0.987$ (\cref{tab:drivaer}). These are single-seed validation results with checkpoints selected on the same cars; \cref{sec:drivaer-appendix} adds three-seed ensembles and inference costs.

\section{Results and Discussion}
\label{sec:results}

For each of the four AirFRANS tasks in \cref{sec:task-airfrans}, we train a unique \globe{} model, with full training details and hyperparameter values in \cref{sec:hyperparameters}. The following sections discuss the accuracy of these models, both quantitatively and qualitatively.

\subsection{Error Metrics}
\label{sec:results-quantitative}

\begin{table}[!tb]
    \centering
    \newcommand{\blank}{--}
    \caption{AirFRANS \textbf{Full} (interpolation) task: mean squared error on the official test cases, normalized by the training-set field variance as in the released AirFRANS evaluator (\cref{sec:airfrans-metric-definitions}); lower is better. External values are as reported in the cited works (not rerun). \globe{} values are single-run point estimates. Metrics with (\blank) were not reported in the associated citation.}
    \label{tab:airfrans-full}
    \resizebox{\linewidth}{!}{%
    \begin{tabular}{l c c c c}
        \toprule
                                                                              & \multicolumn{4}{c}{\textbf{Mean Squared Error, ``Full'' task}}                                                                                           \\
        \cmidrule(lr){2-5}
        {Model}                                                               & {$\bar{u}_x$}                                                  & {$\bar{u}_y$}               & {$\bar{p}$}                 & {$\bar{p}_s$}               \\
        \midrule
        \textbf{\globe{} (our work)}                                          & {\bfseries 0.0042} \enegtwo                                    & {\bfseries 0.0021} \enegtwo & {\bfseries 0.0023} \enegtwo & {\bfseries 0.0070} \enegtwo \\
        \midrule
        enf2enf \citep{catalaniGeometryAwareInference2025}                     & \blank                                                         & \blank                      & 0.1100 \enegtwo             & 0.3200 \enegtwo             \\
        CORAL \citep{serranoOperatorLearningNeural2023,catalaniGeometryAwareInference2025} & \blank                                              & \blank                      & 0.3500 \enegtwo             & 1.2000 \enegtwo             \\
        Transolver \citep{wuTransolverFastTransformer2024}                     & \blank                                                         & \blank                      & 0.3700 \enegtwo             & 1.4200 \enegtwo             \\
        MeshGraphNet \citep{pfaffLearningMeshBasedSimulation2021,catalaniGeometryAwareInference2025} & \blank                                    & \blank                      & 2.1400 \enegtwo             & 3.8700 \enegtwo             \\
        GINO \citep{liGeometryInformedNeuralOperator2023,catalaniGeometryAwareInference2025} & \blank                                          & \blank                      & 2.9700 \enegtwo             & 4.8200 \enegtwo             \\
        MLP \citep{bonnetAirfRANSHighFidelity2023}                             & 0.9490 \enegtwo                                                & 0.9780 \enegtwo             & 0.7370 \enegtwo             & 11.3000 \enegtwo            \\
        GraphSAGE \citep{bonnetAirfRANSHighFidelity2023}                       & 0.8320 \enegtwo                                                & 0.9940 \enegtwo             & 0.6610 \enegtwo             & 6.6200 \enegtwo             \\
        PointNet \citep{bonnetAirfRANSHighFidelity2023}                        & 3.5000 \enegtwo                                                & 3.6450 \enegtwo             & 1.1510 \enegtwo             & 9.2500 \enegtwo             \\
        GUNet \citep{bonnetAirfRANSHighFidelity2023}                           & 1.5170 \enegtwo                                                & 2.0280 \enegtwo             & 0.6570 \enegtwo             & 3.8600 \enegtwo             \\
        \bottomrule
    \end{tabular}%
    }
\end{table}

\begin{table}[!tb]
    \centering
    \newcommand{\blank}{--}
    \caption{AirFRANS \textbf{Scarce} (low-data) task: mean squared error on the official test cases, normalized by the training-set field variance as in \cref{tab:airfrans-full}; lower is better. External values are as reported in the cited works (not rerun). \globe{} values are single-run point estimates.}
    \label{tab:airfrans-scarce}
    \resizebox{\linewidth}{!}{%
    \begin{tabular}{l c c c c}
        \toprule
                                                                              & \multicolumn{4}{c}{\textbf{Mean Squared Error, ``Scarce'' task}}                                                                                           \\
        \cmidrule(lr){2-5}
        {Model}                                                               & {$\bar{u}_x$}                                                    & {$\bar{u}_y$}               & {$\bar{p}$}                 & {$\bar{p}_s$}               \\
        \midrule
        \textbf{\globe{} (our work)}                                          & {\bfseries 0.0238} \enegtwo                                      & {\bfseries 0.0148} \enegtwo & {\bfseries 0.0496} \enegtwo & {\bfseries 0.1889} \enegtwo \\
        \midrule
        MARIO \citep{catalaniScalableSurrogateModels2025}                      & 0.1520 \enegtwo                                                  & 0.1130 \enegtwo             & 0.2400 \enegtwo             & 2.7000 \enegtwo             \\
        Transolver \citep{wuTransolverFastTransformer2024,catalaniScalableSurrogateModels2025} & 2.1050 \enegtwo                                & 2.1080 \enegtwo             & 4.4340 \enegtwo             & 23.6700 \enegtwo            \\
        MLP \citep{bonnetAirfRANSHighFidelity2023}                             & 1.6470 \enegtwo                                                  & 1.4510 \enegtwo             & 3.9040 \enegtwo             & 21.9200 \enegtwo            \\
        GraphSAGE \citep{bonnetAirfRANSHighFidelity2023}                       & 1.4570 \enegtwo                                                  & 1.4540 \enegtwo             & 4.6960 \enegtwo             & 19.4500 \enegtwo            \\
        PointNet \citep{bonnetAirfRANSHighFidelity2023}                        & 3.1110 \enegtwo                                                  & 2.7760 \enegtwo             & 3.2940 \enegtwo             & 18.2700 \enegtwo            \\
        GUNet \citep{bonnetAirfRANSHighFidelity2023}                           & 1.7490 \enegtwo                                                  & 1.8250 \enegtwo             & 3.3880 \enegtwo             & 14.7300 \enegtwo            \\
        \bottomrule
    \end{tabular}%
    }
\end{table}

\paragraph{Metric definitions.} All AirFRANS tables use the official test cases, all original volume-mesh nodes as query points, the same surface masks, pointwise MSE within each case, and equal averaging across cases, as in the released AirFRANS evaluator. The \globe{} rows divide each field's MSE by one point-pooled training-set variance per field, as the released evaluator does. Earlier versions of this work reported case-normalized scores, which divide each case's MSE by that case's truth-field variance; \cref{sec:airfrans-metric-definitions} defines both metrics and lists those scores. The denominator does not affect training or checkpoint selection, which use a separate Huber objective. External values are as reported in the cited works: these are comparisons against published values, not controlled reruns with matched code, seeds, or training budgets. All AirFRANS \globe{} results are single-run point estimates, so we draw no conclusions from small differences. Physically nondimensionalized metrics (using $|\bm{U}_\infty|$ for velocity, $\tfrac{1}{2}\rho|\bm{U}_\infty|^2$ for pressure) are reported in \cref{sec:physical-metrics}.

\paragraph{Comparison with modern surrogates.} As seen in \cref{tab:airfrans-full,tab:airfrans-scarce}, \globe{} has the lowest error of the listed models on every reported field. On the \textit{Full} split, the strongest reported model is enf2enf by \citet{catalaniGeometryAwareInference2025}, for which only pressure metrics are available; its reported volume- and surface-pressure errors are roughly $48\times$ and $46\times$ those of \globe{}. On the \textit{Scarce} split, where the latest state-of-the-art models have been benchmarked by \citet{catalaniScalableSurrogateModels2025}, the reported errors of the next-best model, MARIO, are roughly $6.4\times$, $7.7\times$, $4.8\times$, and $14.3\times$ those of \globe{} on $\bar{u}_x$, $\bar{u}_y$, $\bar{p}$, and $\bar{p}_s$, and those of Transolver \citep{wuTransolverFastTransformer2024} are roughly $90\times$--$140\times$. The original AirFRANS baselines \citep{bonnetAirfRANSHighFidelity2023} remain in the tables as historical context (on \textit{Full}, their best per-field errors are roughly $200\times$--$550\times$ those of \globe{}); because comparisons against weak baselines can overstate progress \citep{mcgreivyWeakBaselinesReporting2024}, we emphasize the modern ones.

On the extrapolation splits, the only published comparison set is the AirFRANS dataset baselines \citep{bonnetAirfRANSHighFidelity2023}. Relative to the best of these in each field, \globe{}'s errors are lower by roughly $126\times$--$380\times$ on Reynolds-number extrapolation and $11\times$--$46\times$ on angle-of-attack extrapolation, including roughly $31\times$ and $46\times$ on AoA volume and surface pressure (\cref{sec:extrapolation-results}). The higher absolute errors in AoA extrapolation likely reflect that angle-of-attack changes induce global shifts in flow topology (stagnation point movement, wake trajectory), causing more severe physics changes than the boundary layer thickening associated with Reynolds number variation.

Qualitative visualizations demonstrating sharp boundary-layer gradients, coherent wake structures, and robust extrapolation are provided in \cref{sec:qualitative-results}. Additionally, \globe{} demonstrates robustness to severely decimated boundary meshes (50\% face removal with destroyed topological connectivity), with errors about $10\times$ those of the intact-mesh Full model and of the same order as the Scarce model's (\cref{sec:non-watertight}). This tests robustness to missing faces, not convergence under mesh refinement, which we did not study.

\paragraph{Ablations and execution cost.} Single-component ablations on the AirFRANS \emph{Scarce} split (\cref{sec:ablation-appendix}) show that, among the substitutions we tested, replacing the \pade{}-approximant kernel with a same-depth-and-width MLP produces the largest degradation, inflating total validation loss by roughly $90\times$ in a single run. Because the \pade{} variant has ${\sim}2\times$ as many parameters, this does not separate rational parameterization from capacity, conditioning, or optimization; the boundary-only representation, hyperlayers, multiscale branches, and equivariance were not individually ablated. Ablations also characterize the accuracy of the dual-tree Barnes-Hut scheme of \cref{sec:hierarchical-acceleration-appendix} as a function of the opening angle $\theta$ and the far-field target expansion option. The Barnes-Hut scheme drops asymptotic cost from $\mathcal{O}(\nsrc^2 + \nsrc \ntrg)$ to $\mathcal{O}(n \log n)$ in $n_{\max} = \max(\nsrc, \ntrg)$, recovers the dense result as $\theta \to 0$, and enables training on full-size DrivAerML cars (\cref{sec:task-drivaer}).

\section{Conclusions}
\label{sec:conclusions}

By combining boundary-element structure with exact Euclidean equivariance, \globe{} achieves strong accuracy on 2D AirFRANS with a compact \mbox{${\sim}\SI{117}{k}$-parameter} model. In single-run comparisons against published results, \globe{} attains lower error than the strongest reported surrogates on both the interpolation and low-data splits, as well as the published baselines on the Reynolds-number and angle-of-attack extrapolation splits. To scale beyond dense $\mathcal{O}(\nsrc^2)$ evaluation, a dual-tree Barnes-Hut scheme exploits the kernel's far-field decay to reduce cost to $\mathcal{O}(n \log n)$, recovering exact dense evaluation as $\theta \to 0$ and enabling end-to-end training on large 3D DrivAerML meshes (\cref{sec:task-drivaer}). Ablation studies (\cref{sec:ablation-appendix}) show that replacing the Pad\'{e}-approximant kernel with a standard MLP of the same depth and width sharply degrades validation loss (though without isolating the cause from parameter count or conditioning) and quantify the accuracy--speed tradeoff of the Barnes-Hut approximation; a mesh-decimation study (\cref{sec:non-watertight}) further shows tolerance to missing boundary faces and broken connectivity.

\paragraph{Limitations and future work.} \globe{} is nonlinear, but its inductive bias is drawn from linear boundary-element structure; we expect it to be least suitable for flows with strong separation, large topology changes, or regimes outside the training distribution. All AirFRANS results are single-run point estimates, and on AirFRANS we did not evaluate integrated lift, drag, or pitching-moment coefficients, so pointwise MSE alone does not establish accuracy for aerodynamic design. On DrivAerML, \globe{}'s field errors are 1.5--2.2$\times$ those of a matched Transolver, and its volume inference is about $6\times$ slower. Separately, even in mostly attached flows, large shifts in global flow topology (stagnation-point movement, wake trajectory) move the test cases far from the training distribution, leading to higher errors on AoA extrapolation than on Reynolds-number extrapolation. The far-field decay envelope assumes algebraic asymptotic behavior, missing logarithmic, exponential, or strongly anisotropic regimes. The current Barnes-Hut configuration uses a zeroth-order source-side monopole approximation; higher-order multipole expansions and a CUDA-implemented tree traversal are natural follow-ups. Closing the gap on 3D cars and transferability to other PDE domains (structural analysis, heat transfer, electrostatics, volumetric-source-term PDEs) are reserved for future work; see \cref{sec:limitations-appendix}.

\section*{Broader Impacts}

This paper aims to advance machine learning for scientific computing. Once validated on integrated design quantities, more accurate surrogate models could help accelerate engineering design in aerospace, automotive, and energy and reduce CFD's computational energy footprint. Aerodynamic surrogates are also inherently dual-use; our intended use is open civil and scientific applications.

\begin{ack}
We thank Corey Adams for valuable feedback and paper review, and the NVIDIA CSRG team for maintaining supercomputing resources that enabled this research. This work was funded by NVIDIA, the author's employer.
\end{ack}

\bibliographystyle{plainnat}
\bibliography{references}

\newpage
\appendix

\section{Design Desiderata for Physics Models}
\label{sec:necessary-attributes-appendix}

This appendix provides detailed discussion of the design desiderata for physics models, summarized in the main text and in \cref{tab:necessary-attributes}.

\begin{table}[htb]
    \centering
    \caption{Design desiderata for physics models and their architectural implications.}
    \label{tab:necessary-attributes}
    \small
    \renewcommand{\arraystretch}{1.2}
    \renewcommand{\tabularxcolumn}[1]{>{\linespread{0.9}\selectfont}m{#1}}
    \begin{tabularx}{\textwidth}{>{\raggedright\arraybackslash}m{0.30\textwidth}>{\raggedright\arraybackslash}m{0.26\textwidth}>{\raggedright\arraybackslash}X}
        \toprule
        \textbf{Attribute}                & \textbf{Physical Basis}          & \textbf{Architectural Implication}                                                        \\
        \midrule
        \multicolumn{3}{l}{\textit{Physical Conservation Laws \& Symmetries}}                                                                                            \\
        \quad Translation-equivariance    & Conservation of momentum         & Use only relative spatial information; no absolute coordinates                            \\
        \quad Rotation-equivariance       & Conservation of angular momentum & Vectors via local basis reprojection, not channel concatenation                           \\
        \quad Parity-equivariance         & Euclidean symmetry group         & Proper handling of vectors vs. pseudovectors                                              \\
        \midrule
        \multicolumn{3}{l}{\textit{Representation Invariances}}                                                                                                          \\
        \quad Discretization-invariance   & Numerical consistency            & Predictions must converge as mesh is refined                                              \\
        \quad Units-invariance            & Physical consistency             & Operate in unitless space via nondimensionalization                                       \\
        \midrule
        \multicolumn{3}{l}{\textit{Problem-Specific Constraints (examples for PDEs)}}                                                                                    \\
        \quad Receptive field structure   & PDE well-posedness               & Global receptive field for elliptic PDEs; match characteristic speeds for hyperbolic PDEs \\
        \quad Boundary condition encoding & PDE well-posedness               & Explicit BC type labels as inputs                                                         \\
        \bottomrule
    \end{tabularx}
\end{table}

\subsection{Physical Conservation Laws and Symmetries}

These requirements follow directly from fundamental physics via Noether's theorem, which connects conservation laws to continuous symmetries.

\paragraph{Translation-equivariance.}
By Noether's theorem, conservation of momentum follows from translational symmetry; any proposed physics model must therefore exhibit translational symmetry: if all inputs are translated in space, outputs should be identically translated. Architecturally, this requires using only relative spatial information -- any use of absolute coordinates in the core learned encoding of a model breaks this symmetry.

\paragraph{Rotation-equivariance.}
Conservation of angular momentum requires rotational symmetry. While rare in ML surrogates for PDEs, rotation-equivariant architectures have achieved remarkable successes in other domains \citep{thomasTensorFieldNetworks2018,fuchsSE3Transformers3DRotoTranslation2020,jumperHighlyAccurateProtein2021,batznerE3equivariantGraphNeural2022}, with libraries such as \texttt{e3nn} \citep{geigerE3nnEuclideanNeural2022} enabling implementation. Common sources of violations include treating vector components as concatenated scalar channels (which implicitly fixes an arbitrary global Cartesian frame) or using traditional convolutional layers on structured grids. If a trained model does not produce identical output when the grid is rotated, it lacks rotational equivariance.

\paragraph{Parity-equivariance (reflection / chirality).}
Almost all physical laws are equivariant under parity.\footnote{With the exception of the weak interaction at subatomic scales \citep{wuExperimentalTestParity1957}.} Intuitively, a pair of vortices spinning clockwise and counterclockwise must have \emph{exactly} identical decay rates. Combining parity symmetry with translation- and rotation-equivariance forms the Euclidean symmetry group. Structurally embedding Euclidean symmetry substantially improves data efficiency and generalizability \citep{geigerE3nnEuclideanNeural2022,batznerE3equivariantGraphNeural2022,zhangThreedimensionalIntegralBoundary2022,sharpeNeuralFoilAirfoilAerodynamics2025,cuEquivariance}. While equivariant ML research has historically focused on materials science, chemistry, and biology, these symmetries are equally relevant in computer-aided engineering (CAE), where their incorporation into PDE surrogates remains rare.

\paragraph{On the infeasibility of data augmentation for equivariance learning.}
An alternative to equivariant architecture design is data augmentation, which avoids specialized architectural constructions. However, augmentation yields only approximate symmetry and is computationally prohibitive on large 3D datasets. For example, the DrivAerML dataset \citep{ashtonDrivAerMLHighFidelityComputational2024} comprises 31\,TB of 3D CFD data before augmentation, and learning 3D rotational symmetry via augmentation requires roughly a $500\times$ increase in dataset size and training compute \citep{smidtEuclideanSymmetryEquivariance2020}. Compounding this across continuous translations and rotations over $SE(3)$ is impractical at scale.

\subsection{Representation Invariances}

In addition to respecting physical symmetries, any valid physics model must produce results that are independent of arbitrary choices in how we represent and discretize the problem.

\paragraph{Discretization-invariance (numerical consistency).}
For a valid solver, discretization error must approach zero as mesh resolution increases -- this is \emph{numerical consistency} in CFD. Architecturally, this requires that the model's predictions converge to a well-defined limit independent of the specific mesh used. CNNs and GNNs tie their representations to specific discretizations; neural operators and neural fields achieve this property by parameterizing continuous functions.

\paragraph{Units-invariance.}
Another fundamental requirement is invariance to the choice of physical units. The same \emph{physical} scenario, whether represented in SI or Imperial units, should yield the same \emph{physical} solution. This necessitates dimensional consistency throughout the entire modeling pipeline. Notably, most ML surrogates violate this principle during data preprocessing, before the model architecture is even considered -- for example, by concatenating features with incompatible units (such as combining velocities, $[L][T]^{-1}$, with lengths, $[L]$). A practical way to achieve units-invariance is to learn in unitless spaces via rigorous nondimensionalization, as in \citet{sharpeNeuralFoilAirfoilAerodynamics2025}.

\subsection{Problem-Specific Constraints}

These requirements vary depending on the PDE type and problem formulation, but are equally critical for learning valid solution operators.

\paragraph{PDE information flow and receptive fields.}

Beyond the symmetries discussed above, any valid physics model must respect the information flow structure dictated by the PDE itself. Different PDE types\footnote{Here, we restrict this classification discussion to second-order-and-higher PDEs, as these form the vast majority of computationally expensive PDEs used in engineering applications.} impose fundamentally different requirements on how information propagates through space:

\textit{Elliptic PDEs} (e.g., steady-state aerodynamics, electrostatics, static structural analysis) exhibit global coupling: every point in the domain is meaningfully influenced by every boundary condition and source term. Mathematically, well-posedness requires that boundary data at \emph{any} location can affect the solution at \emph{any} other location. For a surrogate model, this necessitates a \emph{global receptive field}: the model's output at a query point depends on information from the entire domain.

\textit{Parabolic PDEs} (e.g., heat diffusion, most unsteady problems) exhibit infinitely fast spatial information propagation, but with rapid decay of influence with distance. For small time steps $\Delta t$, local receptive fields can provide reasonable approximations; however, large-timestep solution operators still require spatially distant information access.

\textit{Hyperbolic PDEs} (e.g., wave propagation, inviscid compressible flow) have finite characteristic speeds. Information travels along characteristic curves at well-defined wave speeds. A valid solver must resolve these characteristics: the receptive field at time $t$ must encompass all points reachable via characteristics originating from initial/boundary data. This requirement is in some ways an analogue of the Courant-Friedrichs-Lewy (CFL) condition in explicit time-stepping: if the numerical receptive field fails to capture physical wave propagation, the solver becomes fundamentally unstable or inaccurate.

Architectures with local or limited-range receptive fields cannot satisfy well-posedness requirements for PDEs requiring global coupling. Vanilla GNNs, for instance, require a number of message-passing layers proportional to the mesh graph diameter to propagate information globally, which does not scale to fine meshes. Many neural operator implementations similarly employ localized kernels (ball queries, k-NN) that lack global coupling.

When such architectures nonetheless achieve strong benchmark performance on elliptic PDEs despite these architectural limitations, it suggests they may be learning correlations with problem-specific features (mesh patterns, geometric heuristics) rather than the underlying PDE physics. For example, a local GNN applied to aerodynamics might learn that fine mesh cells correlate with boundary layer regions and hence lower velocities, or that distance to a boundary (measured by neighbor-hops) correlates with pressure gradients. Such patterns can produce good metrics on validation sets from the same distribution but fail catastrophically under genuine distribution shift to different geometries, flow regimes, or mesh strategies. For applications requiring robust out-of-distribution generalization, architecturally ensuring that receptive fields match the PDE's intrinsic information flow provides stronger guarantees than relying on training data to implicitly capture these patterns.

\paragraph{Explicit boundary condition encoding.}
A related and equally fundamental issue is that many PDE problem encodings in the ML literature fail to encode any boundary condition (BC) type information whatsoever, rendering the learning task mathematically ill-posed: it is not possible to truly learn a PDE's solution operator if the boundary conditions are not provided. Consider an automotive aerodynamics case with mixed BCs: no-slip (vehicle body) and free-slip (ground plane). Without explicit BC labels, models either a) cannot distinguish boundary condition differences, leading to inaccurate solutions, or b) worse, they infer boundary condition types from nearby geometry, learning spurious correlations (e.g., ``flat surfaces tend to be road surfaces, which tend to be slip walls'') that also fail catastrophically under distribution shift. Proper generalization requires BC types as explicit inputs, ensuring physically grounded learning rather than dataset artifacts.

\subsection{Architecture Design Philosophy for ML Surrogates for PDEs}

The prevalence of architectures in the field of ML for PDEs that violate these requirements reflects a mismatch between the domains where modern ML has historically achieved its greatest successes and the requirements of modeling physics-governed systems. While breakthroughs in perception, language, and game-playing have naturally influenced and advanced PDE surrogates, these domains differ fundamentally from physics in their treatment of structure and generalization. For example, symmetries can be useful heuristics in computer vision,\footnote{For example, see the data augmentation techniques in \citet{krizhevskyImageNetClassificationDeep2017}.} but the inevitable exceptions and edge cases limit the usefulness of structurally enforcing these symmetries.

By contrast, in physics, conservation laws and coordinate invariances are exact, universal mathematical constraints. Hence, by encoding these symmetries, we reduce the hypothesis space to physically plausible functions, improving data efficiency and out-of-distribution accuracy. In CAE applications, this out-of-distribution generalization is particularly crucial, due to adverse selection bias: engineers reach for new simulations precisely when encountering scenarios that are far from any previously analyzed case. ML benchmarks with i.i.d. train/test splits may therefore inadvertently reward architectures that achieve high in-distribution accuracy through mechanisms that preclude practical generalization (encoding absolute position, architecture-specific discretizations, local receptive fields), as discussed by \citet{mcgreivyWeakBaselinesReporting2024}.

As the field matures toward foundation-model aspirations, we believe that incorporating physical structure at the architectural level is essential. Foundation models must generalize across vast ranges of geometries, flow regimes, and boundary conditions. Models that violate conservation laws, break under coordinate transformations, or depend on arbitrary discretization choices are fundamentally learning correlations in the training distribution rather than underlying physics. If a model violates basic physical requirements known to be exact, it has likely learned dataset-specific patterns rather than a general solution operator, regardless of its performance on held-out test sets from the same distribution.

\section{Related Work Comparison}
\label{sec:related-work-appendix}

\cref{tab:related-work} provides a structured comparison of ML surrogate approaches for PDEs.

\begin{table}[htb]
    \centering
    \caption{ML surrogate approaches for PDEs and their characteristics.}
    \label{tab:related-work}
    \small
    \renewcommand{\arraystretch}{1.2}
    \begin{tabularx}{\textwidth}{>{\raggedright\arraybackslash}m{0.20\textwidth}>{\raggedright\arraybackslash}X>{\raggedright\arraybackslash}X}
        \toprule
        \textbf{Method}  & \textbf{Key Strengths}                                                                                                                    & \textbf{Key Limitations}                                                                                                                                                     \\
        \midrule
        CNNs / UNets     & Hierarchical features; efficient on structured grids                                                                                      & Structured grids only; lack geometric symmetries; costly interpolation for unstructured meshes                                                                               \\
        GNNs             & Handle unstructured meshes; effective for parabolic PDEs                                                                                  & Local receptive fields require $\mathcal{O}(\text{diameter})$ layers for elliptic PDEs; typically lack rotation equivariance                                                 \\
        Neural Operators & Discretization invariance; function-space mappings; well-suited for heterogeneous PDEs that are chiefly driven by field forcing functions & Often use localized kernels to incorporate geometry information, limiting information flow; limited ability to incorporate boundary information for boundary-driven problems \\
        Neural Fields    & Arbitrary query points; continuous representations; discretization invariance                                                             & Global compression bottleneck; limited geometric equivariance                                                                                                                \\
        \bottomrule
    \end{tabularx}
\end{table}

\subsection{BEM-Inspired Neural Surrogates}

A growing body of work integrates boundary integral or Green's function structure into neural architectures. We summarize the most relevant approaches and clarify how \globe{} differs.

\textbf{BI-GreenNet} \citep{linBIGreenNetLearningGreens2023} learns Green's functions for Poisson and Helmholtz equations by combining known fundamental solutions with boundary integral networks. The requirement for explicit fundamental solutions to remove singularities limits applicability to PDEs where such solutions are analytically available. \globe{} learns kernel functions entirely end-to-end without requiring analytical Green's functions or fundamental solutions.

\textbf{BENO} \citep{wangBENOBoundaryEmbedded2024} addresses elliptic PDEs with complex geometries and inhomogeneous boundary values using dual GNN branches for interior sources and boundary values, coupled with a Transformer encoder over the boundary nodes. While inspired by Green's functions, BENO relies on graph connectivity and attention mechanisms rather than kernel superposition. \globe{} avoids graph structure entirely, operating on unordered face clouds with all-to-all kernel evaluation.

\textbf{Neural Green's Functions} \citep{yooNeuralGreensFunction2025} learns solution operators for linear PDEs by decomposing the Green's function via eigendecompositions of the differential operator. This provides strong generalization across source and boundary functions, as well as across unseen geometries within a fixed shape category (one model is trained per category in their experiments), but requires the PDE operator to admit such decomposition. \globe{} imposes no such requirement.

\textbf{Neural FMM} \citep{fogniniNeuralFMM2025} integrates the hierarchical computation flow of the fast multipole method into neural networks for learning Green's operators of elliptic PDEs. \globe{} similarly equips a neural kernel with a hierarchical acceleration scheme (dual-tree Barnes-Hut; \cref{sec:hierarchical-acceleration-appendix}), but the two approaches differ in several respects: Neural FMM embeds the multi-level FMM hierarchy into the network parameterization itself, whereas \globe{} leaves the kernel parameterization unchanged and applies the spatial tree decomposition to the kernel evaluation; \globe{}'s acceleration is enabled by a structurally enforced far-field decay envelope (\cref{eq:far-field-decay-full}) rather than a trained-in hierarchical structure; and \globe{} simultaneously enforces Euclidean equivariance, discretization-invariance, and units-invariance, which Neural FMM does not target.

\textbf{BI-DeepONet / BI-TDONet} \citep{mengBIDeepONet2024} formulate operator learning based on boundary integral equations, enabling trained models to predict solutions across different 2D domains without retraining; like BI-GreenNet, the framework requires the analytical fundamental solution of the PDE class to construct the BIE kernel. \textbf{BINO} \citep{wuBINOBoundaryIntegral2026} learns boundary-integral Green's traction kernels for mesh deformation, demonstrating generalization across boundary conditions. \textbf{PIBNet} \citep{marsalPIBNetPhysicsInspired2025} combines a physics-inspired graph-construction strategy (favoring short-range edges motivated by Green's-function decay) with a multiscale graph neural network to approximate BEM boundary solution traces for 3D multiple scattering problems.

\paragraph{Key distinctions of \globe{}.} While these works share the insight that boundary integral formulations provide useful inductive bias, \globe{} differs in several important respects: (1) it enforces exact Euclidean equivariance (translation, rotation, and parity) throughout the architecture, which no prior BEM-neural hybrid achieves; (2) it enforces units-invariance and discretization-invariance as additional structural constraints; (3) it uses communication hyperlayers for boundary-to-boundary coupling rather than traditional BEM linear solves or global latent compression; and (4) it operates on unordered boundary face clouds without requiring graph connectivity, mesh topology, or analytical fundamental solutions.

\subsection{Physics-Constrained Neural PDE Surrogates}

Incorporating physical priors into neural PDE solvers is an active research direction. \citet{brandstetterLiePointSymmetry2022} exploit Lie point symmetries of PDEs for data augmentation, reporting sample-efficiency gains of roughly $\times 4$ to $\times 16$ across the 1D evolution equations they study; this is complementary to \globe{}'s approach of structural enforcement, which provides exact (rather than approximate) symmetry preservation. \citet{wangIncorporatingSymmetryDeep2021} similarly demonstrate that incorporating equivariances (rotation, uniform motion, and scaling) into deep dynamics models yields large improvements in generalization on Rayleigh-B\'enard convection and ocean-current data; their work focuses on convolutional dynamics models (ResNet and U-Net variants) on regular grids, with each model enforcing a single symmetry at a time, while \globe{} integrates multiple simultaneous equivariances with a boundary-element formulation operating on unstructured surface meshes. G-FNO \citep{helwigGFNOGroupEquivariant2023} extends Fourier neural operators with group equivariance to rotations, translations, and reflections in the frequency domain, though it operates on structured grids. Most recently, \citet{huangGeometricPhysicalConstraints2025} systematically demonstrate that combining geometric symmetries with physical constraints (e.g., divergence-free conditions) yields synergistic improvements on fluid dynamics problems. \globe{} shares the philosophy that structural enforcement of physical constraints is preferable to learning them from data, while extending this principle to boundary-driven formulations with a broader set of simultaneously enforced invariances (Euclidean equivariance, discretization-invariance, and units-invariance).

\section{Kernel Function Formulation}
\label{sec:kernel-appendix}

This appendix provides detailed mathematical formulations for the kernel function components summarized in \cref{sec:kernel-functions}.

The kernel function operates on source-target pairs, with all positional information entering through the relative position vector from the source face center to the target point, nondimensionalized by a reference length scale $\ell$ to preserve units-invariance:
\begin{equation}
    \label{eq:relative-position}
    \vect{r}_{ts} \,=\, \frac{\vect{x}_t - \vect{x}_s}{\ell} \in \RR^{\nsp}.
\end{equation}

\subsection{Invariant Engineered Features}
\label{sec:kernel-engineered-features-appendix}

The scalar collection $\vect{\xi}$ can be used directly as model inputs. However, the vector collection $\mat{V}$ cannot be used directly without breaking rotational equivariance. We therefore transform these vectors into rotationally invariant features. (Rotational equivariance is later restored by vector reprojection in \cref{sec:vector-reprojection-appendix}.)

One possible rotationally invariant feature is the magnitude $\norm{\vect{v}_i}$. However, this has two issues: (a) very large magnitudes can dominate the scalar collection, saturating downstream steps, and (b) the magnitude function breaks $C^1$ continuity near the zero vector, violating continuous differentiability.\footnote{While not strictly required for forward evaluation, smooth derivatives are valuable for downstream tasks such as gradient-based aerodynamic shape optimization.}

Instead, we encode the value $\smoothlog(\norm{\vect{v}_i})$ for each vector $\vect{v}_i$ as an additional scalar in the collection $\vect{\xi}$. Here, we define the function $\smoothlog : \RR^1 \to \RR^1$ as
\begin{equation}
    \smoothlog(x) = (1 - e^{-x}) \cdot \ln(1 + x),
\end{equation}
which is crafted to have several desirable properties on the domain $x \geq 0$ (as is the case when applied to the raw magnitude values):
\begin{itemize}[noitemsep]
    \item $\smoothlog(x) \sim x^2$ as $x \to 0$
    \item $\smoothlog(x) \sim \ln(x)$ as $x \to \infty$
    \item $\smoothlog(x)$ is monotonically increasing for $x > 0$, and hence forms a bijection with $x$
    \item $\smoothlog(x)$ is $C^\infty$ continuous for all $x \geq 0$, with higher-degree derivatives that grow relatively slowly with degree
    \item $\smoothlog(x)$ can be efficiently and accurately computed using the \texttt{expm1} and \texttt{log1p} special functions
\end{itemize}

\begin{figure}[htb]
    \centering
    \includegraphics[width=\textwidth]{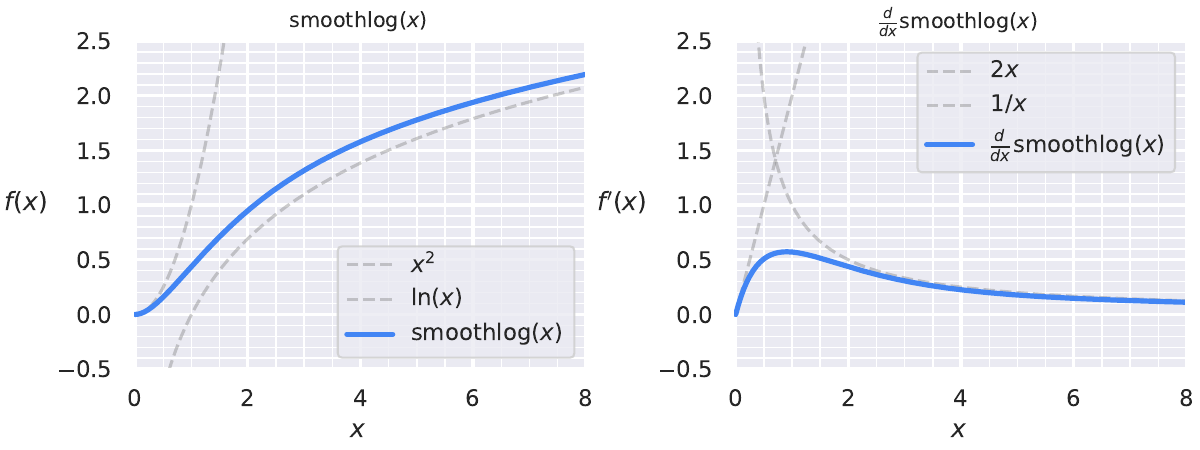}
    \caption{The $\smoothlog(x)$ function and its derivative, with near- and far-field asymptotic limits shown as dashed lines.}
    \label{fig:smoothlog}
\end{figure}

\noindent \cref{fig:smoothlog} illustrates these asymptotic properties.

Next, we compute additional engineered features by considering every possible unordered pair of vectors $\vect{v}_a$ and $\vect{v}_b$ in the collection $\mat{V}$. For each such pair, we evaluate the magnitude-weighted spherical harmonic features
\begin{equation}
    \label{eq:spherical-harmonics}
    \smoothlog\!\left(\norm{\vect{v}_a} \cdot \norm{\vect{v}_b}\right) \cdot \Leg_i\!\left(\cos(\theta_{ab})\right),
\end{equation}
where $\Leg_i$ is the $i$-th Legendre polynomial (for $i = 1, 2, \ldots, n_{\text{harmonics}}$, a hyperparameter), and $\theta_{ab}$ is the angle between vectors $\vect{v}_a$ and $\vect{v}_b$. These are added to the collection $\vect{\xi}$. Multiplication by $\smoothlog\!\left(\norm{\vect{v}_a} \cdot \norm{\vect{v}_b}\right)$ is necessary to preserve $C^\infty$ continuity: without this scaling, $\theta_{ab}$ can change discontinuously as either vector crosses zero. Effectively, \cref{eq:spherical-harmonics} forms a Legendre-polynomial expansion on the pairwise angle (the axisymmetric/$m{=}0$ component of a spherical-harmonic expansion in 3D), weighted by a nonlinear function of the magnitudes.

At this point, all relevant information has been gathered and encoded into a collection of scalars $\vect{\xi}$. As scalar-valued quantities, these are inherently rotationally invariant (in addition to translation-invariant), facilitating further processing.

\subsection{\pade{}-Approximant MLPs}
\label{sec:pade-approx-mlp-appendix}

At this stage, the scalar collection $\vect{\xi}$ could be processed by a standard multilayer perceptron (MLP). However, we find empirically that using a custom \pade{}-approximant MLP instead yields substantial improvements in model accuracy and training speed.

We introduce what we call a ``\pade{}-approximant MLP'' as a learnable mapping of the form:

\begin{equation}
    \label{eq:pade-approx-def}
    \pade\left(\vect{\xi}\right) \;=\; \frac{\operatorname{sgn}\!\left(\phi_n\left(\vect{\xi}\right)\right)\,\left\lvert\phi_n\left(\vect{\xi}\right)\right\rvert^{N}}{1 + \left\lvert\phi_d\left(\vect{\xi}\right)\right\rvert^{D}}.
\end{equation}
Here, $\phi_n$ and $\phi_d$ are independent learnable MLPs with $C^\infty$ differentiable activations (SiLU in the reference implementation); $N$ and $D$ are the integer-valued orders of the numerator and denominator, respectively. The numerator uses sign-preserving exponentiation, while the denominator uses magnitude-only exponentiation; when combined with the constant term, this ensures that the denominator is always $\ge 1$, preventing arbitrarily large Lipschitz constants.

We typically set $N = 2$ and $D = 2$; while other choices can be made, the property $N = D$ is quite useful as it ensures that the \pade{}-approximant layer asymptotes to a constant value in the far-field limit, for any direction in the input space $\vect{\xi}$. This will later aid in careful control of the asymptotic behavior of the kernel function, as discussed in \cref{sec:far-field-decay-appendix}.

The motivation stems from Green's function structure: these typically exhibit rational-function-like behavior with algebraic singularities near sources and algebraic decay at infinity (e.g., $1/r$ for 3D Poisson). While standard MLPs are universal approximators, \pade{}-approximant MLPs provide explicit inductive bias toward this functional form. The algebraic far-field decay of the kernel, however, comes from the explicitly imposed envelope of \cref{sec:far-field-decay-appendix}, not from the \pade{} parameterization, and our ablation (\cref{sec:ablation-pade}) does not isolate which property of the \pade{} form helps. \cref{sec:pade-illustrations} provides visualizations demonstrating the expressivity of randomly initialized \pade{}-approximant MLPs compared to standard MLPs.

\subsection{Far-Field Decay Constraint}
\label{sec:far-field-decay-appendix}

To bias the model toward physically plausible decay with distance, we then scale all the raw outputs of the \pade{}-approximant MLP by a function of the relative position vector $\vect{r}_{ts}$:
\begin{equation}
    \label{eq:far-field-decay-full}
    g\big(\norm{\vect{r}_{ts}}\big)
    \,=\, \frac{1 - e^{-\norm{\vect{r}_{ts}}^2}}{\left(\norm{\vect{r}_{ts}}^2 + 1\right)^{(\nsp-1)/2}},
\end{equation}
where $\nsp$ is the number of spatial dimensions (2 or 3), yielding $\sim 1/r$ decay in 2D and $\sim 1/r^2$ in 3D as $\norm{\vect{r}_{ts}}\to\infty$ in any direction.

This introduces two useful effects. First, self-influence at $\vect{r}_{ts}=\vect{0}$ is driven to zero, encouraging kernels to share influence across faces and preventing overly compact support (the ``bed-of-nails interpolant'' failure mode in radial basis interpolation). The chosen decay is inspired by the vorticity profile of the Lamb-Oseen vortex. Second, the far-field decay encodes an inductive bias toward the inverse square law: a common large-distance behavior of particle-mediated interactions (or alternatively, Gauss's law for fields with conserved fluxes) that is observed across a vast array of physical systems, while still allowing the model to learn deviations where appropriate (e.g., advection-dominated regimes).

This also introduces favorable generalization properties in spatial extrapolation. When a \globe{} model is queried with target points that are much more distant from PDE boundaries than any target points seen in the training data, the kernel function cannot extrapolate wildly: ``in the absence of other information, physical phenomena tend to zero at infinity.'' Additionally, this guaranteed far-field decay enables the formulation of infinite far-field boundary conditions, where boundary effects naturally vanish at large distances without requiring explicit far-field boundary meshes.

\subsection{Vector Reprojection for Rotational Equivariance}
\label{sec:vector-reprojection-appendix}

The \pade{}-approximant MLP outputs a latent vector $\vect{z}$ that is \emph{not} a physical-space vector; its dimensionality typically far exceeds $\nsp$. Interpreting any subset of channels as a physical vector would implicitly select an arbitrary global Cartesian frame, breaking rotational symmetry. Therefore, while individual channels of $\vect{z}$ can be directly interpreted as physical \emph{scalar} fields, constructing physical \emph{vector} outputs requires a separate procedure.

To illustrate how we can achieve this, consider the simplified case of constructing a single physical vector $\vect{u}$ associated with the pairwise interaction between a source face and a target point. We fundamentally need to project channel(s) of $\vect{z}$ onto a local basis vector set with non-arbitrary directionality to preserve rotational symmetry. A convenient source of non-arbitrary directionality can be found in the set of vectors $\mat{V}$ that were gathered as inputs to the kernel function. We then assemble a set of basis vectors\footnote{Strictly speaking, this is not a true basis in the linear algebra sense, as (a) we typically have more basis vectors than spatial dimensions, and (b) the collection may not always span all of $\RR^{\nsp}$ (e.g., if every vector in $\mat{V}$ is the zero vector). This is not problematic for our approach, though the resulting reprojection is generally not invertible.} $\mat{B} = [\vect{b}_1, \vect{b}_2, \ldots, \vect{b}_n]$ by including the following components:

\begin{enumerate}[leftmargin=*]
    \item Base direction:
          \begin{itemize}
              \item The radial unit vector $\vect{\hat{r}}$ from the source face to the target point.
          \end{itemize}
    \item For each non-$\vect{r}$ input vector $\vect{v}_i$ with unit direction $\vect{\hat{v}}_i$:
          \begin{enumerate}
              \item Add the axis direction $\vect{\hat{v}}_i$.
              \item In the 2D case ($\nsp=2$): include a dipole-like direction $\vect{e}_\kappa$ (projection of $-\vect{\hat{v}}_i$ onto the line orthogonal to $\vect{\hat{r}}$).
              \item In the 3D case ($\nsp=3$): include a polar/meridional direction $\vect{e}_\theta$ (projection of $-\vect{\hat{v}}_i$ onto the plane orthogonal to $\vect{\hat{r}}$).
              \item Finally, scale each of the above basis vectors (steps within this enumeration) by $\smoothlog(\norm{\vect{v}_i})$. Similar to the use of $\smoothlog(\cdot)$ in \cref{sec:kernel-engineered-features-appendix}, this is subtly required to preserve $C^\infty$ continuity of the resulting field, and has the side benefit of mitigating saturation effects.
          \end{enumerate}
\end{enumerate}

\noindent\textit{Optional tangential/vortex basis vectors.} The above construction preserves parity symmetry. However, for physical phenomena that are strongly linked to vorticity, additional tangential basis vectors can provide useful inductive bias: in 2D, a tangential direction $\vect{e}_\theta$ orthogonal to $\vect{\hat{r}}$; in 3D, an azimuthal direction $\vect{e}_\phi$ orthogonal to both $\vect{\hat{r}}$ and $\vect{e}_\theta$. These directions are pseudovectors rather than true vectors, so including them naïvely breaks parity symmetry. A parity-preserving implementation is possible but requires explicit tracking of vector versus pseudovector semantics throughout the architecture, which is not yet implemented in the reference code. Consequently, these tangential/vortex-like directions are disabled by default, though users may enable them for applications where the inductive bias outweighs the loss of exact parity equivariance. They are disabled in every reported AirFRANS and DrivAerML configuration, and we have not measured the accuracy change from enabling them.

\subsection{Multiscale Kernel Composition}
\label{sec:multiscale-appendix}

This appendix collects the formal definitions for the multiscale kernel composition summarized in \cref{sec:multiscale}. For each user-provided reference length $\ell_i$, a separate kernel branch $\kernel^{(i)}$ is instantiated with effective reference length
\begin{equation}
    \ell^\text{eff}_i \;=\; \ell_i\,\exp(\alpha_i),
\end{equation}
\begin{eqexpl}
    \item{$\ell_i$} user-provided reference length for branch $i$
    \item{$\alpha_i$} learned scalar that fine-tunes the scale during training
\end{eqexpl}
\noindent The relative position fed into branch $i$ is normalized by $\ell^\text{eff}_i$ (compare \cref{eq:relative-position}), so each branch ``zooms'' the same geometric interaction to a different scale. Letting $\kernel^{(k)}_{ts}$ denote the output of branch $k$ for source-target pair $(t,s)$, the multiscale field is
\begin{equation}
    \label{eq:multiscale-sum-appendix}
    \vect{u}_t\;=\; \sum_{k}\;\sum_{s=1}^{\nsrc} w^{(k)}_{s}\, a_s\, \kernel^{(k)}_{ts},
\end{equation}
\begin{eqexpl}
    \item{$\vect{u}_t$} final field value (scalar or vector) at target point $t$
    \item{$k$} kernel branch index, one per reference length scale
    \item{$w^{(k)}_{s}$} per-source, per-branch strength produced by the communication hyperlayers (\cref{sec:hyperlayers})
    \item{$a_s$} area of source face $s$
\end{eqexpl}
\noindent which extends \cref{eq:aggregation-over-sources} from a single kernel to a superposition across multiple kernel branches. In the reference implementation, evaluation supports optional chunking over target points to control memory\footnote{At training time, chunking has no effect on memory due to the need to store intermediate primals; memory reduction there is better achieved by downsampled training. At inference time, chunking can drive peak memory to extremely low levels (700\,MB at chunk size 128 query points) due to the independence of pairwise interactions.} without changing results.

\paragraph{Choosing reference lengths.} The set $\{\ell_i\}$ should be guided by the governing physics. For maximum generalizability, these should be fundamental length scales from the governing equations. In external aerodynamics, candidates include:
\begin{itemize}[noitemsep]
    \item The viscous length $\nu / U_\infty$
    \item A characteristic length such as chord $c_\text{ref}$
    \item Boundary layer thickness estimates (e.g., $\delta \approx 5c/\sqrt{\Rey}$ for laminar)
    \item For rarefied flows, mean free path $\lambda$
\end{itemize}
In practice, 2--4 scales suffice; the learnable $\alpha_i$ adapt initial choices during training.

\subsection{Hierarchical Acceleration via Dual-Tree Barnes-Hut}
\label{sec:hierarchical-acceleration-appendix}

The dense all-to-all kernel evaluation incurs $\mathcal{O}(\nsrc^2 + \nsrc \ntrg)$ complexity (analogous to transformer self-attention). \globe{} mitigates this with a dual-tree Barnes-Hut scheme that exploits the kernel's guaranteed far-field decay (\cref{sec:far-field-decay-appendix}). The same neural kernel evaluates both individual and cluster-centroid interactions. However, the cluster tree uses axis-aligned bounding boxes, which break exact rotational equivariance (by a small error that depends on the opening angle $\theta$); exact equivariance of the dense construction is recovered as $\theta \to 0$. Tuning the opening angle $\theta$ smoothly interpolates between $\mathcal{O}(n \log n)$ approximate evaluation at moderate $\theta$ (typically $\theta \in [0.5, 1.5]$) and exact evaluation at $\theta = 0$.

\paragraph{Monopole approximation.} For a target $t$ and a cluster $C_S$ of source faces, the exact kernel sum
\begin{equation*}
    \sum_{s \in C_S} w_s\, a_s\, \kernel(\vect{x}_t, \vect{x}_s, \mathrm{data}_s)
\end{equation*}
is approximated by a single kernel evaluation at the cluster centroid using area-weighted aggregates of the per-source data:
\begin{equation}
    \label{eq:bh-monopole}
    \sum_{s \in C_S} w_s\, a_s\, \kernel(\vect{x}_t, \vect{x}_s, \mathrm{data}_s) \;\approx\; W_{C_S}\, A_{C_S}\, \kernel\!\left(\vect{x}_t,\, \bar{\vect{x}}_{C_S},\, \overline{\mathrm{data}}_{C_S}\right),
\end{equation}
where $\bar{\vect{x}}_{C_S}$ is the area-weighted source centroid, $\overline{\mathrm{data}}_{C_S}$ is the area-weighted average of source features (normals, latents), $A_{C_S} = \sum_{s \in C_S} a_s$ is the total cluster area, and $W_{C_S} = \sum_{s \in C_S} w_s$ is the summed strength. Area-weighting (rather than strength-weighting) is used for two reasons. First, \emph{correctness}: unlike the mass weights in classical N-body Barnes-Hut, per-source strengths $w_s \in \mathbb{R}$ are signed, so a strength-weighted centroid $\sum_s w_s a_s \vect{x}_s / \sum_s w_s a_s$ is undefined (and ill-conditioned nearby) for any cluster whose strengths nearly cancel -- even though such clusters can carry meaningful far-field content (e.g., the dipole field of two equal-and-opposite sources is nonzero despite zero net strength). Areas $a_s > 0$ are positive by construction, so the area-weighted centroid is unconditionally well-defined, and clusters with $W_{C_S} \approx 0$ correctly contribute (near-)zero to the monopole, with the missing dipole-and-higher content recovered as the cluster is opened during traversal. Second, \emph{reusability}: areas are fixed geometric properties, so the same aggregate is reusable across all kernel branches in a multiscale composition (\cref{sec:multiscale-appendix}) and across communication hyperlayers in which strengths change while geometry does not. This is a zeroth-order monopole approximation in the source coordinate, with the area-weighted geometric centroid and the summed strength factor handled separately rather than fused into a single strength-weighted centroid; classical higher-order multipole expansions in the source coordinate (FMM-style) are a natural extension and are left as future work.

\paragraph{Spatial data structure.} Source and target points are independently organized into binary radix trees built from Morton-code-sorted positions \citep{karrasMaximizingParallelismConstruction2012}, with axis-aligned bounding boxes (AABBs) propagated bottom-up. Each leaf holds a single point, and each internal node stores the area-weighted aggregates of its subtree. Tree construction is $\mathcal{O}(N \log N)$ time, $\mathcal{O}(N)$ memory, and is performed once per forward pass; the resulting tree depends only on geometry and is shared across all multiscale branches and hyperlayers.

\paragraph{Dual-tree traversal and the opening criterion.} Starting from the root pair, the traversal classifies pairs of (target node, source node) at each level using an AABB-distance acceptance criterion that accounts for the spatial extent of \emph{both} nodes:
\begin{equation}
    \label{eq:bh-opening}
    (D_T + D_S)^2 \;<\; \theta^2\, r^2,
\end{equation}
where $D_T, D_S$ are the AABB diagonals of the target and source nodes and $r$ is the minimum distance between the two AABBs (zero if they overlap). The squared-form is used to avoid division when $r = 0$; in that overlap case the criterion always fails, so overlapping nodes are never declared far-field. The criterion is equivalent to the more familiar $(D_T + D_S)/r < \theta$ when $r > 0$. When the criterion holds, the pair is recorded as far-field and evaluated via \cref{eq:bh-monopole}. When it fails and at least one node is internal, the larger node is split and the children are re-tested. When it fails and both nodes are leaves, the underlying points are expanded into the Cartesian product of individual (target, source) pairs that are evaluated exactly. The combined-diameter form (rather than the single-tree $D/r$ test of classical Barnes-Hut) is appropriate because the broadcast approximation also assumes the kernel is roughly constant across the target node, which degrades when the target node is large relative to inter-node distance.

\paragraph{Order of expansion: \texttt{expand\_far\_targets}.} The default zeroth-order configuration evaluates each far-field node pair once at $(\bar{\vect{x}}_{C_T}, \bar{\vect{x}}_{C_S})$ and broadcasts the result to all targets in $C_T$ via a scatter-add. This introduces a target-side approximation that can produce visible blocky artifacts when target clusters are large. An optional flag \texttt{expand\_far\_targets=True} keeps the source-side aggregates but evaluates the kernel at every individual target $\vect{x}_t \in C_T$ paired with the same source aggregate -- eliminating the target-side approximation at the cost of more kernel evaluations per far-field pair. This is \emph{not} a higher-order multipole expansion in the FMM sense (no dipole/quadrupole terms are introduced); the source side is still treated by zeroth-order monopole approximation. Empirically, \texttt{expand\_far\_targets=True} is the more accurate default at large $\nsrc$.

\paragraph{Acceptance criterion semantics.} The parameter $\theta$ follows the standard Barnes-Hut convention: $\theta = 0$ forces all interactions to be exact, larger $\theta$ admits more approximations. Typical operating values are $\theta \in [0.5, 1.5]$ for production use, with $\theta = 1.0$ as the default. As $\theta \to 0$, the dual-tree output converges to the exact dense result, while the kernel's far-field decay envelope bounds the approximation error from distant clusters.

\paragraph{Cross-BC plans and reuse.} For models with multiple boundary-condition partitions, a separate dual-interaction plan is computed for every (source BC, destination BC) pair -- $B^2$ plans for $B$ BC types. Communication hyperlayers reuse all plans across all $H$ layers because the geometry (and hence the trees) does not change between hyperlayers; only the source aggregates are recomputed when latents update. Final-prediction plans (boundary-to-volume) are built once at evaluation time. With $B$ small (typically $1$--$4$ in practice), the additional traversal cost is negligible.

\paragraph{Complexity.} \cref{tab:bh-complexity} summarizes the asymptotic costs.

\begin{table}[htb]
    \centering
    \caption{Asymptotic complexity of the dual-tree Barnes-Hut acceleration applied to \globe{}'s kernel evaluation, with comparison to the dense baseline. Here $N = \max(\nsrc, \ntrg)$ and $n_{\text{far}}$ is the number of far-field node pairs (typically $\mathcal{O}(N)$ for well-separated geometries). The dense baseline memory assumes unchunked evaluation; with target-side chunking it drops to $\mathcal{O}(\text{chunk})$ per call (matching the Barnes-Hut near-field row), at constant time cost.}
    \label{tab:bh-complexity}
    \small
    \begin{tabular}{l c c}
        \toprule
                                         & \textbf{Time}              & \textbf{Memory}           \\
        \midrule
        \textbf{Dense baseline}          & $\mathcal{O}(\nsrc \ntrg)$ & $\mathcal{O}(\nsrc \ntrg)$ \\
        \midrule
        Tree construction                & $\mathcal{O}(N \log N)$    & $\mathcal{O}(N)$           \\
        Aggregate computation            & $\mathcal{O}(N)$           & $\mathcal{O}(N)$           \\
        Dual-tree traversal              & $\mathcal{O}(N \log N)$    & $\mathcal{O}(N \log N)$    \\
        Near-field evaluation            & $\mathcal{O}(N \log N)$    & $\mathcal{O}(\text{chunk})$ \\
        Far-field evaluation             & $\mathcal{O}(n_{\text{far}})$ & $\mathcal{O}(n_{\text{far}})$ \\
        Far-field broadcast (zeroth-order) & $\mathcal{O}(N)$         & $\mathcal{O}(N)$           \\
        \midrule
        \textbf{Barnes-Hut total}        & $\mathcal{O}(N \log N)$    & $\mathcal{O}(N \log N)$    \\
        \bottomrule
    \end{tabular}
\end{table}

\begin{figure}[htb]
    \centering
    \includegraphics[width=0.8\textwidth]{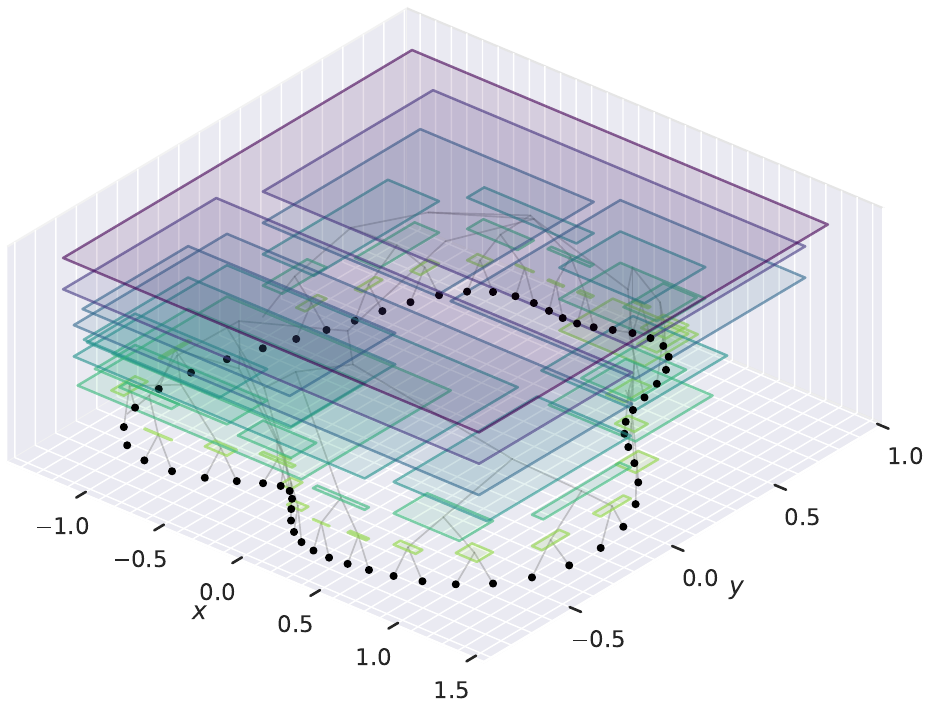}
    \caption{Hierarchical AABB decomposition produced by \globe{}'s cluster tree (\cref{sec:hierarchical-acceleration-appendix}), on a synthetic 2D point cloud. Each AABB is the bounding box of its subtree, drawn at a height proportional to its tree depth (root at the top, leaves at the bottom); color encodes depth.}
    \label{fig:cluster-tree-viz}
\end{figure}

\paragraph{Tree visualization.} \cref{fig:cluster-tree-viz} illustrates the spatial decomposition produced by \globe{}'s cluster tree on a representative point set, with all depths drawn simultaneously to expose the recursive AABB nesting and parent-child structure that drive the dual-tree traversal.

\paragraph{Caveats and future directions.} (i) The dual-tree traversal is currently implemented in Python; an accelerated kernel for the traversal stage would close most of the remaining gap to the kernel-evaluation cost. (ii) Memory efficiency relies on gradient checkpointing in the kernel evaluation; we omit further engineering details that are not central to the algorithm. (iii) Higher-order source-side multipole expansions (true FMM) and an analogous higher-order target-side expansion remain natural follow-ups. (iv) Decimation-robust training (\cref{sec:non-watertight}) is complementary to hierarchical acceleration: training on decimated meshes and evaluating with Barnes-Hut on full meshes provides a practical workflow for large 3D geometries.

\subsection{Hyperlayer Design Rationale}
\label{sec:hyperlayer-rationale}

The architecture of the communication hyperlayers is inspired by the linear solve step of traditional boundary-element methods. There, the strengths $\vect{w}$ associated with each Green's function kernel are obtained by solving a linear system of the form $\mat{A}\vect{w}=\vect{b}$, where $\mat{A}$ corresponds to influence coefficients between Green's functions and is typically square and dense; $\vect{b}$ are the boundary condition residuals in the $\vect{w}=\vect{0}$ case. While future work could explore using this linear solve step, we choose to use this hyperlayer architecture instead for several reasons:
\begin{itemize}[noitemsep]
    \item A linear solve would dominate computational cost as the number of boundary faces grows, as in traditional boundary-element methods: a direct solve requires $\mathcal{O}(\nsrc^{3})$ time, while unrolling an iterative solve through automatic differentiation requires storing every intermediate iterate in memory. Implicit differentiation via a custom vector--Jacobian product avoids storing iterates, but adds implementation complexity and yields approximate gradients when the forward solve is truncated.
    \item This linear solve step requires symbolic knowledge of the governing PDEs (to evaluate boundary condition residuals). This is not necessarily a problem, though it does increase implementation complexity, especially since the need for symbolics has so far been avoided in this architecture.
    \item Hyperlayers propagate boundary-condition values between boundaries. Consider an external aerodynamics problem posed with an explicit far-field ``freestream'' boundary mesh at a large finite distance from the body (rather than an implicit infinite far field as in \cref{sec:far-field-decay-appendix}), where the freestream velocity is prescribed as a Dirichlet condition on the outer boundary. Without boundary-to-boundary communication, the near-field flow around the body would have to be driven directly by kernels evaluated from the distant outer boundary. With communication hyperlayers, the freestream boundary instead passes its state to the body surface, which then governs the near-field flow locally without requiring large long-range kernel weights.
\end{itemize}

\section{Architecture Diagrams}
\label{sec:architecture-diagrams-appendix}

\cref{fig:globe_io_mapping} illustrates the input-output mapping of a \globe{} model.

\begin{figure}[htb]
    \input{figures/globe_io_mapping}
    \caption{The input-output mapping of a \globe{} model. Inputs consist of boundary meshes partitioned by boundary condition type, global scalar and vector parameters, reference length scales, and query point locations. Outputs are scalar and vector fields evaluated at the query points.}
    \label{fig:globe_io_mapping}
\end{figure}

\section{Detailed Background on Integral Boundary Layer Methods}
\label{sec:ibl-detailed}

For 2D aerodynamic flows without massive separation, specialized integral boundary layer (IBL) methods such as XFOIL and MSES by \citet{drelaXFOILAnalysisDesign1989,drelaUsersGuideMSES2007} provide an existence proof that dramatically faster and more accurate methods are possible. IBL methods decompose the flow into a thin boundary layer and an inviscid outer region, which are then coupled. The inviscid outer flow can be modeled as a potential flow (a linear homogeneous PDE) and discretized using the boundary-element method (BEM), which represents only the boundary geometry rather than the entire volume \citep{martinssonFastDirectSolvers2020}. This yields 100--1000$\times$ reductions in degrees of freedom and runtime relative to volumetric RANS, and on their target attached-flow problems IBL methods often achieve \emph{higher} accuracy than RANS CFD \citep{coderComparisonsTheoreticalMethods2014,morgadoXFOILVsCFD2016,adlerCFDNotCFD2022}.

At high Reynolds numbers and with mostly attached flows, thin shear layer approximations apply, as described in detail by \citet{drelaFlightVehicleAerodynamics2013,drelaAerodynamicsViscousFluids2019}. The boundary layers when represented in integral form are governed by the von Kármán momentum integral equation, which is parabolic in the streamwise coordinate and hence far more tractable than the full RANS system, as in \citet{drelaXFOILAnalysisDesign1989}.

This boundary-centric formulation is the conceptual ancestor of \globe{}: explicitly decomposing the flow into boundary and outer regions, with global information transfer mediated through boundary-only integrals, is a powerful inductive bias that we adopt and extend. Unlike analytical IBL methods, however, \globe{} does not rely on closed-form decompositions specific to 2D attached flow; the kernel functions are learned end-to-end and the architecture extends naturally to 3D and to mixed boundary conditions. Extending classical IBL itself to 3D has long remained elusive despite considerable effort \citep{drelaThreeDimensionalIntegralBoundary2013,zhangThreedimensionalIntegralBoundary2022}; \globe{} sidesteps this difficulty by replacing analytical decomposition with learned kernel superposition.

\section{Illustrations of Kernel Functions}
\label{sec:kernel-illustrations}

In this appendix we visualize an example kernel function $\kernel$, which aids in demonstrating some of the kernel's key mathematical properties, as well as its expressive range. While the example shown here is in 2D for ease of visualization, all constructions and guarantees (translation/rotation/parity equivariance, discretization invariance, and far-field behavior) are identical in 3D; see \crefrange{sec:kernel-functions}{sec:multiscale} for descriptions on the minor architectural changes associated with this.

Each figure shows a scalar field $\phi$ and a vector field $\vect{u}$ produced by $\kernel$ when evaluated from boundary face ``sources'' to a dense set of ``target'' points (\cref{eq:relative-position}) in physical space ($\RR^2$). The axes in all figures represent the $x$- and $y$-coordinates of the target points in nondimensional spatial coordinates (i.e., divided by the reference length $\ell$, as discussed in \cref{sec:kernel-appendix}). For simplicity, and to facilitate a later discussion on symmetry, all visualizations use kernel functions that do not include any global scalars or global vectors as inputs; only per-source vectors (when present) are provided. The parameters of the kernel function's learned \pade{}-approximant MLP core are their initial pseudorandomly generated values, and are the same for the kernels across all figures.

\begin{figure}[htb]
    \centering
    \includegraphics[width=\textwidth]{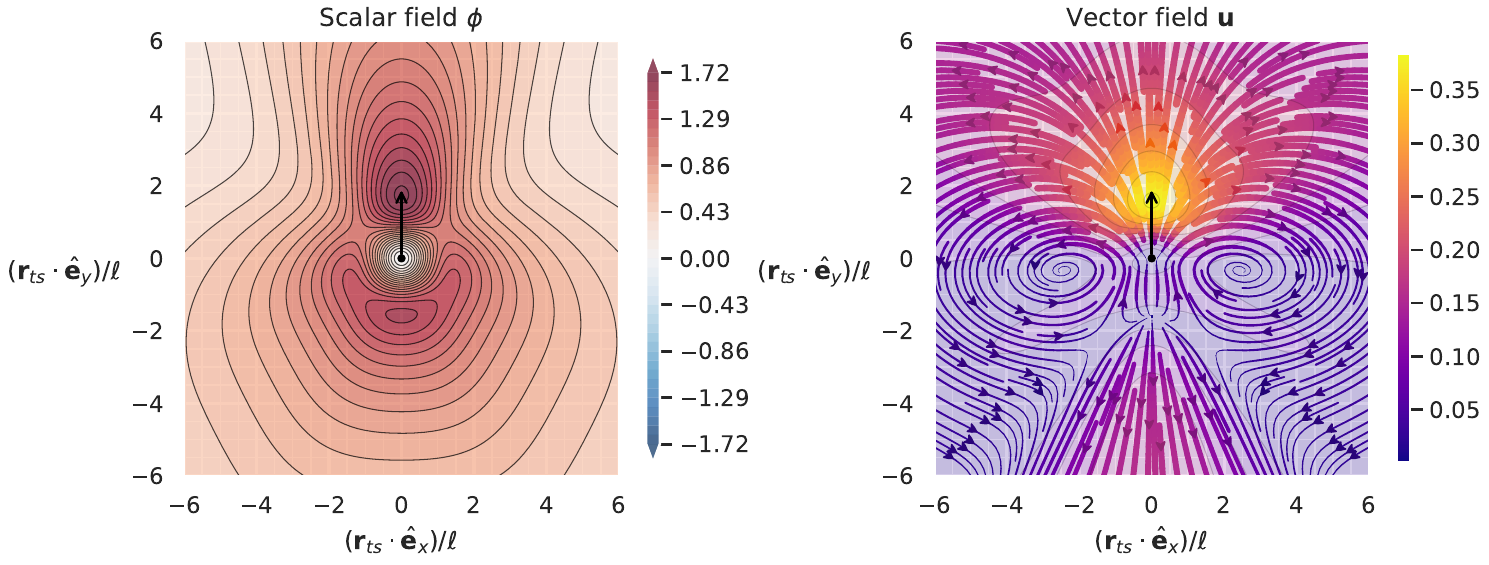}
    \caption{Single-source kernel evaluation in 2D. The source (black dot) is located at the origin and carries a vector input (black arrow). The far-field envelope (\cref{sec:far-field-decay-appendix}) drives self-influence to zero and enforces inverse-distance decay.}
    \label{fig:kernel-viz-0}
\end{figure}

\paragraph{Single source at the origin.} We begin with a single source at the origin. The source carries a vector-valued input (e.g., the unit normal of the boundary face, or a vector-valued boundary condition), which participates in the invariant feature construction (\cref{sec:kernel-engineered-features-appendix}) and later vector reprojection (\cref{sec:vector-reprojection-appendix}). \cref{fig:kernel-viz-0} shows the resulting fields, with the source denoted by a black dot and its associated vector-valued input denoted by a black arrow. The envelope in \cref{sec:far-field-decay-appendix} enforces two asymptotic properties: (i) self-influence vanishes at the origin via the $(1-e^{-\norm{\bm r_{ts}}^2})$ factor, avoiding overly localized kernels; and (ii) influence decays as $\sim 1/r$ in 2D and $\sim 1/r^2$ in 3D as $\norm{\bm r_{ts}}\to\infty$, ensuring bounded far-field behavior and stable spatial extrapolation.

\begin{figure}[htb]
    \centering
    \includegraphics[width=\textwidth]{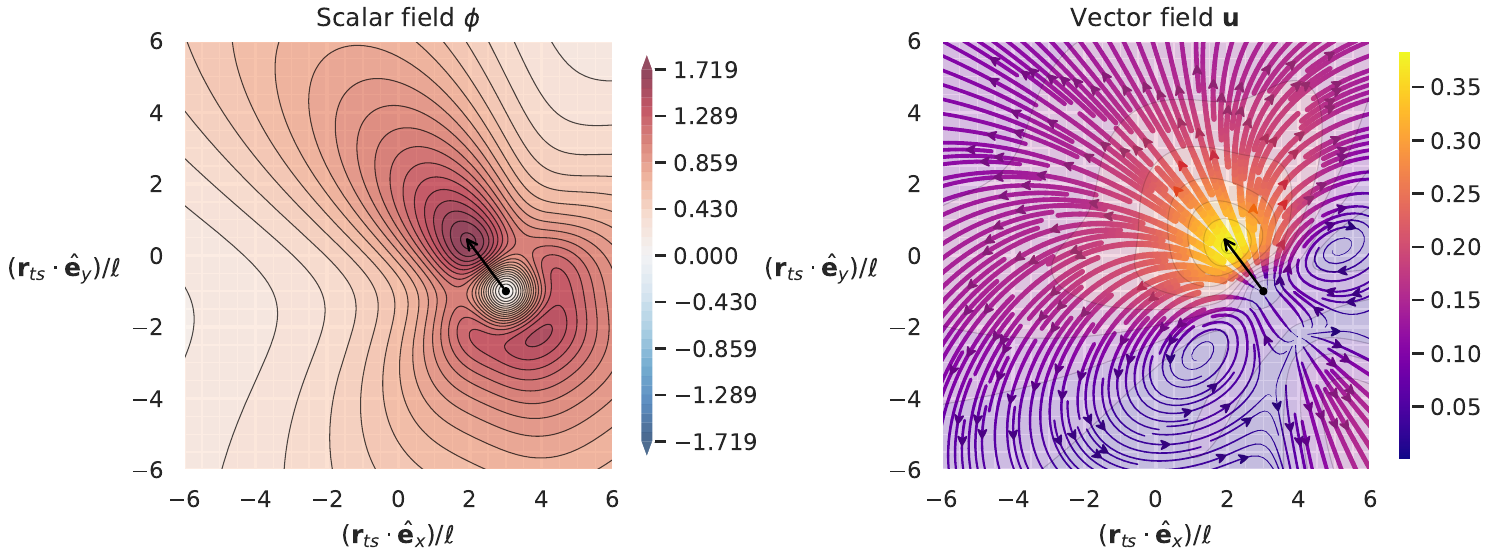}
    \caption{Equivariance demonstration. Rigidly translating/rotating the inputs (source point and its vector) rigidly translates/rotates the output fields without distortion (\cref{sec:vector-reprojection-appendix}).}
    \label{fig:kernel-viz-1}
\end{figure}

\paragraph{Equivariance under translation and rotation.} If we translate and rotate all inputs (the source location and any associated vectors), the outputs translate and rotate \emph{exactly}. This follows directly from using only the relative position vector $\bm r_{ts}=(\bm x_t-\bm x_s)/\ell$ (\cref{eq:relative-position}) and invariant scalar features; rotational information is reintroduced solely via local reprojection (\cref{sec:vector-reprojection-appendix}). Because no global vectors or scalars are present and the single source vector in \crefrange{fig:kernel-viz-0}{fig:kernel-viz-1} is vertical, the input set is invariant under the reflection $x\mapsto -x$, and parity equivariance therefore implies mirror-symmetric outputs. \cref{fig:kernel-viz-1} shows the same single-source pattern after a rigid transformation of the inputs; the fields are correspondingly transformed, confirming exact translation- and rotation-equivariance.

\begin{figure}[htb]
    \centering
    \includegraphics[width=\textwidth]{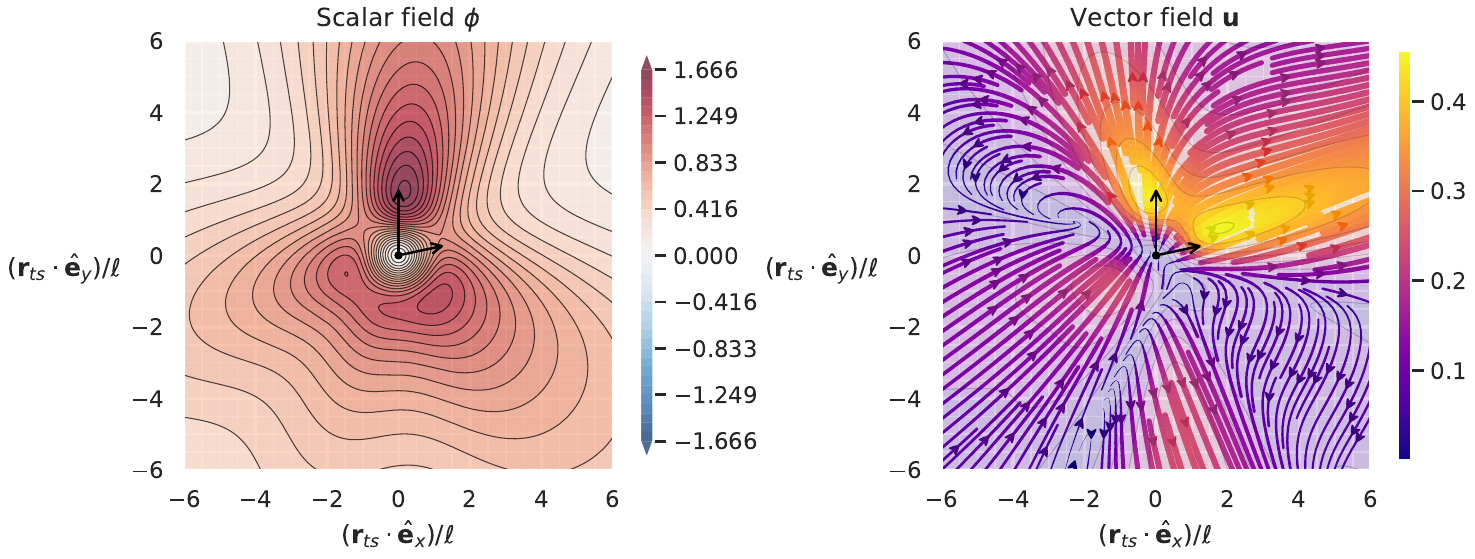}
    \caption{Adding a second source-associated vector breaks left-right symmetry, yielding richer, directional patterns (\cref{sec:vector-reprojection-appendix}).}
    \label{fig:kernel-viz-2}
\end{figure}

\paragraph{Breaking left-right symmetry with an additional vector.} Up to now, the configuration has one source-associated vector. The fields are left-right symmetric because, with no global vectors or scalars and a vertical source vector, the inputs are invariant under left-right reflection; parity-equivariance then mandates symmetry. To explicitly break this invariance, we provide a second source-associated vector (here oriented toward the right), so the input set is no longer reflection-symmetric. The resulting outputs need not be mirror-symmetric, as shown in \cref{fig:kernel-viz-2}. This behavior follows from the invariant feature construction together with equivariant vector reprojection; the network receives no absolute coordinate frame, and directionality enters only through bases derived from the input vectors (\cref{sec:vector-reprojection-appendix}).

Note, however, that the resulting fields of \cref{fig:kernel-viz-2} are \emph{not} simply a superposition of the kernel functions evaluated twice with different source-associated input vectors -- the two input vectors mutually interact in a manner that is, in general, nonlinear with respect to the input vectors.

\begin{figure}[htb]
    \centering
    \includegraphics[width=\textwidth]{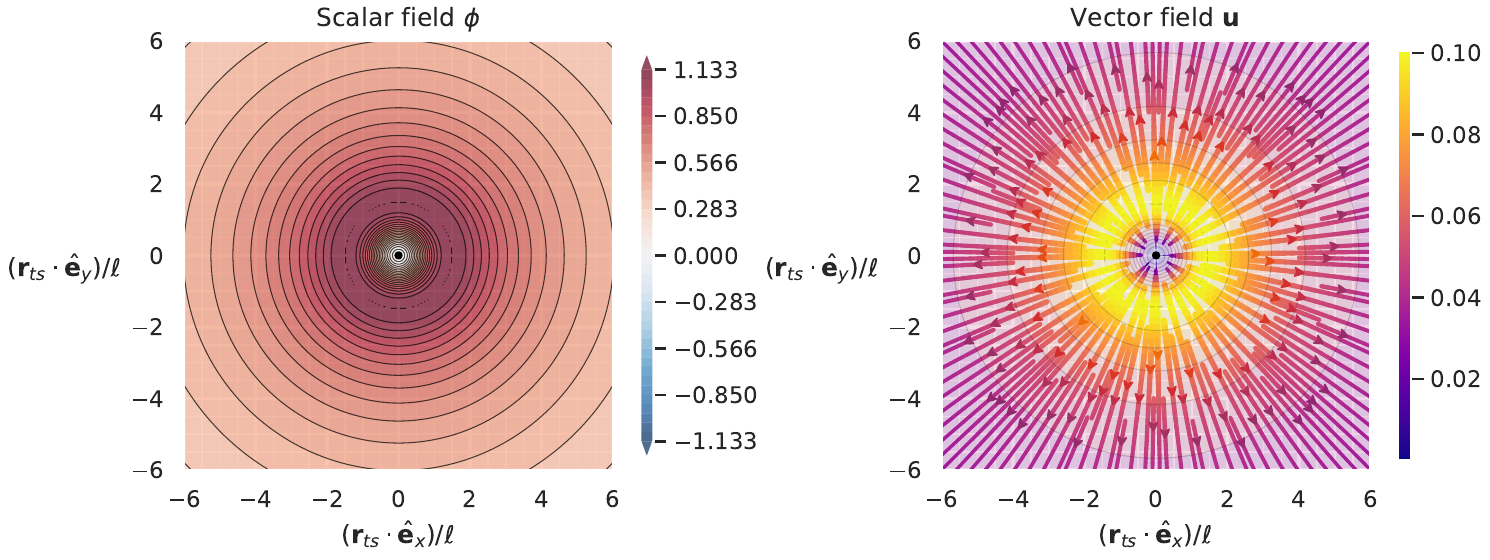}
    \caption{With no source-associated input vectors, the scalar and vector outputs are necessarily radially symmetric. Directionality cannot be inferred without directional inputs.}
    \label{fig:kernel-viz-3}
\end{figure}

\paragraph{Necessary radial symmetry without vector information.} Conversely, if no source-associated vectors are provided, there is no directional information with which to break symmetry -- this is a direct consequence of the symmetry principle described by \citet{curieSymetrieDansPhenomenes1894}. The only geometric input is $\bm r_{ts}$'s magnitude, and the learned core only processes (rotationally invariant) scalars; therefore both $\phi$ and $\bm u$ must be radially symmetric. \cref{fig:kernel-viz-3} illustrates the resulting radially symmetric fields. Once again, this symmetry is a consequence of the model's invariants rather than an explicit hand-coded constraint.

\begin{figure}[htb]
    \centering
    \includegraphics[width=\textwidth]{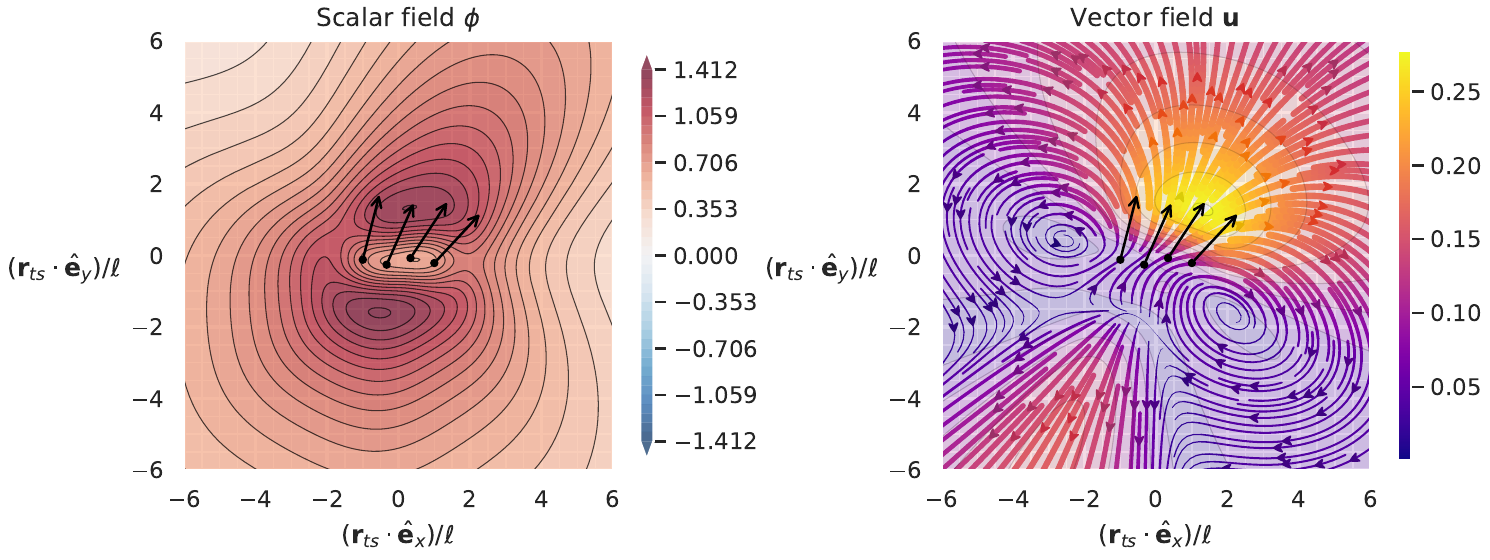}
    \caption{Superposition from four sources placed along a parametric curve, with per-source strengths reduced to preserve total influence (cf. area weighting in \cref{eq:aggregation-over-sources}).}
    \label{fig:kernel-viz-4}
\end{figure}

\paragraph{Superposition of multiple sources.} The final aggregation over sources (\cref{eq:aggregation-over-sources}) linearly superimposes influences weighted by per-face strengths $w_s$ and face areas $a_s$. \cref{fig:kernel-viz-4} shows four sources placed along a parametric curve, each with its own associated vector. To preserve total influence, each source's effective strength is reduced ($\propto 1/4$), mimicking the area scaling one would obtain from a coarser boundary discretization.

\begin{figure}[htb]
    \centering
    \includegraphics[width=\textwidth]{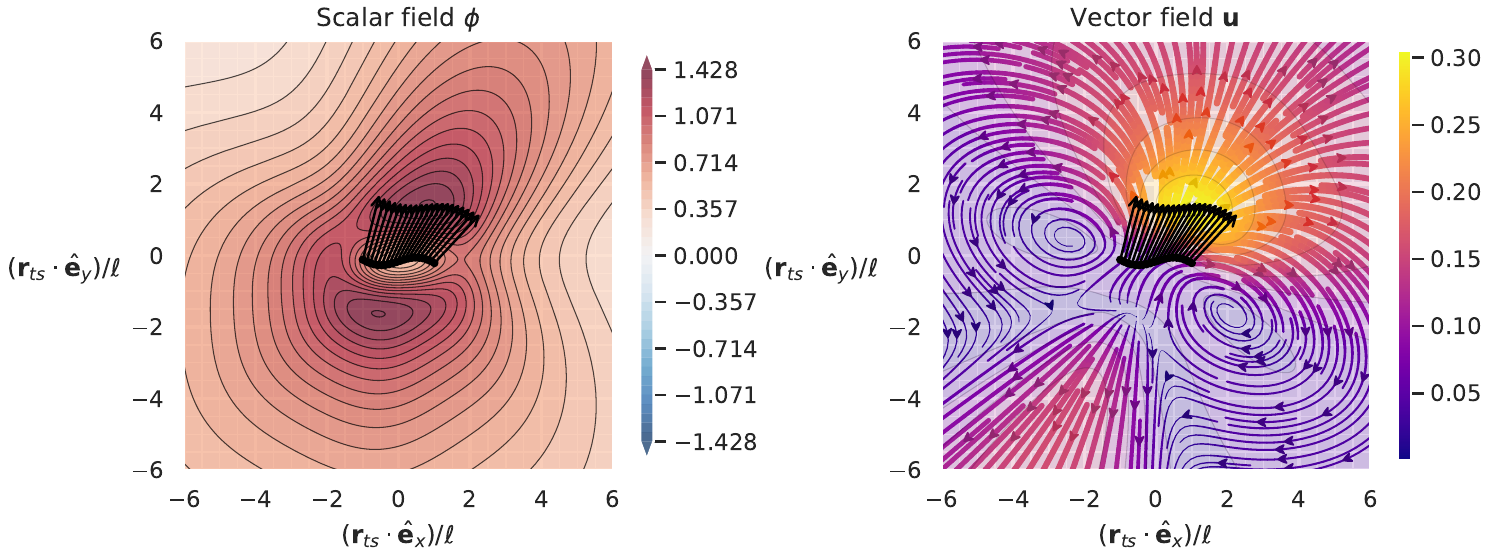}
    \caption{Refinement of \cref{fig:kernel-viz-4} to twenty sources. Despite the increased sampling density, the summed fields converge, demonstrating discretization-invariance under face-area weighting (\cref{eq:aggregation-over-sources}).}
    \label{fig:kernel-viz-5}
\end{figure}

\paragraph{Refinement and discretization invariance.} Refining the parametric sampling to twenty sources (each again with reduced strength to maintain the same total) yields fields that closely match those in \cref{fig:kernel-viz-4}. This mirrors mesh refinement in practice: as boundary faces are subdivided, areas $a_s$ decrease and the weighted sum converges. In the limit of vanishing face areas, the discrete contraction in \cref{eq:aggregation-over-sources}, $\sum_s w_s\,a_s\,\kernel_{ts}$, converges to the exact boundary surface integral $\int_{\partial\Omega} w(\bm{x}_s)\,\kernel(\bm{x}_t,\bm{x}_s)\,\mathrm{d}A_s$. \cref{fig:kernel-viz-5} shows this convergence in action, empirically illustrating the discretization-invariance discussed in \cref{eq:aggregation-over-sources}.

\section{Illustrations of \pade{}-Approximant MLPs}
\label{sec:pade-illustrations}

This appendix visualizes the functional behavior of randomly initialized \pade{}-approximant MLPs introduced in \cref{sec:pade-approx-mlp-appendix}. We use the notation \pade{} $[N/D]$ to denote a \pade{} MLP with numerator order $N$ and denominator order $D$, mirroring mathematical convention from traditional polynomial-based \pade{} approximants.

\begin{figure}[htb]
    \centering
    \includegraphics[width=\textwidth]{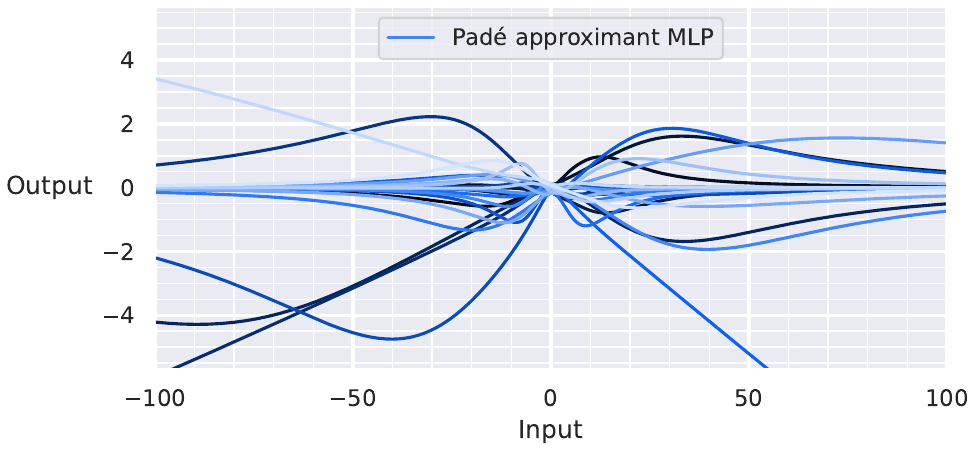}
    \caption{25 random initializations of a $\RR^1 \to \RR^1$ \pade{} $[1/3]$ MLP with layer sizes $[1, 64, 64, 1]$ and SiLU activations on each subnetwork, with each network represented as a different line.}
    \label{fig:pade-random-init}
\end{figure}

\begin{figure}[htb]
    \centering
    \includegraphics[width=\textwidth]{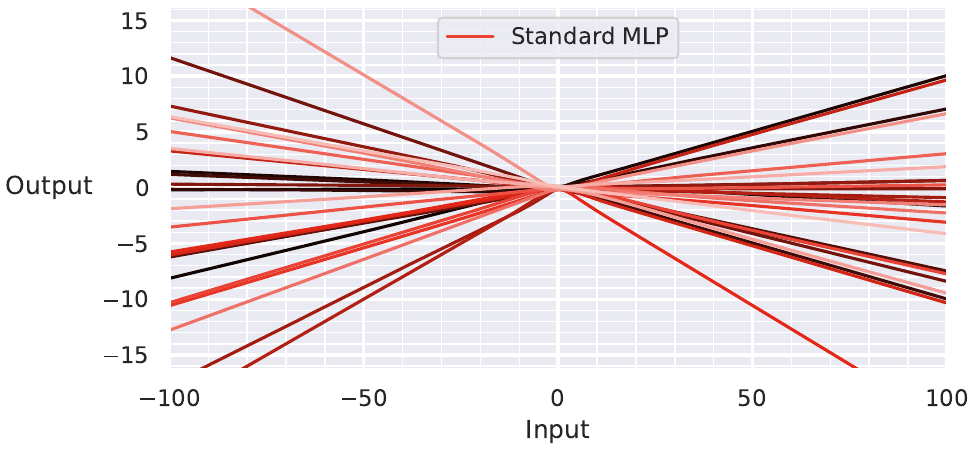}
    \caption{25 random initializations of a $\RR^1 \to \RR^1$ standard MLP with layer sizes $[1, 64, 64, 1]$ and SiLU activations, with each network represented as a different line.}
    \label{fig:mlp-random-init}
\end{figure}

To illustrate the diversity of functional forms that can be learned by \pade{}-approximant MLPs, \cref{fig:pade-random-init} shows 25 random initializations of a \pade{} $[1/3]$ MLP. For comparison, \cref{fig:mlp-random-init} shows 25 random initializations of a standard MLP with the same layer sizes and activations.

While both types of MLPs satisfy the universal approximation theorem, the classes of functions that these networks most readily express are quite different. We suggest that the functional forms expressed by \pade{}-approximant MLPs may be better suited for representing typical Green's functions arising in physics-based PDEs.

\section{Discussion of \globe{}'s Mathematical Properties}
\label{sec:mathematical-proofs}

We now formally establish that \globe{} satisfies the design desiderata of \cref{sec:necessary-attributes}.

\paragraph{Translation-equivariance.} All positional information enters the kernel through the relative position $\vect{r}_{ts} = (\vect{x}_t - \vect{x}_s)/\ell$ (\cref{eq:relative-position}). Under translation $\vect{x} \mapsto \vect{x} + \vect{c}$, we have $\vect{r}_{ts} \mapsto \big[(\vect{x}_t + \vect{c}) - (\vect{x}_s + \vect{c})\big]/\ell = \vect{r}_{ts}$, unchanged. All subsequent operations depend only on $\vect{r}_{ts}$ and other translation-invariant quantities (normals, boundary data), so translating all inputs translates outputs identically.

\paragraph{Rotation-equivariance.} The kernel processes only rotationally invariant scalars: magnitudes $\smoothlog(\norm{\vect{v}_i})$ and pairwise angles via Legendre polynomials. Physical vectors are reconstructed by reprojecting onto local bases derived from input vectors (\cref{sec:vector-reprojection-appendix}). Under rotation $\mathcal{R}$, each input vector transforms as $\vect{v}_i \mapsto \mathcal{R}\vect{v}_i$, but magnitudes and angles remain invariant. The basis vectors transform as $\vect{b}_j \mapsto \mathcal{R}\vect{b}_j$, and the reprojection $\sum_j c_j \vect{b}_j \mapsto \sum_j c_j (\mathcal{R}\vect{b}_j) = \mathcal{R}\left(\sum_j c_j \vect{b}_j\right)$ yields exact equivariance. The final calibration (linear for vectors) preserves this property.

\paragraph{Parity-equivariance.} The architecture respects parity when the optional tangential/vortex basis vectors -- $\vect{e}_\theta$ in 2D and $\vect{e}_\phi$ in 3D -- are disabled (these are pseudovectors), as is the default. With only true vectors in the reprojection basis (radial, axis, dipole directions), all bases transform correctly under reflection, and the Legendre polynomial features are parity-even. Full parity preservation including tangential bases requires tracking vector vs. pseudovector semantics (\cref{sec:vector-reprojection-appendix}); this is left as future work. Tangential bases are disabled in every reported configuration. AirFRANS uses dense evaluation, so its predictions come from the parity-equivariant dense path; DrivAerML uses the approximate hierarchical path, for which we have not measured an empirical parity tolerance.

\paragraph{Numerical checks and scope.} The arguments above establish equivariance of the dense construction in exact arithmetic. Separately, the reference implementation's unit tests check translations, a $60^\circ$ rotation, and a Householder reflection for 2D and 3D scalar and vector cases; all 18 tests pass at \texttt{rtol} $=$ \texttt{atol} $= 10^{-5}$. The Barnes-Hut path is not exactly rotation-equivariant at finite $\theta$ (\cref{sec:hierarchical-acceleration-appendix}), and discretization-invariance below is an argument about the continuum limit: the decimation study of \cref{sec:non-watertight} tests robustness to missing faces, not convergence under mesh refinement.

\paragraph{Discretization-invariance.} The aggregation in \cref{eq:aggregation-over-sources} weights each source by its face area $a_s$. As the boundary mesh is refined with $a_s \to 0$, the discrete sum $\sum_s w_s a_s \kernel_{ts}$ converges to the continuous boundary integral $\int_{\partial\Omega} w(\vect{x}_s) \kernel(\vect{x}_t, \vect{x}_s) \,\mathrm{d}A_s$. The kernel itself is mesh-independent (parameterized by continuous positions), so predictions converge to a well-defined limit as discretization is refined.

\paragraph{Units-invariance.} All positional features are nondimensionalized by reference lengths $\ell_i$. All input scalars and vectors must be dimensionless, a requirement that should be enforced during data preprocessing. Our implementation for the AirFRANS case study (\cref{sec:task-airfrans,sec:preprocessing-details}) demonstrates one approach to ensuring this property. Kernel operations preserve dimensionlessness throughout, and physical units are recovered only during post-processing. Therefore, the entire forward pass operates in a unitless regime, guaranteeing units-invariance.

\section{Experimental Details}
\label{sec:experimental-details}

\subsection{Preprocessing and Nondimensionalization Details}
\label{sec:preprocessing-details}

We use the preprocessed AirFRANS meshes and follow a strictly unitless pipeline. For each sample, we load the internal VTU and the two VTP boundaries (freestream and airfoil) and construct an explicit boundary mesh for the airfoil with a \texttt{no\_slip} BC using the face centers and normals. Let \(\bm{U}_\infty\) denote the mean freestream velocity on the freestream boundary, with magnitude \(|\bm{U}_\infty|\), density \(\rho=1\,\mathrm{kg/m^3}\)\footnote{While some secondary sources report \(\rho=1.204\,\mathrm{kg/m^3}\), the OpenFOAM cases in AirFRANS are configured with \(\rho=1\). This choice preserves constant far-field total pressure, consistent with incompressible flow assumptions.}, and kinematic viscosity \(\nu=1.56\times10^{-5}\,\mathrm{m^2/s}\). We define the dynamic pressure \(q_\infty=\tfrac{1}{2}\rho|\bm{U}_\infty|^2\) and compute the following nondimensional fields on the internal point set:
\begin{align*}
    \frac{\bm{U}}{|\bm{U}_\infty|},\quad \frac{\Delta\bm{U}}{|\bm{U}_\infty|} & = \frac{\bm{U}-\bm{U}_\infty}{|\bm{U}_\infty|},                                                                        \\
    C_p                                                                       & = \frac{p}{q_\infty},\qquad C_{pt} = \frac{p + q}{q_\infty},\quad q = q_\infty\norm{\frac{\bm{U}}{|\bm{U}_\infty|}}^2, \\
    \ln\bigl(1+\nu_t/\nu\bigr)                                                & ,\qquad \bm{C}_{F,\text{shear}}=\frac{2\nu\,\bm{S}\,\vect{\hat{n}}_w}{q_\infty},
\end{align*}
where \(\bm{S}=\tfrac{1}{2}(\nabla\bm{U}+\nabla\bm{U}^{\!T})\) is the symmetric strain-rate tensor and \(\vect{\hat{n}}_w\) are per-point airfoil normals sampled at the surface-only points.

During training, we mask out all points in the dataset with obviously non-physical total pressures ($C_{pt}>1.02$), as these denote significant violations of energy conservation -- this affects 0.003\% of points across the entire dataset.

Reference lengths provided to \globe{} are \(\ell\in\{c_\text{ref},\ \delta_{FS}\}\), where \(c_\text{ref}\) is the airfoil chord and \(\delta_{FS} = \sqrt{\nu\,c_\text{ref}/|\bm{U}_\infty|}\) is a freestream viscous thickness scale.

\paragraph{On excluding signed distance fields.} While AirFRANS includes signed distance fields (SDFs), we deliberately exclude them as model input features. SDF features pose two fundamental problems in PDE surrogate modeling: (1) SDFs have no direct physical role in the governing equations, as they are purely geometric constructs. If proximity information were physically meaningful, better approaches would use fields derived from physical PDEs, such as the signed heat method of \citet{fengHeatMethodGeneralized2024}. (2) More critically, SDFs are inherently non-smooth: gradients exhibit discontinuities at equidistant loci (medial axes/surfaces), violating the $C^\infty$ continuity that \globe{}'s kernel architecture maintains. The relative position $\vect{r}_{ts}$ (\cref{eq:relative-position}) provides sufficient geometric information without these pathologies.

\paragraph{On redundant output fields, gradients, and self-consistency.}
Several output fields are not independent: for example, given velocity and $C_p$, one can infer $C_{pt}$ (and conversely), and wall-shear vectors could be obtained from velocity gradients.\footnote{Specifically, one can differentiate the velocity field to obtain the velocity gradient tensor, apply the Newtonian constitutive relation to form the viscous stress tensor, and contract with the surface normal to obtain the wall-shear stress vector.} Therefore, it would be possible to implement a Sobolev-style objective to enforce these relationships implicitly through the loss.

However, we find that directly predicting both underlying and derived quantities yields measurably better accuracy and faster convergence, mirroring empirical observations in Sobolev training, where such redundancy stabilizes optimization and improves fidelity in sharp-gradient regions (e.g., near-wall shear). A further practical benefit is trustworthiness assessment during extrapolation: redundant channels enable lightweight ``self-inconsistency'' diagnostics at inference (e.g., comparing the explicitly predicted $C_{pt}$ to the value implied by $\vect{U}$ and $C_p$). Elevated self-inconsistency flags untrustworthy predictions, an important consideration for CAE workflows that is often underemphasized in ML surrogate literature. Conversely, computing derived quantities from a minimal set of primary fields via exact differential relations could further strengthen the physical inductive bias, and remains an interesting direction for future work.

\subsection{Hyperparameters and Training Details}
\label{sec:hyperparameters}

\paragraph{Implementation note: TensorDict data structure.}
The reference implementation uses the \td{} data structure developed by \citet{bou2023torchrl} to pass data between components. \globe{} differs from most other ML architectures for PDEs in that it explicitly tracks tensorial semantics (distinguishing physical scalars from physical vectors) throughout the network, as required for rotational equivariance. Inputs and outputs of various components are therefore commonly dictionaries of scalars and vectors, rather than the single large concatenated tensor typical in other model architectures. This tends to result in many code patterns where an operation is performed for each tensor leaf of a dictionary, causing a speed penalty due to GPU kernel launch overhead. Fortunately, \td{} allows such operations to be fused into a single launch, yielding significant speedups.\footnote{In benchmarks on GB200 systems on the AirFRANS task (\cref{sec:task-airfrans}), \td{} fused operations gave a $1.7\times$ training-epoch speedup over an equivalent implementation using Python dictionaries, with \texttt{torch.compile} enabled in both cases.}

\paragraph{Implementation used for the reported results.}
The reported AirFRANS models were trained with a pre-release implementation of \globe{}, while the Full and Scarce results are evaluated with the public NVIDIA PhysicsNeMo implementation (commit \texttt{e3dae21a3} with two minor compatibility fixes). Because the public release reorders the internal kernel-network features, loading pre-release checkpoints requires only a deterministic permutation of each kernel network's first-layer input columns and last-layer output rows. Under exact ($\theta = 0$) float32 evaluation, the two implementations agree to within $2\times10^{-4}$ relative error across all Full and Scarce metrics, matching to every reported digit; only the wall-shear rows of \cref{tab:airfrans-physical-metrics} differ, as their ground truth depends on the velocity-gradient reconstruction during preprocessing.

\paragraph{Training protocol.}
During training, we employ dynamic subsampling of prediction points: on each iteration, a limited number of prediction points are randomly chosen to be the query points for training (default: 2,048). The loss is a rotationally invariant per-field Huber objective\footnote{Chosen to reduce sensitivity to any remaining numerical artifacts in the CFD training data; for vector fields, the Huber threshold is applied to the vector norm rather than componentwise to preserve rotational invariance.} with scales $\left\{\Delta U/|U_\infty|:1,\ C_p:1,\ C_{pt}:1,\ \ln(1+\nu_t/\nu):5,\ C_{F,\text{shear}}:10^{-2}\right\}$, applied after masking invalid ground-truth entries. Optimization uses a combined optimizer: Muon by \citet{jordan2024muon} for matrix-shaped parameters and RAdamW by \citet{liuVarianceAdaptiveLearning2021} for all other parameters, with \texttt{ReduceLROnPlateau} scheduling. The base learning rate of $10^{-3}$ is scaled by $\sqrt{n_\text{GPU}\,n_q/2048}$, where $n_q$ is the number of query points per iteration. This initial learning rate is set at the start of training and retained across checkpoint resumptions; for the Full and Scarce runs, which launched on 32 GPUs with $n_q = 2048$, it is $10^{-3} \times \sqrt{32} \approx 5.66\times10^{-3}$. The scheduler's minimum learning rate is $1/64$ of this scaled value ($8.84\times10^{-5}$). We train in float32, without automatic mixed precision, and use \texttt{torch.compile} in the \texttt{max-autotune} mode (i.e., including CUDA Graphs) for the Full and Scarce models. The Full and Scarce models were trained on NVIDIA H100 GPUs (Full: 32 GPUs, increased to 80 late in training; Scarce: 32 GPUs). Distributed data-parallel training is used when multiple GPUs are available, and parallelization experiments show near-linear scaling of train-time effective throughput with increasing accelerator count, even at the largest-benchmarked configuration with 128 GB200 GPUs. Validation follows the official split per regime and uses the official prediction set (i.e., without any subsampling or masking) for both the accuracy metrics and qualitative visualizations in \cref{sec:results}.

\cref{tab:hyperparameters} lists all hyperparameters used in the main AirFRANS experiments. These values were selected through systematic but limited exploration: we compared 2-layer (128 neurons) versus 3-layer (64 neurons) architectures, swept learning rate over \{$10^{-2}$, $10^{-3}$, $10^{-4}$\}, and explored three weight decay values. No extensive hyperparameter tuning was performed; the reported configuration represents a reasonable baseline rather than an exhaustively optimized setup, suggesting that further improvements may be possible with more thorough search.
In total, these hyperparameters lead to a model with 117,282 trainable parameters.

\begin{table}[!htb]
    \centering
    \caption{Hyperparameters for AirFRANS experiments. The learning-rate and hardware rows are those of the Full and Scarce runs.}
    \label{tab:hyperparameters}
    \begin{tabular}{ll}
        \toprule
        Hyperparameter                                               & Value                             \\
        \midrule
        \multicolumn{2}{l}{\textit{Architecture}}                                                        \\
        \quad Number of communication hyperlayers ($H$)              & 2                                 \\
        \quad Hidden layer sizes                                     & [64, 64, 64]                      \\
        \quad Number of latent scalars per hyperlayer                & 12                                \\
        \quad Number of latent vectors per hyperlayer                & 6                                 \\
        \quad Number of spherical harmonics ($n_{\text{harmonics}}$) & 1                                 \\
        \quad \pade{} numerator order ($N$)                          & 2                                 \\
        \quad \pade{} denominator order ($D$)                        & 2                                 \\
        \quad Reference lengths                                      & $\{c_{\text{ref}}, \delta_{FS}\}$ \\
        \midrule
        \multicolumn{2}{l}{\textit{Training}}                                                            \\
        \quad Query points per iteration                             & 2,048                             \\
        \quad Boundary face downsampling ratio                       & 1.0 (No downsampling)             \\
        \quad Learning rate (base)                                   & $10^{-3} \times \sqrt{32} \approx 5.66\times10^{-3}$ \\
        \quad Weight decay                                           & $10^{-4}$                         \\
        \quad Scheduler                                              & ReduceLROnPlateau                 \\
        \quad \quad Patience                                         & 400 epochs                        \\
        \quad \quad Factor                                           & 0.5                               \\
        \quad \quad Min LR                                           & $8.84\times10^{-5}$ (base LR$/64$) \\
        \quad Optimizer                                              & Muon (matrices) + RAdamW (others) \\
        \quad Precision                                              & float32 (no AMP)                  \\
        \quad Training hardware                                      & NVIDIA H100, 32--80 GPUs          \\
        \quad Huber loss delta                                       & 1.0                               \\
        \bottomrule
    \end{tabular}
\end{table}

\subsection{Computational Efficiency on AirFRANS}
\label{sec:results-cost}

We report the execution cost of \globe{} during both training and inference. These are execution measurements of \globe{} alone, not a comparison with CFD or other surrogates.

\paragraph{Training efficiency (AirFRANS).} Training was performed using distributed data parallelism; with the pre-release implementation, we observed near-linear throughput scaling when tested up to the largest-benchmarked configuration of 128 NVIDIA GB200 GPUs. This efficient scaling is enabled by the dense compute intensity of the $\mathcal{O}(n_{\text{src}}^2)$ boundary-to-boundary interactions, which dominates communication overhead.

With the pre-release implementation on 40 GB200 GPUs and the \emph{Full} split (\cref{sec:task-airfrans}), we record a training epoch time of 6.83 seconds, or 0.341 GPU-seconds per training sample across the 800 training cases. Validation is substantially faster because it requires only a forward pass without storing intermediate activations for backpropagation, taking 0.083 GPU-seconds per validation sample (at 2,048 subsampled query points). With the public NVIDIA PhysicsNeMo implementation (float32, 2,048 query points, leaf size 512), the Full-split epoch loop -- including the validation pass, logging, and checkpointing -- takes 16.9 seconds on 40 GB300 GPUs and 9.2 seconds on 80 GB300 GPUs.

\paragraph{Inference efficiency (AirFRANS).} Full-field inference is slower than a training-time validation pass because it queries every vertex of the volume mesh (${\sim}180\times10^3$ points) rather than a 2,048-point subsample. On a single GB200 GPU, the pre-release implementation requires 2.326 seconds per test case. On a single GB300 GPU in float32 ($\theta = 0$, \texttt{leaf\_size=512}, \texttt{prediction\_chunk\_size=32768}), full-mesh inference takes 3.4 seconds per case with the public NVIDIA PhysicsNeMo implementation and 3.20 seconds with the pre-release implementation.

Inference memory footprint can be driven down to very low levels by chunking over query points. With the pre-release implementation and a chunk size of 128 points, peak memory consumption is held to approximately 700 MB, allowing deployment on consumer-grade hardware.

\section{Extended Results}
\label{sec:extended-results}

This appendix provides extended results including extrapolation task performance, qualitative visualizations, and non-watertight mesh experiments.

\subsection{Reynolds and Angle-of-Attack Extrapolation}
\label{sec:extrapolation-results}

\cref{tab:airfrans-reynolds,tab:airfrans-aoa} report performance on the extrapolation tasks. \globe{} maintains strong accuracy even when evaluated on Reynolds numbers and angles of attack outside the training distribution.

\begin{table}[htb]
    \centering
    \newcommand{\blank}{--}
    \caption{AirFRANS \textbf{Reynolds} (extrapolation) task: mean squared error on the official test cases, normalized by the training-set field variance as in \cref{tab:airfrans-full}; lower is better. External values are as reported in the cited works (not rerun). \globe{} values are single-run point estimates.}
    \label{tab:airfrans-reynolds}
    \resizebox{\linewidth}{!}{%
    \begin{tabular}{l c c c c}
        \toprule
                                     & \multicolumn{4}{c}{\textbf{Mean Squared Error, ``Reynolds'' task}}                                                                                           \\
        \cmidrule(lr){2-5}
        {Model}                      & {$\bar{u}_x$}                                                      & {$\bar{u}_y$}               & {$\bar{p}$}                 & {$\bar{p}_s$}               \\
        \midrule
        \textbf{\globe{} (our work)}                    & {\bfseries 0.0601} \enegtwo                                        & {\bfseries 0.0165} \enegtwo & {\bfseries 0.0144} \enegtwo & {\bfseries 0.0473} \enegtwo \\
        \midrule
        MLP \citep{bonnetAirfRANSHighFidelity2023}       & 9.5050 \enegtwo                                                    & 4.9240 \enegtwo             & 4.3000 \enegtwo             & 208.9800 \enegtwo           \\
        GraphSAGE \citep{bonnetAirfRANSHighFidelity2023} & 7.5580 \enegtwo                                                    & 3.4980 \enegtwo             & 3.8260 \enegtwo             & 17.9700 \enegtwo            \\
        PointNet \citep{bonnetAirfRANSHighFidelity2023}  & 9.4220 \enegtwo                                                    & 7.1290 \enegtwo             & 4.0110 \enegtwo             & 20.1300 \enegtwo            \\
        GUNet \citep{bonnetAirfRANSHighFidelity2023}     & 8.3830 \enegtwo                                                    & 5.2500 \enegtwo             & 4.4830 \enegtwo             & 20.5900 \enegtwo            \\
        \bottomrule
    \end{tabular}%
    }
\end{table}

\begin{table}[htb]
    \centering
    \newcommand{\blank}{--}
    \caption{AirFRANS \textbf{Angle of Attack} (extrapolation) task: mean squared error on the official test cases, normalized by the training-set field variance as in \cref{tab:airfrans-full}; lower is better. External values are as reported in the cited works (not rerun). \globe{} values are single-run point estimates.}
    \label{tab:airfrans-aoa}
    \resizebox{\linewidth}{!}{%
    \begin{tabular}{l c c c c}
        \toprule
                                     & \multicolumn{4}{c}{\textbf{Mean Squared Error, ``Angle of Attack'' task}}                                                                                             \\
        \cmidrule(lr){2-5}
        {Model}                      & {$\bar{u}_x$}                                                             & {$\bar{u}_y$}               & {$\bar{p}$}                  & {$\bar{p}_s$}                \\
        \midrule
        \textbf{\globe{} (our work)}                    & {\bfseries 0.4158} \enegtwo                                               & {\bfseries 0.2116} \enegtwo & {\bfseries 0.3536} \enegtwo  & {\bfseries 1.2665} \enegtwo  \\
        \midrule
        MLP \citep{bonnetAirfRANSHighFidelity2023}       & 6.9650 \enegtwo                                                           & 10.6300 \enegtwo            & 11.7110 \enegtwo             & 87.6200 \enegtwo             \\
        GraphSAGE \citep{bonnetAirfRANSHighFidelity2023} & 4.4350 \enegtwo                                                           & 9.4000 \enegtwo             & 10.9080 \enegtwo             & 76.3800 \enegtwo             \\
        PointNet \citep{bonnetAirfRANSHighFidelity2023}  & 8.6800 \enegtwo                                                           & 15.7960 \enegtwo            & 16.2370 \enegtwo             & 58.4600 \enegtwo             \\
        GUNet \citep{bonnetAirfRANSHighFidelity2023}     & 5.6890 \enegtwo                                                           & 10.3420 \enegtwo            & 14.8870 \enegtwo             & 69.6700 \enegtwo             \\
        \bottomrule
    \end{tabular}%
    }
\end{table}

\subsection{Qualitative Field Visualizations}
\label{sec:qualitative-results}

\begin{figure}[htb]
    \centering
    \includegraphics[width=\textwidth]{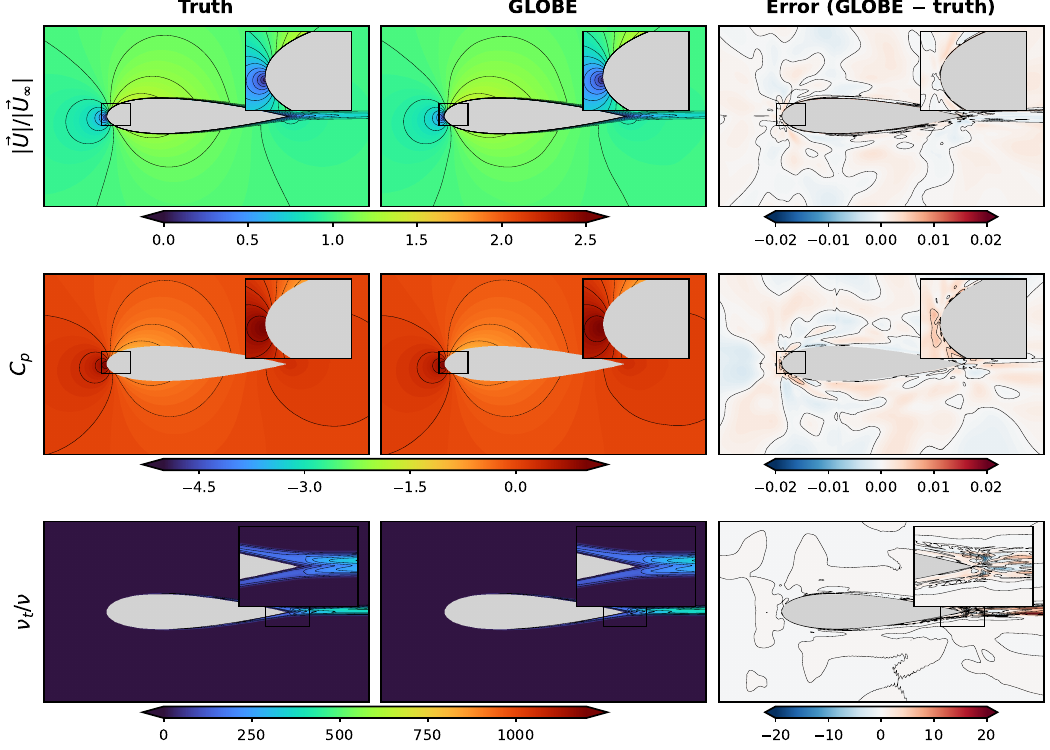}
    \caption{\globe{} model predictions on the first test case of the \textbf{Full} split. Rows show the nondimensional velocity magnitude $|\bm{U}|/|\bm{U}_\infty|$, pressure coefficient $C_p$, and turbulent viscosity ratio $\nu_t/\nu$; columns show ground truth, \globe{} prediction, and error (prediction minus truth). Each field and each error uses a fixed color range, shared with \cref{fig:airfrans-scarce-qual}. Insets zoom in on the leading edge for $|\bm{U}|/|\bm{U}_\infty|$ and $C_p$, and on the trailing edge and wake for $\nu_t/\nu$.}
    \label{fig:airfrans-full-qual}
\end{figure}

\begin{figure}[htb]
    \centering
    \includegraphics[width=\textwidth]{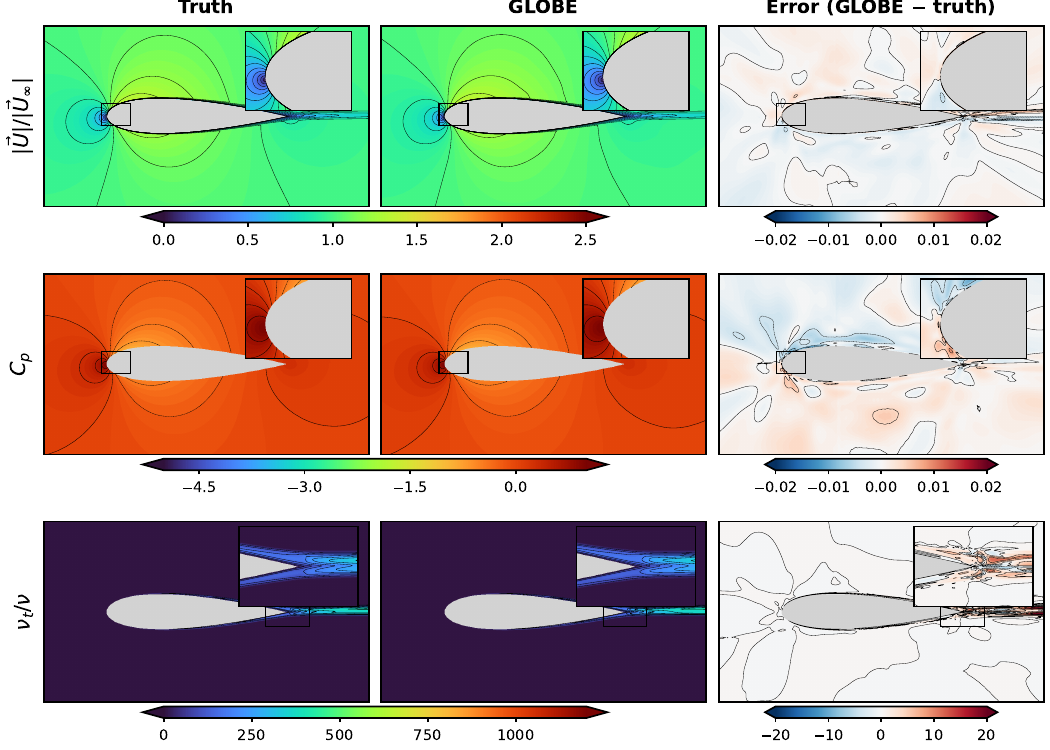}
    \caption{\globe{} model predictions on the first test case of the \textbf{Scarce} split (the same case as \cref{fig:airfrans-full-qual}). Rows show the nondimensional velocity magnitude $|\bm{U}|/|\bm{U}_\infty|$, pressure coefficient $C_p$, and turbulent viscosity ratio $\nu_t/\nu$; columns show ground truth, \globe{} prediction, and error (prediction minus truth). Each field and each error uses a fixed color range, shared with \cref{fig:airfrans-full-qual}. Insets zoom in on the leading edge for $|\bm{U}|/|\bm{U}_\infty|$ and $C_p$, and on the trailing edge and wake for $\nu_t/\nu$. With 200 training samples (one-quarter of the Full split), the boundary-layer and wake profiles remain well resolved.}
    \label{fig:airfrans-scarce-qual}
\end{figure}

\begin{figure}[htb]
    \centering
    \includegraphics[width=\textwidth]{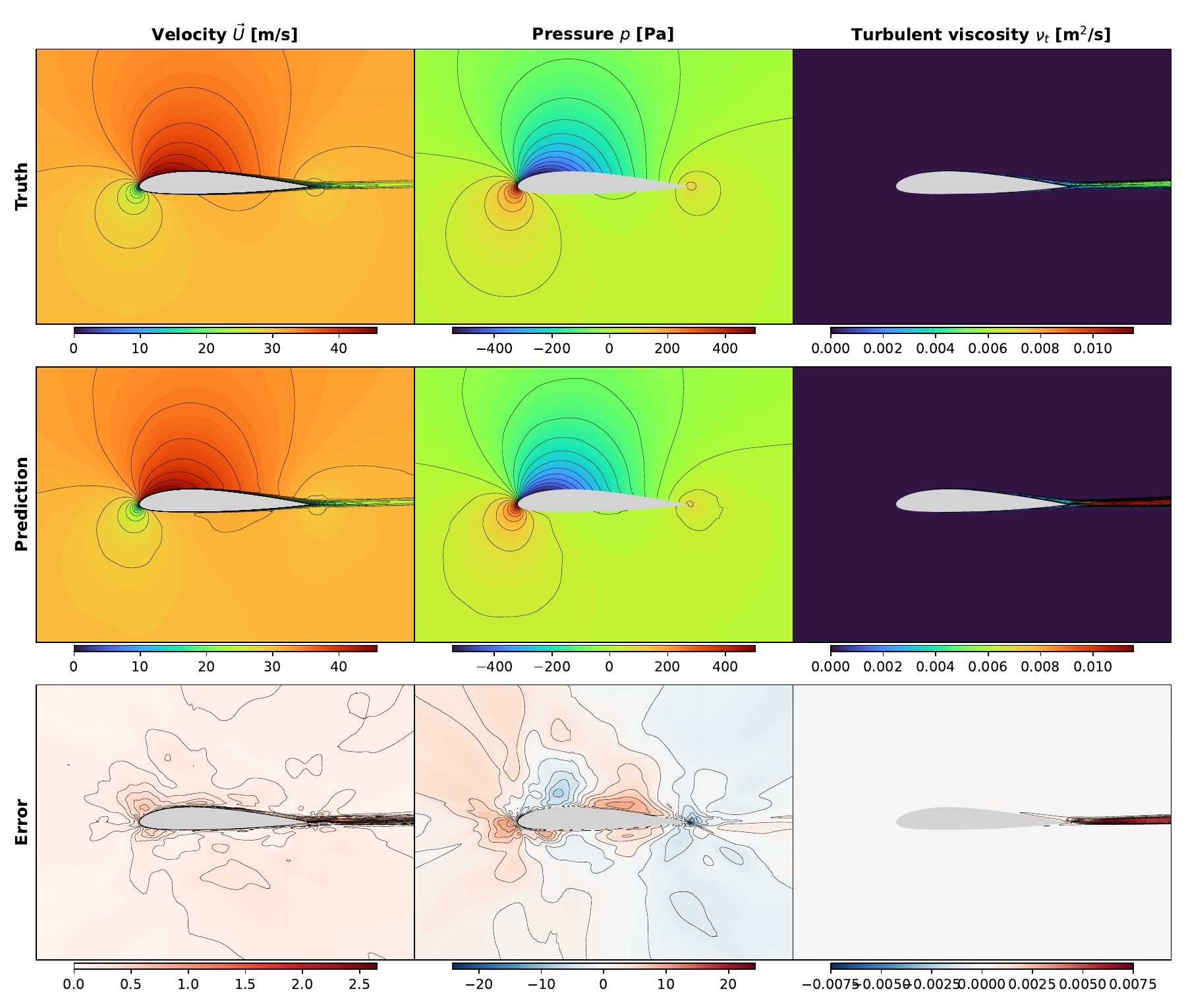}
    \caption{\globe{} model predictions on a test case from the \textbf{Reynolds extrapolation} split. Rows show ground truth, \globe{} predictions, and error; columns show velocity magnitude, pressure, and turbulent viscosity in dimensional units. Ground-truth and predicted fields share a color scale within each column, while the error panels use a separate scale per figure. Here, the largest errors occur in the boundary layer and thin wake.}
    \label{fig:airfrans-re-qual}
\end{figure}

\begin{figure}[htb]
    \centering
    \includegraphics[width=\textwidth]{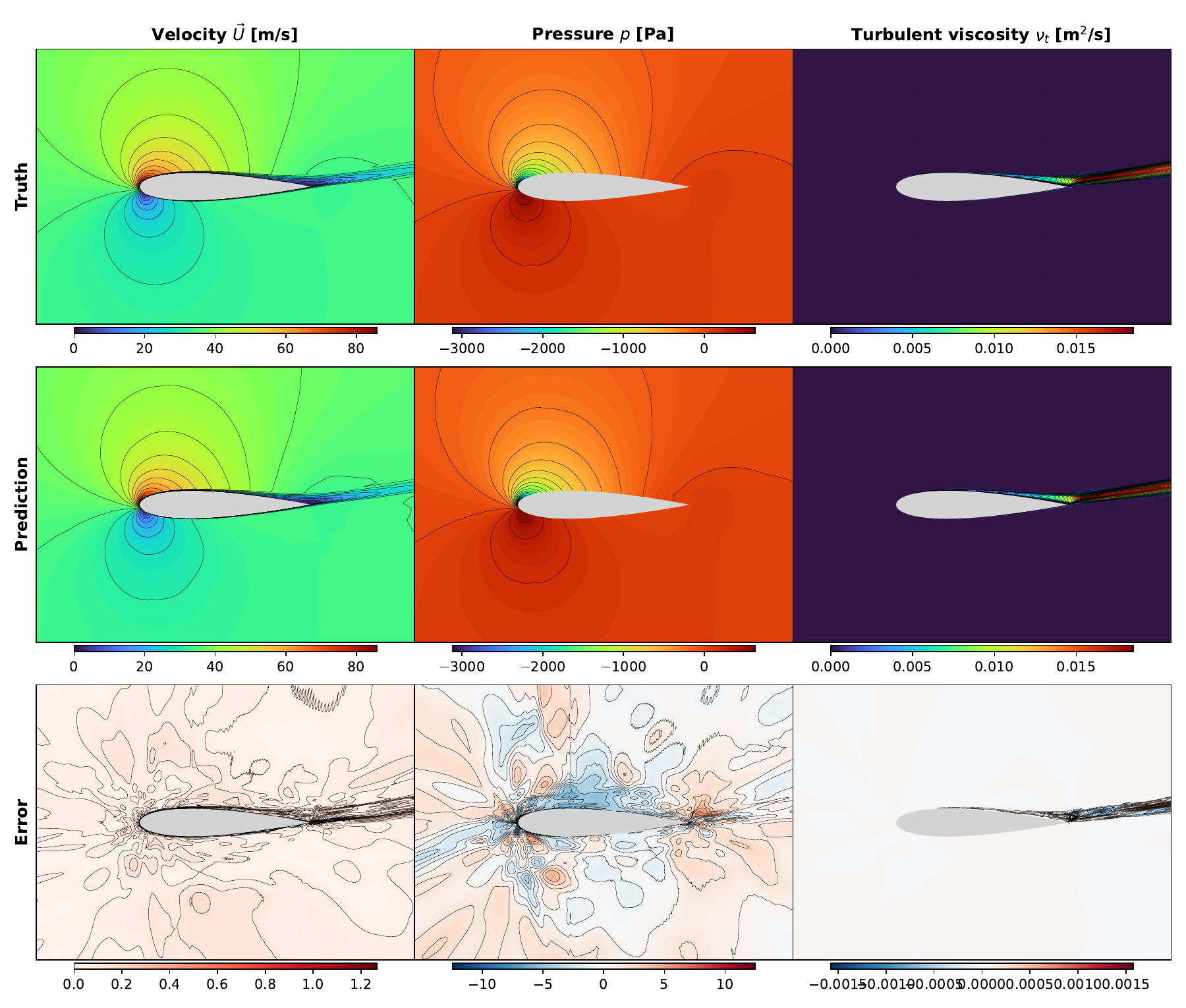}
    \caption{\globe{} model predictions on a test case from the \textbf{AoA extrapolation} split. Rows show ground truth, \globe{} predictions, and error; columns show velocity magnitude, pressure, and turbulent viscosity in dimensional units. Ground-truth and predicted fields share a color scale within each column, while the error panels use a separate scale per figure. At this angle of attack, just above the training range, the model captures the leading-edge suction peak and shifted stagnation point, with the largest errors concentrated in the suction peak, upper-surface boundary layer, and thickened wake.}
    \label{fig:airfrans-aoa-qual}
\end{figure}

\crefrange{fig:airfrans-full-qual}{fig:airfrans-aoa-qual} show \globe{} predictions on official test cases from each of the four tasks, demonstrating sharp boundary-layer gradients, coherent wake structures, and maintained accuracy even under extrapolation.\footnote{Test cases were selected systematically (first in alphabetical order for Full/Scarce; first low-Reynolds and first high-AoA sample for extrapolation splits) rather than cherry-picked for low error.}

\subsection{Non-Watertight Mesh Training}
\label{sec:non-watertight}

A critical challenge in industrial CAE workflows is the preprocessing required to prepare CAD geometry for simulation. In many cases, \emph{defeaturing} (the manual removal or simplification of small geometric features) is the most time-consuming step in traditional simulation pipelines. Even after defeaturing, practitioners frequently encounter issues with non-watertight meshes: small gaps, overlapping faces, or topological inconsistencies that cause traditional CFD solvers to fail or require further manual intervention. For ML surrogates to achieve practical deployment in industry, they must be robust to such geometric imperfections rather than requiring the idealized meshes typical of academic benchmarks.

We expect \globe{} to tolerate boundary mesh imperfections because of its boundary-element-method-like formulation. Because the kernel evaluation treats each boundary face centroid as an independent source point, with influences aggregated via area-weighted summation (\cref{eq:aggregation-over-sources}), the architecture has no explicit dependence on topological connectivity, shared vertices, or edge relationships between faces. Unlike volume-mesh-based methods or graph neural networks that rely on explicit connectivity, \globe{} operates on a cloud of oriented surface elements. This suggests that small gaps, overlaps, or topological defects should degrade performance gracefully rather than catastrophically.

To test this hypothesis, we trained a model on the \textit{Full} split after corrupting each training airfoil's boundary mesh in two steps: (1) randomly deleting 50\% of the boundary faces, and (2) doubling the area of each remaining face by scaling its vertices radially outward from its centroid, preserving the total boundary area in \cref{eq:aggregation-over-sources}. The mesh is passed to the model as an unordered set of faces without connectivity.

This perturbation introduces both large gaps where faces were removed and self-intersections and disconnected edges where the remaining faces were expanded, yielding non-watertight meshes that standard mesh-based CFD solvers cannot accept without repair.

Despite this severe degradation of input quality, \globe{} produces reasonable predictions. \cref{tab:airfrans-decimated} compares performance on the official test cases across three configurations: the standard Full model, the decimated-mesh model, and the Scarce model (included as a data-scarcity reference point). While errors increase under decimation (as expected when discarding half the boundary information), the model remains functional, with errors of the same order as the Scarce model's.

\begin{table}[htb]
    \centering
    \caption{Robustness to non-watertight meshes: \globe{} trained on severely decimated boundary meshes (50\% of faces randomly removed, remaining faces doubled in area via vertex expansion, destroying topological connectivity) compared to baseline Full and Scarce models. Training uses decimated meshes; evaluation on the official test cases uses intact meshes. Metrics are case-normalized mean squared error, $E^{\mathrm{case}}$ (\cref{sec:airfrans-metric-definitions}), as in \cref{tab:airfrans-case-normalized}.}
    \label{tab:airfrans-decimated}
    \begin{tabular}{l c c c c}
        \toprule
                                            & \multicolumn{4}{c}{\textbf{Mean Squared Error}}                                                                                  \\
        \cmidrule(lr){2-5}
        {Model}                             & {$\bar{u}_x$}                                   & {$\bar{u}_y$}            & {$\bar{p}$}              & {$\bar{p}_s$}            \\
        \midrule
        \globe{} (Full)                     & 0.0047 \enegtwo                                 & 0.0039 \enegtwo          & 0.0031 \enegtwo          & 0.0039 \enegtwo          \\
        \textbf{\globe{} (Full, decimated)} & \textbf{0.0510} \enegtwo                        & \textbf{0.0402} \enegtwo & \textbf{0.0311} \enegtwo & \textbf{0.0336} \enegtwo \\
        \globe{} (Scarce)                   & 0.0208 \enegtwo                                 & 0.0180 \enegtwo          & 0.0286 \enegtwo          & 0.0392 \enegtwo          \\
        \bottomrule
    \end{tabular}
\end{table}

\begin{figure}[htb]
    \centering
    \includegraphics[width=\textwidth]{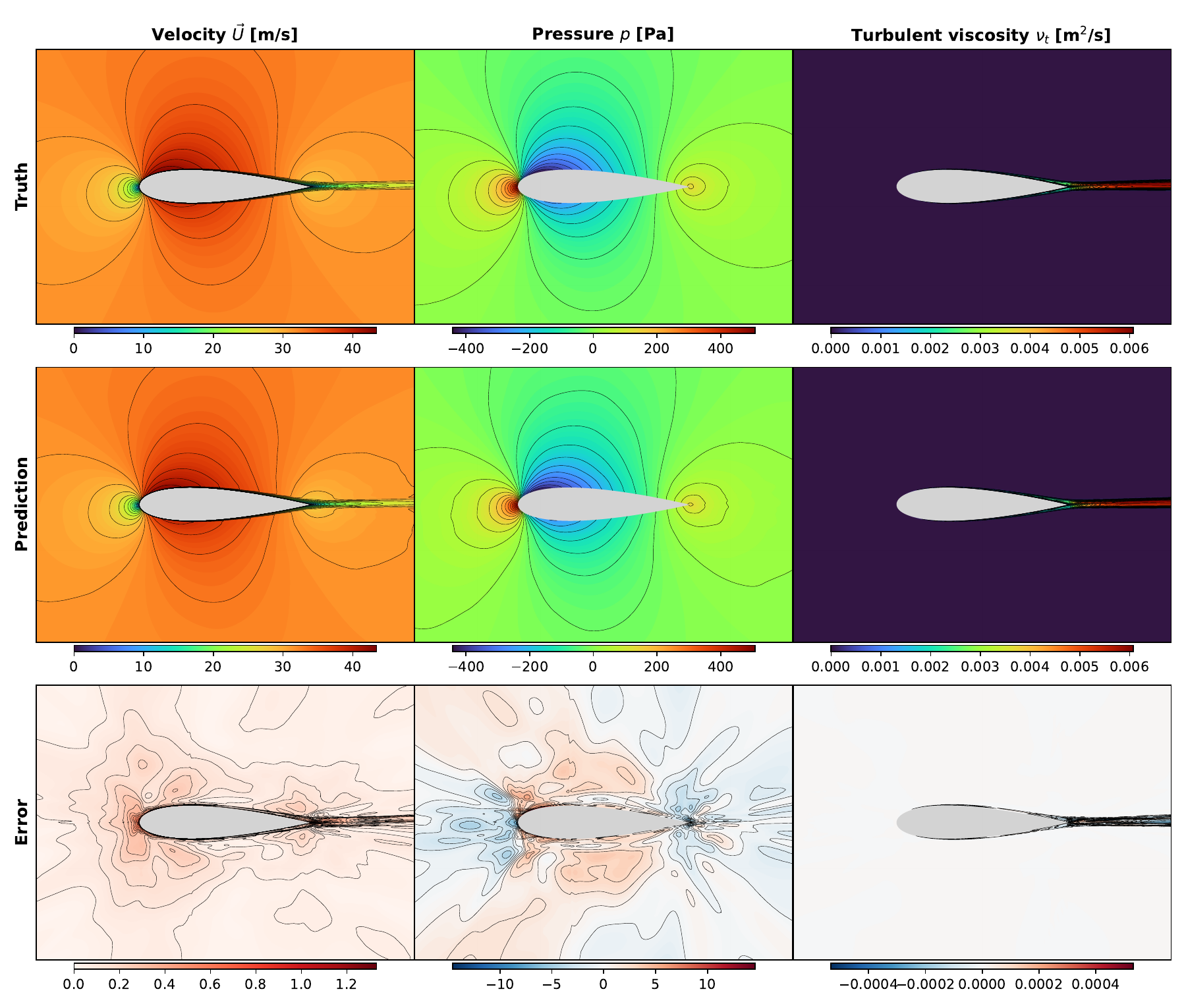}
    \caption{\globe{} model predictions on a test case when trained on severely decimated meshes (50\% boundary face removal). Rows show ground truth, \globe{} predictions, and error; columns show velocity magnitude, pressure, and turbulent viscosity in dimensional units. Truth and prediction share a color scale per field; error panels use their own scale, which differs between figures. Despite training exclusively on topologically invalid meshes with massive gaps and intersections, the model recovers the main flow features.}
    \label{fig:airfrans-decimated-qual}
\end{figure}

\cref{fig:airfrans-decimated-qual} shows a representative test case predicted by the decimated-mesh model. Even when trained exclusively on meshes with 50\% missing faces and severe topological defects, the model recovers coherent flow structures, sharp boundary-layer gradients, and physically plausible wakes. This suggests that \globe{} may tolerate minor geometric imperfections in CAD-derived meshes.

Beyond handling defeaturing artifacts, this decimation robustness also provides a near-term path to scalability for large-scale 3D datasets. Training on heavily downsampled boundary meshes does not catastrophically degrade accuracy, which is useful when scaling to 3D geometries with millions of boundary faces.

\section{AirFRANS Error-Metric Definitions}
\label{sec:airfrans-metric-definitions}

For field $j$ and $C$ test cases, the case-normalized score is
\begin{equation}
    \label{eq:metric-case}
    E^{\mathrm{case}}_j = \frac{1}{C}\sum_{c=1}^{C}\frac{\operatorname{MSE}_{c,j}}{s_{c,j}^{2}},
\end{equation}
where $\operatorname{MSE}_{c,j}$ is the pointwise mean squared error of field $j$ over the evaluation nodes of case $c$, and $s_{c,j}$ is the sample standard deviation of the ground-truth field over those nodes. The released AirFRANS evaluator instead normalizes targets by a training-global mean $\mu_{\mathrm{train},j}$ and population standard deviation $\sigma_{\mathrm{train},j}$, pooled over all points of the training statistic subset; the mean cancels from prediction errors, giving
\begin{equation}
    \label{eq:metric-ref}
    E^{\mathrm{ref}}_j = \frac{1}{C}\sum_{c=1}^{C}\frac{\operatorname{MSE}_{c,j}}{\sigma_{\mathrm{train},j}^{2}}.
\end{equation}
For surface pressure, the numerator is restricted to surface nodes, while the denominator remains the training-set volume-pressure variance. Both evaluators use the official test cases, all original volume-mesh nodes, the same surface masks, and equal weighting of cases; the denominator is the only material difference. $E^{\mathrm{case}}$ is invariant to any case-dependent rescaling of the fields, since the scale cancels between numerator and denominator, whereas $E^{\mathrm{ref}}$ is not; we therefore computed $E^{\mathrm{ref}}$ directly in the released evaluator's dimensional representation, from the same fixed predictions. The \globe{} rows of \cref{tab:airfrans-full,tab:airfrans-scarce,tab:airfrans-reynolds,tab:airfrans-aoa} report $E^{\mathrm{ref}}$. Earlier versions of this work reported $E^{\mathrm{case}}$; \cref{tab:airfrans-case-normalized} lists those values for the same models, and \cref{tab:airfrans-decimated} also uses $E^{\mathrm{case}}$. Published baselines use the released evaluator's denominator, so only $E^{\mathrm{ref}}$ is compared with them: under $E^{\mathrm{case}}$, for example, \globe{}'s AoA volume-pressure error appears close to the best baseline's, whereas under $E^{\mathrm{ref}}$ it is roughly $31\times$ lower. MARIO and enf2enf also use training-global output normalization in their released pipelines, but exact statistic subsets and other preprocessing choices are not controlled across repositories, so this aligns the error denominator only.

\begin{table}[htb]
    \centering
    \caption{Case-normalized scores $E^{\mathrm{case}}$ (\cref{eq:metric-case}) of the \globe{} models in \cref{tab:airfrans-full,tab:airfrans-scarce,tab:airfrans-reynolds,tab:airfrans-aoa}; lower is better.}
    \label{tab:airfrans-case-normalized}
    \begin{tabular}{l c c c c}
        \toprule
        {Task}   & {$\bar{u}_x$}   & {$\bar{u}_y$}   & {$\bar{p}$}      & {$\bar{p}_s$}    \\
        \midrule
        Full     & 0.0047 \enegtwo & 0.0039 \enegtwo & 0.0031 \enegtwo  & 0.0039 \enegtwo  \\
        Scarce   & 0.0208 \enegtwo & 0.0180 \enegtwo & 0.0286 \enegtwo  & 0.0392 \enegtwo  \\
        Reynolds & 0.1897 \enegtwo & 0.0811 \enegtwo & 0.2446 \enegtwo  & 0.2903 \enegtwo  \\
        AoA      & 0.9706 \enegtwo & 1.1928 \enegtwo & 11.3786 \enegtwo & 21.8215 \enegtwo \\
        \bottomrule
    \end{tabular}
\end{table}

\section{Physically Nondimensionalized Performance Metrics}
\label{sec:physical-metrics}

As discussed in \cref{sec:results-quantitative}, the primary results in \crefrange{tab:airfrans-full}{tab:airfrans-scarce} normalize errors by training-set field variances (\cref{sec:airfrans-metric-definitions}) to enable comparison with values reported by prior work on AirFRANS. However, this approach obscures physical meaning. Here we report performance using proper physical nondimensionalization, where velocity fields are normalized by the freestream velocity magnitude $|\bm{U}_\infty|$, pressure by the dynamic pressure $\frac{1}{2}\rho|\bm{U}_\infty|^2$ (yielding the pressure coefficient $C_p$), and turbulent viscosity by the kinematic viscosity $\nu$.

\cref{tab:airfrans-physical-metrics} presents mean squared error (MSE) and mean absolute error (MAE) on the official test cases of both the Full and Scarce splits using these physically meaningful nondimensional quantities. Unlike \crefrange{tab:airfrans-full}{tab:airfrans-scarce}, values are reported without the $\times10^{-2}$ multiplier (which was used in the main tables only for consistency with AirFRANS reporting style). We report MAE as the primary metric: it provides direct physical interpretability in the base units of each field. For example, an MAE of $2.1\times10^{-3}$ on $C_p$ indicates that typical errors are $\pm0.0021$ in pressure coefficient -- immediately meaningful to practitioners. In contrast, MSE values require taking the square root for physical interpretation, and overweight outliers in ways that may not align with engineering significance.

\begin{table}[htb]
    \centering
    \caption{\globe{} model performance on the official AirFRANS test cases, reported using physical nondimensionalization. All fields are nondimensionalized by physical characteristic scales ($|\bm{U}_\infty|$ for velocity, $\frac{1}{2}\rho|\bm{U}_\infty|^2$ for pressure). These metrics provide physically interpretable performance assessment, in contrast to the variance-normalized metrics in \crefrange{tab:airfrans-full}{tab:airfrans-scarce}. The shear-stress ground truth is derived from velocity gradients and wall normals, so it depends on the gradient reconstruction; the $C_{F,\text{shear}}$ values use the public implementation's preprocessing.}
    \label{tab:airfrans-physical-metrics}
    \small
    \begin{tabular}{l c c c c}
        \toprule
                                & \multicolumn{2}{c}{\textbf{Full Dataset}} & \multicolumn{2}{c}{\textbf{Scarce Dataset}}                                             \\
        \cmidrule(lr){2-3} \cmidrule(lr){4-5}
        \textbf{Field}          & \textbf{MSE}                              & \textbf{MAE}                                & \textbf{MSE}        & \textbf{MAE}        \\
        \midrule
        \multicolumn{5}{l}{\textit{Nondimensional velocity}}                                                                                                          \\
        $U_x/|\bm{U}_\infty|$   & $9.53\times10^{-6}$                       & $1.57\times10^{-3}$                         & $4.53\times10^{-5}$ & $2.38\times10^{-3}$ \\
        $U_y/|\bm{U}_\infty|$   & $5.45\times10^{-6}$                       & $1.14\times10^{-3}$                         & $4.62\times10^{-5}$ & $2.11\times10^{-3}$ \\
        \midrule
        \multicolumn{5}{l}{\textit{Pressure coefficient}}                                                                                                             \\
        $C_p$                   & $3.49\times10^{-5}$                       & $2.07\times10^{-3}$                         & $1.30\times10^{-3}$ & $5.02\times10^{-3}$ \\
        $C_p$ (surface only)    & $1.06\times10^{-4}$                       & $3.71\times10^{-3}$                         & $6.04\times10^{-3}$ & $1.09\times10^{-2}$ \\
        $C_{pt}$                & $2.59\times10^{-5}$                       & $1.73\times10^{-3}$                         & $5.19\times10^{-4}$ & $3.41\times10^{-3}$ \\
        $C_{pt}$ (surface only) & $1.54\times10^{-4}$                       & $5.08\times10^{-3}$                         & $8.32\times10^{-3}$ & $1.47\times10^{-2}$ \\
        \midrule
        \multicolumn{5}{l}{\textit{Turbulent viscosity ratio}}                                                                                                        \\
        $\ln(1+\nu_t/\nu)$      & $1.97\times10^{-3}$                       & $8.30\times10^{-3}$                         & $6.24\times10^{-3}$ & $1.40\times10^{-2}$ \\
        \midrule
        \multicolumn{5}{l}{\textit{Skin friction coefficient (surface only)} }                                                                                        \\
        $C_{F,\text{shear},x}$  & $1.07\times10^{-8}$                       & $3.78\times10^{-5}$                         & $5.43\times10^{-8}$ & $7.17\times10^{-5}$ \\
        $C_{F,\text{shear},y}$  & $1.86\times10^{-8}$                       & $4.84\times10^{-5}$                         & $1.40\times10^{-7}$ & $9.44\times10^{-5}$ \\
        \bottomrule
    \end{tabular}
\end{table}

\paragraph{Interpretation.} These physically nondimensionalized metrics reveal \globe{}'s strong performance in aerodynamically meaningful terms. On the Full split, velocity MAE is $\sim\!1.4\times10^{-3}$ (0.14\% of freestream), pressure coefficient MAE is $2.1\times10^{-3}$ in the volume and $3.7\times10^{-3}$ on the surface, and skin friction MAE is $\sim\!4.3\times10^{-5}$. The Scarce split shows graceful degradation with roughly 2--3$\times$ higher MAE errors despite 4$\times$ less training data. These pointwise errors do not by themselves establish the accuracy of integrated design quantities such as lift, drag, or pitching moment, which we did not evaluate on AirFRANS.

In contrast, the variance-normalized metrics in \crefrange{tab:airfrans-full}{tab:airfrans-scarce} obscure this physical interpretation: without knowing dataset statistics, one cannot infer whether a given error magnitude corresponds to 0.1\%, 1\%, or 10\% physical error. This demonstrates why physical nondimensionalization should be standard practice for ML surrogates in engineering domains, and is indeed longstanding practice in the fluid dynamics literature.

\section{DrivAerML: 3D Car Aerodynamics}
\label{sec:drivaer-appendix}

\paragraph{Data.} DrivAerML \citep{ashtonDrivAerMLHighFidelityComputational2024} is a public dataset of \num{500} scale-resolving CFD simulations of parametrically varied DrivAer cars in a wind tunnel, with ${\sim}\SI{8.8}{M}$ surface cells and ${\sim}\SI{150}{M}$ volume cells per car. We train on \num{400} cars and report errors on \num{34} held-out validation cars. The car body and wheels are stationary no-slip walls; the floor is slip upstream of $x = \SI{-2.34}{m}$ and no-slip downstream.

\paragraph{Models.} Separate surface and volume \globe{} models share one architecture: the car, the slip floor patch, and the no-slip floor patch are separate boundary partitions; the kernel networks have $[512, 512, 512]$ hidden layers, with 16 latent scalars and 8 latent vectors per hyperlayer and 4 spherical-harmonic terms; there are 2 communication hyperlayers; and the dual-tree Barnes-Hut acceleration (\cref{sec:hierarchical-acceleration-appendix}) runs at $\theta = 1.5$ with target-side expansion enabled. Each model has 12.2M parameters. The surface model predicts pressure and wall-shear stress at surface cells, with an area-weighted loss; the volume model predicts velocity, pressure, and turbulent viscosity at volume cells, with loss weights 1.25, 0.67, and 0.1. At each training step, both models take a random sample of \num{50000} car-surface faces plus the two floor patches as boundary input and evaluate \num{50000} sampled prediction points per car; at inference, the \num{50000}-face boundary sample is drawn once per car with a fixed seed while all surface or volume cells are queried in \num{50000}-point chunks. Training uses a Huber loss, Muon for matrix-shaped parameters and AdamW for the rest, a learning rate of $3\times10^{-3}$ that decays by $10\times$ every 200 epochs, and weight decay $10^{-4}$, in bfloat16 with \texttt{torch.compile}. Each model trains on one node of 4 GB300 GPUs for about 49\,h (\SI{3.84}{\second} per step, median), until the learning rate reaches its floor after about 430 epochs; the checkpoint with the lowest validation loss is kept.

\paragraph{Baseline.} Transolver \citep{wuTransolverFastTransformer2024} is trained on the same data, as separate surface and volume models with the same checkpoint selection, using 16 layers of width 512 (47.2M parameters per model), \num{100000} points per step in float32, and up to 500 epochs.

\paragraph{Metrics.} For each validation car, we compute the relative $L_2$ error of surface pressure and wall-shear stress over all surface cells, weighted by cell area, and of velocity and pressure over all volume cells, weighted equally, then average over the 34 cars. The drag coefficient is integrated from each car's predicted surface pressure and wall-shear stress, and $R^2$ is computed across the 34 cars.

\begin{table}[htb]
    \centering
    \caption{DrivAerML validation errors over 34 held-out cars: mean relative $L_2$ error (\%) of surface pressure $p_s$, wall-shear stress $\bm{\tau}_w$, volume velocity $\bm{U}$, and volume pressure $p_v$ (lower is better), and $R^2$ of the integrated drag coefficient $C_D$ across cars (higher is better). Ensembles average the predictions of three seeds. Checkpoints are selected on the same cars.}
    \label{tab:drivaer}
    \begin{tabular}{l c c c c c}
        \toprule
        Model & $p_s$ & $\bm{\tau}_w$ & $\bm{U}$ & $p_v$ & $R^2(C_D)$ \\
        \midrule
        \globe{}, one seed            & 8.87 & 9.67 & 8.29 & 6.74 & 0.976 \\
        \globe{}, 3-seed ensemble     & 7.67 & 8.67 & 7.36 & 5.89 & 0.959 \\
        Transolver, one seed          & 4.11 & 6.15 & 5.26 & 3.81 & 0.987 \\
        Transolver, 3-seed ensemble   & 3.88 & 5.97 & 4.97 & 3.57 & 0.989 \\
        \bottomrule
    \end{tabular}
\end{table}

\begin{figure}[htb]
    \centering
    \includegraphics[width=\textwidth]{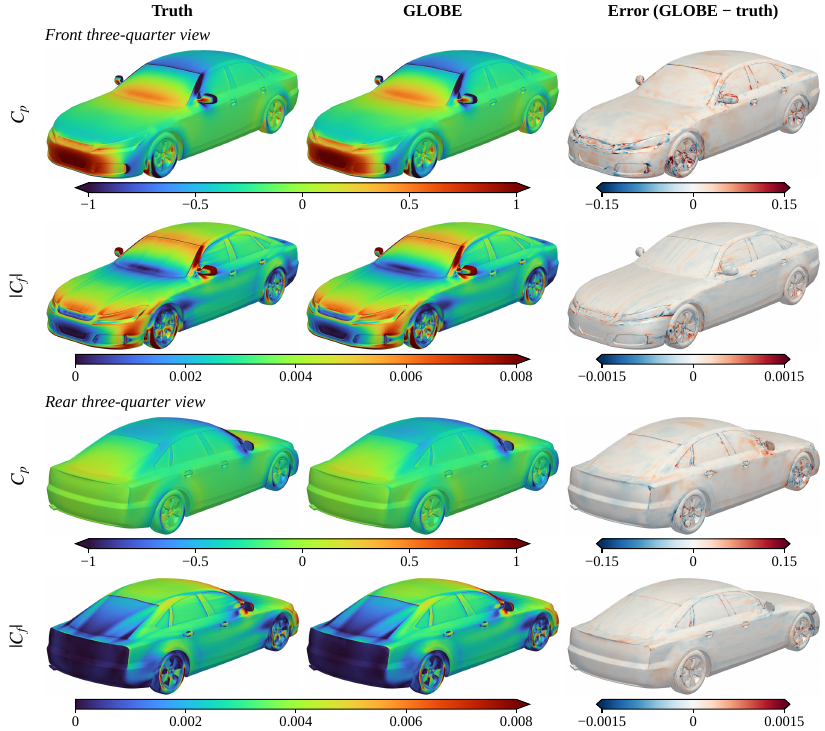}
    \caption{The predictions of \cref{fig:drivaer} (same car, model, and color ranges) from front and rear three-quarter views.}
    \label{fig:drivaer-views}
\end{figure}

\paragraph{Results and cost.} \Cref{tab:drivaer} shows that \globe{}'s field errors are 1.5--2.2$\times$ those of Transolver, while both predict the drag coefficient with $R^2 > 0.95$. On one GB300 GPU, \globe{} predicts all surface cells of a car in about \SI{165}{\second} and all volume cells in about \SI{2750}{\second}; the Transolver volume model takes about \SI{475}{\second} with the same prediction code. \Cref{fig:drivaer-views} shows the surface predictions of \cref{fig:drivaer} from front and rear three-quarter views.

\section{Ablation Studies}
\label{sec:ablation-appendix}

This appendix reports two single-component ablation campaigns on the AirFRANS \emph{Scarce} split (the fastest-to-train regime, with $200$ training samples). The reported metric is the per-epoch \emph{total validation loss} -- the sum of the five scaled component losses optimized during training: $\Delta U / \norm{\bm{U}_\infty}$, $C_p$, $C_{p_t}$, $\ln(1 + \nu_t / \nu)$, and $C_{F,\text{shear}}$, with scale factors $1.0$, $1.0$, $1.0$, $5.0$, and $0.01$ respectively. Lower is better. Both campaigns were run with the pre-release implementation of \globe{} (\cref{sec:hyperparameters}). The two campaigns were run several weeks apart and are therefore each anchored to their own internal reference row rather than a shared global baseline; within each campaign, all rows share identical hardware, optimizer, and training-budget settings.

\subsection{Pad\'{e} Kernel vs.\ MLP Kernel}
\label{sec:ablation-pade}

\cref{tab:architectural-ablation} contrasts the reference Pad\'{e}-approximant kernel network ($N = D = 2$) with a plain MLP of equivalent width and depth. Among the substitutions we tested, this produces the largest degradation: in a single run, the standard-MLP variant trains unstably and fails to descend below an order-of-magnitude-higher loss plateau, inflating total validation loss by roughly $90\times$. Because the Pad\'{e}-approximant block contains two independent MLPs ($\phi_n$ and $\phi_d$), it carries ${\sim}2\times$ the trainable parameters of the standard-MLP comparison at this width and depth. This ablation therefore shows that replacing the Pad\'{e} network with a plain MLP of the same width and depth severely degrades training, though it does not isolate the mechanism (rational approximation, parameter count, conditioning, or optimization); the algebraic far-field decay is imposed by the envelope of \cref{sec:far-field-decay-appendix} in both variants. Retraining an MLP-kernel \globe{} with stabilization (e.g., a lower learning rate or gradient clipping) would be a useful additional study.

\begin{table}[htb]
    \centering
    \caption{Pad\'{e} vs.\ MLP kernel network on the AirFRANS \emph{Scarce} split. Lower total validation loss is better.}
    \label{tab:architectural-ablation}
    \begin{tabular}{l c}
        \toprule
        Configuration                       & Total validation loss \\
        \midrule
        Reference \globe{} (Pad\'{e} kernel) & $0.00159$             \\
        MLP kernel (no Pad\'{e})            & $0.14144$             \\
        \bottomrule
    \end{tabular}
\end{table}

\subsection{Hierarchical Acceleration: Opening Angle $\theta$ and Far-Field Target Expansion}
\label{sec:ablation-bh}

\cref{tab:bh-ablation} characterizes the accuracy of the dual-tree Barnes-Hut scheme as a function of the opening angle $\theta$ and the \texttt{expand\_far\_targets} flag. With \texttt{expand\_far\_targets~=~False}, both source and target sides of a far-field interaction are monopole-approximated; with \texttt{expand\_far\_targets~=~True}, the target side is evaluated at individual points while the source side retains the monopole approximation (see \cref{sec:hierarchical-acceleration-appendix} for definitions).

\begin{table}[htb]
    \centering
    \caption{Hierarchical-acceleration ablation on AirFRANS-scarce: opening angle $\theta$ and the \texttt{expand\_far\_targets} flag. The dense reference ($\theta = 0$) recovers exact evaluation. Lower total validation loss is better.}
    \label{tab:bh-ablation}
    \begin{tabular}{l l c}
        \toprule
        $\theta$              & \texttt{expand\_far\_targets} & Total validation loss \\
        \midrule
        $0$ (dense reference) & ---                           & $0.00162$             \\
        $0.5$                 & False                         & $0.01310$             \\
        $1.0$                 & False                         & $0.01964$             \\
        $0.5$                 & True                          & $0.00365$             \\
        $1.0$                 & True                          & $0.00311$             \\
        $2.0$                 & True                          & $0.00496$             \\
        \bottomrule
    \end{tabular}
\end{table}

With \texttt{expand\_far\_targets~=~False}, even moderate opening angles increase validation loss by ${\sim}8\times$ at $\theta = 0.5$ and ${\sim}12\times$ at $\theta = 1.0$, indicating that approximating both source \emph{and} target clusters by their centroids is too coarse on AirFRANS-scale meshes. With \texttt{expand\_far\_targets~=~True}, by contrast, total validation loss remains within $2$--$3\times$ of the dense reference across $\theta \in \{0.5, 1.0, 2.0\}$ and varies little across that range; the small difference between $\theta = 0.5$ ($0.00365$) and $\theta = 1.0$ ($0.00311$) is within single-seed run-to-run variation.

Target-side expansion (\texttt{expand\_far\_targets~=~True}) is also used for the DrivAerML models in \cref{sec:drivaer-appendix}. Source-side monopole approximation remains the dominant remaining error mechanism; lifting it to a higher-order multipole expansion (e.g., dipole, quadrupole) is left as future work.

\section{Expanded Limitations}
\label{sec:limitations-appendix}

This appendix expands on the limitations summarized in the main-paper Conclusions (\cref{sec:conclusions}).

\paragraph{Volumetric source terms and spatially varying coefficients.} The \globe{} architecture represents PDE solutions as superpositions of kernel evaluations from \emph{boundary} faces, deliberately matching boundary-driven elliptic PDE structure (\cref{sec:necessary-attributes-appendix}). PDEs governed by interior source terms, spatially varying material coefficients, or distributed body forces are not captured by a boundary-only representation. In principle the architecture could be extended to handle such cases by introducing volumetric source ``faces'' as an additional BC partition with non-trivial 3D source data, but the current formulation does not do this and we have not evaluated it.

\paragraph{Strongly nonlinear regimes.} \globe{} is nonlinear: its learned source representations, communication hyperlayers, and kernels are nonlinear functions of the geometry, boundary conditions, and global parameters, and only the final source aggregation is additive. Its inductive bias is nonetheless drawn from linear boundary-element structure, which suits flows where much of the domain is well-approximated by superpositions of localized influences (e.g., vorticity-free regions of subsonic external flow, well-approximated by linear potential flow, and weakly coupled boundary layers). We expect it to be less well-suited to regimes dominated by strongly nonlinear interactions: massive separation, deep stall, transonic shock-boundary-layer interactions, etc.

\paragraph{Rigidity of the far-field decay envelope.} The Lamb-Oseen-style envelope (\cref{eq:far-field-decay-full}) hard-codes algebraic $\sim 1/r^{\nsp - 1}$ decay, which matches the velocity fields induced by single-layer sources (equivalently, the leading-order behavior of double-layer/dipole potentials) for the elliptic PDEs we target (Laplace, Stokes, screened Helmholtz, etc.). While this is required for the Barnes-Hut far-field assumption to hold, it may be overly rigid for some PDEs -- e.g., logarithmic 2D Laplace potentials or strongly anisotropic decays. We leave exploration of alternatives for future work.

\paragraph{Distribution shifts in flow topology.} Even for mostly attached flows, large changes in global flow topology can move test cases far from the training distribution, which accounts for the higher errors on AoA extrapolation in \cref{tab:airfrans-aoa} (\cref{sec:results-quantitative}).

\newpage
\input{checklist.tex}

\end{document}

%% file: commands.tex
\newcommand{\mat}[1]{
    \mathbf{#1}
}

\newcommand{\Rey}{\mathrm{Re}}

\usepackage{xspace}

%% file: globe_macros.tex
\usepackage{bm}

\newcommand{\vect}[1]{\bm{#1}}
\renewcommand{\mat}[1]{\bm{#1}}
\newcommand{\norm}[1]{\left\lVert #1 \right\rVert}

\newcommand{\RR}{\mathbb{R}}
\newcommand{\nsp}{d}             %
\newcommand{\ntrg}{N_t}          %
\newcommand{\nsrc}{N_s}          %

\newcommand{\td}{\textsc{TensorDict}}
\newcommand{\kernel}{\mathcal{K}}
\newcommand{\smoothlog}{\operatorname{smoothlog}}
\newcommand{\Leg}{\operatorname{P}} %
\newcommand{\pade}{\text{Padé}}
\newcommand{\globe}{\text{GLOBE}}
\newcommand{\enegtwo}{{\color{gray}\ensuremath{\times 10^{-2}}}}

%% file: preamble.tex
\usepackage[T1]{fontenc}

\usepackage{microtype}
\usepackage{amsmath}

\usepackage{amssymb}
\usepackage{bm}
\usepackage{textcomp, gensymb}
\usepackage{graphicx}
\usepackage{pgf}
\usepackage{float}

\usepackage{ifdraft}
\usepackage{hyperref}
\usepackage[capitalize,noabbrev]{cleveref}
\usepackage{enumitem}

\usepackage{eqexpl}
\eqexplSetIntro{where:}
\eqexplSetDelim{=}

\usepackage{multicol}

\usepackage{siunitx}
\usepackage{tabularx}
\usepackage{booktabs}
\usepackage{multirow}
\usepackage{tablefootnote}
\usepackage{tabularray}
\UseTblrLibrary{booktabs}
\UseTblrLibrary{counter}

\usepackage{tikz}
\usetikzlibrary{positioning}
\usetikzlibrary{shapes.geometric}
\usetikzlibrary{shapes.arrows}
\usetikzlibrary{decorations.pathmorphing, decorations.pathreplacing, calc}
\usetikzlibrary{fit}
\usetikzlibrary{backgrounds}

\usepackage{tikz-cd}
\tikzcdset{
    math mode=false
}

\usepackage[dvipsnames,svgnames]{xcolor}

\definecolor{myorange}{RGB}{255, 165, 0}
\definecolor{mydarkseagreen}{RGB}{143, 188, 143}
\definecolor{mydodgerblue}{RGB}{30, 144, 255}

\usepackage{titlesec}

\usepackage{caption}
\usepackage{adjustbox}

\usepackage[titletoc,title]{appendix}

\usepackage{svg}

\usepackage{subcaption}

\usepackage{afterpage}
\usepackage[section]{placeins}

\usepackage{nth}

\usepackage{xstring}

\usepackage{authblk}

\usepackage{todonotes}
\setuptodonotes{inline,backgroundcolor=yellow!20,prepend,caption={TODO}}

%% file: figures/globe_arch.tex
\begin{tikzpicture}[
    x=6.5in,
    y=1cm,
    box/.style={
        draw, rounded corners,
        inner xsep=3pt, inner ysep=3pt, outer sep=0pt,
        minimum height=0.6cm,
        font=\small,
        align=flush center,
        anchor=center,
    },
    smallbox/.style={
        box,
        minimum height=0.5cm,
        font=\footnotesize
    },
    arrow/.style={->, >=latex, line width=0.8pt},
    thickarrow/.style={->, >=latex, line width=1.2pt},
    blowup/.style={->, >=latex, line width=0.6pt, dashed, gray},
    zoomline/.style={gray, line width=0.4pt, shorten >=4pt},
    label/.style={font=\footnotesize, align=center}
]

\pgfmathsetmacro{\fullwidth}{6.5in}
\pgfmathsetmacro{\levelTopY}{10.5}
\pgfmathsetmacro{\levelMiddleY}{6.2}
\pgfmathsetmacro{\levelBottomY}{1.4}

\node[label, anchor=south] at (0.5, \levelTopY+1.5) {\textbf{Top Level: Communication Hyperlayers} (Section \ref{sec:hyperlayers})};

\node[box, fill=Peach!15, text width=12ex, anchor=west] (bc_input) at (0.0, \levelTopY) {
    \textbf{Inputs}\\
    \scriptsize
    \begin{flushleft}
    Boundary meshes;\\
    Reference lengths;\\
    Global scalars \& vectors\\
    Query points
    \end{flushleft}
};

\node[smallbox, fill=Peach!15, text width=8.2ex] (init_latent) at (0.19, \levelTopY) {
    \scriptsize
    Initialize
    \begin{flushleft}
    $w_s^{(k,0)} = 1$\\
    latents $= \emptyset$
    \end{flushleft}
};

\node[box, fill=mydarkseagreen!15, text width=13ex] (hyper0) at (0.325, \levelTopY) {
    \textbf{Hyperlayer 0}\\
    \footnotesize
    $\sum_i$ Multiscale kernels, with:\\
    \scriptsize
    \begin{flushleft}
    \textbf{Sources}: $\text{BC}_i$ faces \\
    \textbf{Targets}: all faces\\
    \textbf{Outputs}: $w_s^{(k)}$ for $H_1$ kernels; latent scalars \& vectors
    \end{flushleft}
};

\node[box, fill=mydarkseagreen!15, text width=13ex] (hyper1) at (0.50, \levelTopY) {
    \textbf{Hyperlayer 1}\\
    \footnotesize
    $\sum_i$ Multiscale kernels, with:\\
    \scriptsize
    \begin{flushleft}
    \textbf{Sources}: $\text{BC}_i$ faces \\
    \textbf{Targets}: all faces\\
    \textbf{Outputs}: $w_s^{(k)}$ for $H_2$ kernels; latent scalars \& vectors
    \end{flushleft}
};

\node[label, outer sep=-2pt] (ellipsis) at (0.6045, \levelTopY) {$\cdots$};

\node[box, fill=mydarkseagreen!15, text width=13ex] (hyperH) at (0.710, \levelTopY) {
    \textbf{Hyperlayer $H$}\\
    \footnotesize
    $\sum_i$ Multiscale kernels, with:\\
    \scriptsize
    \begin{flushleft}
    \textbf{Sources}: $\text{BC}_i$ faces \\
    \textbf{Targets}: query pts.\\
    \textbf{Outputs}: raw solution fields at query points
    \end{flushleft}
};

\node[box, fill=RoyalPurple!15, text width=10ex] (calib) at (0.852, \levelTopY) {
    \textbf{Calibration}\\
    \scriptsize 
    \begin{flushleft}
    Scalars: affine;\\
    Vectors: linear
    \end{flushleft}
};

\node[box, fill=RoyalPurple!15, text width=6.5ex, anchor=east] (output_fields) at (1.0, \levelTopY) {
    \textbf{Output}\\
    \scriptsize
    \begin{flushleft}
    Solution fields at query pts.
    \end{flushleft}
};

\draw[arrow] (bc_input.east) -- (init_latent.west);
\draw[arrow] (init_latent.east) -- (hyper0.west);
\draw[arrow] (hyper0.east) -- node[above, font=\tiny, text width=5ex, align=flush center] {latent data} (hyper1.west);
\draw[arrow] (hyper1.east) -- (ellipsis.west);
\draw[arrow] (ellipsis.east) -- (hyperH.west);
\draw[arrow] (hyperH.east) -- (calib.west);
\draw[arrow] (calib.east) -- (output_fields.west);

\node[box, fill=mydarkseagreen!5, text width=\fullwidth, minimum height=4.2cm] (ms_kernel) at (0.5, \levelMiddleY) {};
\draw[zoomline] (hyper1.south west) -- (ms_kernel.north west);
\draw[zoomline] (hyper1.south east) -- (ms_kernel.north east);
\path (hyper1.south) -- node[align=center, font=\scriptsize, text=gray] {Zooming into a single multiscale kernel\\associated with a single source BC type} (ms_kernel.north);

\node[label, anchor=south] (levelMiddleLabel) at (0.5, \levelMiddleY+1.5) {\textbf{Middle Level: Multiscale Kernel} (one per source BC type, e.g., `inlet', `outlet', `wall'; Section \ref{sec:multiscale})};

\node[smallbox, fill=Peach!15, text width=23ex, anchor=west] (ms_input) at (0.0, \levelMiddleY) {
    \textbf{Inputs}\\
    \scriptsize
    \begin{flushleft}
    Source points $\{\vect{x}_s\}$;\\
    Target points $\{\vect{x}_t\}$;\\
    Source scalars \& vectors;\\
    Global scalars \& vectors;\\
    Source strengths $w_s^{(k)}$ (per branch);\\
    Log-ratios $\ln(\ell_i/\ell_j)$
    \end{flushleft}
};

\node[box, fill=TealBlue!15, text width=22ex] (k1) at (0.45, \levelMiddleY+1.0) {
    \textbf{Kernel} $\kernel^{(1)}$ with ref. length $\ell^\text{eff}_1 = \ell_1\,\exp(\alpha_1)$
};

\node[box, fill=TealBlue!15, text width=22ex] (k2) at (0.45, \levelMiddleY) {
    \textbf{Kernel} $\kernel^{(2)}$ with ref. length $\ell^\text{eff}_2 = \ell_2\,\exp(\alpha_2)$
};

\node[label] (ms_ellipsis) at (0.45, \levelMiddleY-0.6) {$\vdots$};
\node[label, anchor=west] (ms_ellipsis) at (0.45, \levelMiddleY-0.7) {\scriptsize ($n$ branches)};

\node[box, fill=TealBlue!15, text width=22ex] (kn) at (0.45, \levelMiddleY-1.4) {
    \textbf{Kernel} $\kernel^{(n)}$ with ref. length $\ell^\text{eff}_n = \ell_n\,\exp(\alpha_n)$
};

\node[box, fill=TealBlue!15] (ms_sum) at (0.70, \levelMiddleY) {
    $\displaystyle\sum_{k}$
};

\node[smallbox, fill=RoyalPurple!15, text width=19ex, anchor=east] (ms_output) at (1.0, \levelMiddleY) {
    \textbf{Output}\\
    \scriptsize
    \begin{flushleft}
    Field values at targets\\
    $\vect{u}_t$ (Eq.~\ref{eq:multiscale-sum-appendix})
    \end{flushleft}
};

\draw[arrow] (ms_input.east) to[out=0, in=180] (k1.west);
\draw[arrow] (ms_input.east) to[out=0, in=180] (k2.west);
\draw[arrow] (ms_input.east) to[out=0, in=180] (kn.west);
\draw[arrow] (k1.east) to[out=0, in=180] (ms_sum.west);
\draw[arrow] (k2.east) to[out=0, in=180] (ms_sum.west);
\draw[arrow] (kn.east) to[out=0, in=180] (ms_sum.west);
\draw[arrow] (ms_sum.east) -- (ms_output.west);

\node[label, anchor=north, text width=40ex] at (0.15, \levelMiddleY-1.25) {
    \scriptsize Each kernel function has its own learnable parameters; shared architecture (Sec.~\ref{sec:multiscale})
};

\node[box, fill=TealBlue!5, text width=\fullwidth, minimum height=4.5cm] (kernel_detail) at (0.5, \levelBottomY-0.25) {};
\draw[zoomline] (kn.south west) -- (kernel_detail.north west);
\draw[zoomline] (kn.south east) -- (kernel_detail.north east);
\path (kn.south) -- node[align=center, font=\scriptsize, text=gray] {Zooming into a single kernel $\kernel^{(k)}$} (kernel_detail.north);

\node[label, anchor=south] (levelBottomLabel) at (0.5, \levelBottomY+1.35) {\textbf{Bottom Level: Kernel Function $\kernel$} (source $\to$ target evaluation; Section \ref{sec:kernel-functions})};

\node[smallbox, fill=Peach!15, text width=12ex, anchor=west] (k_input) at (0.0, \levelBottomY) {
    \textbf{Input Assembly}\\
    \scriptsize
    \begin{flushleft}
    Relative position\\
    $\vect{r}_{ts} = \frac{\vect{x}_t - \vect{x}_s}{\ell}$;\\
    Source scalars $\vect{s}$;\\
    Source vectors $\mat{V}$;\\
    Global scalars \& vectors
    \end{flushleft}
};

\node[smallbox, fill=Peach!15, text width=15ex] (k_features) at (0.225, \levelBottomY) {
    \textbf{Invariant Features}\\
    \scriptsize
    \begin{flushleft}
    Scalars $\vect{s}$;\\
    $\smoothlog(\norm{\vect{v}_i})$ for vectors;\\
    Weighted $\Leg_i(\cos\theta_{ab})$ on vector pairs
    \end{flushleft}
};

\node[box, fill=ProcessBlue!15, text width=10ex] (k_network) at (0.375, \levelBottomY) {
    \textbf{\pade{} MLP}\\
    \scriptsize
    \begin{flushleft}
    Process concat. scalar features
    \end{flushleft}
};

\node[smallbox, fill=ProcessBlue!15, text width=14ex] (k_decay) at (0.52, \levelBottomY) {
    \textbf{Far-Field Decay}\\
    \scriptsize
    \begin{flushleft}
    Scale by envelope (Eq.~\ref{eq:far-field-decay-full}):
    \end{flushleft}
    $\frac{1-\exp(-\norm{\vect{r}}^2)}{(\norm{\vect{r}}^2+1)^{(\nsp-1)/2}}$
};

\node[smallbox, fill=ProcessBlue!15, text width=12ex] (k_reproj) at (0.675, \levelBottomY) {
    \textbf{Vector Reprojection}\\
    \scriptsize
    \begin{flushleft}
    Project onto basis\\
    $\mat{B} = [\vect{\hat{r}}, \vect{\hat{v}}_i, \ldots]$
    \end{flushleft}
};

\node[box, fill=ProcessBlue!15, text width=11ex] (k_agg) at (0.82, \levelBottomY) {
    \textbf{Aggregation}\\
    \scriptsize
    \begin{flushleft}
    $\sum_s w_s\, a_s\, \kernel_{ts}$\\(Eq.~\ref{eq:aggregation-over-sources})
    \end{flushleft}
};

\node[smallbox, fill=RoyalPurple!15, text width=9ex, anchor=east] (k_output) at (1.0, \levelBottomY) {
    \textbf{Output}\\
    \scriptsize
    \begin{flushleft}
    Aggregated kernel\\
    output $\kernel_t$
    \end{flushleft}
};

\draw[arrow] (k_input.east) -- (k_features.west);
\draw[arrow] (k_features.east) -- (k_network.west);
\draw[arrow] (k_network.east) -- (k_decay.west);
\draw[arrow] (k_decay.east) -- (k_reproj.west);
\draw[arrow] (k_reproj.east) -- (k_agg.west);
\draw[arrow] (k_agg.east) -- (k_output.west);

\node[label, anchor=north, text width=40ex] at (0.45, \levelBottomY-0.95) {
    \scriptsize Shared kernel parameters across all $(t,s)$ pairs; aggregate over sources (Eq.~\ref{eq:aggregation-over-sources})
};

\node[box, draw=gray, thick, dashed, fill=white, text width=30ex, align=left, anchor=north east] at (0.95, \levelBottomY-0.9) {
    \textbf{\footnotesize Equivariance Mechanisms:}\\
    \begin{itemize}[noitemsep,topsep=0pt, leftmargin=1.0em]
        \scriptsize
        \item Translation: relative positions only
        \item Rotation: invariant features + reprojection
        \item Parity: parity-preserving reprojection
        \item Discretization: area-weighted aggregation
        \item Units: rigorous nondimensionalization
    \end{itemize}
};

\end{tikzpicture}

%% file: figures/globe_io_mapping.tex
\begin{tikzpicture}[x=\textwidth]
    \tikzset{
        box/.style={
            draw, rounded corners,
            inner xsep=3pt, inner ysep=6pt, outer sep=3pt,
            minimum height=1cm
        },
        arrow/.style={->, >=latex, line width=1pt, shorten >=0pt, shorten <=0pt}
    }
    \node[box, 
    fill=Peach!15, text width=\dimexpr\textwidth\relax, anchor=south
    ] (input) at (0.5,1.2){
        {\centering \textbf{Problem Inputs}\par}
        \begin{itemize}[noitemsep, topsep=0pt, left=1.2em]
            \item A sequence of one or more boundary meshes (one per BC type), each of which consists of:
            \begin{itemize}[noitemsep, topsep=0pt, left=1.2em]
                \item Geometry: triangle faces in 3D or line edges in 2D; represented as vertices and face indices (e.g., an unstructured STL file)
                \item The boundary condition (BC) type as a string label (e.g., ``no\_slip'', ``inlet'', ``outlet'')
                \item Any unitless scalar or vector values associated with this BC type (e.g., values of a Dirichlet boundary condition)
            \end{itemize}
            \item A set of scalar, dimensional reference lengths $\{\ell_i\}$, obtained through combinations of dimensional parameters in the governing equations (via Buckingham $\pi$ theorem or domain expertise)
            \item (Optional) Global scalars and/or vectors (unitless), which could be drawn from the governing equations or infinite far-field boundary conditions
        \end{itemize}
    };

    \node[box,
    fill=Peach!15, text width=\dimexpr0.35\textwidth\relax, anchor=west
    ] (coords) at (0.0,0){
        {\centering \textbf{Query Points}\par}
        \begin{itemize}[noitemsep, topsep=0pt, left=1.2em]
            \item Spatial coordinates where the solution fields are evaluated
        \end{itemize}
    };

    \node[box, 
    fill=TealBlue!15, text width=\dimexpr0.15\textwidth\relax, anchor=center, align=flush center
    ] (solver) at (0.5,0){
        \textbf{GLOBE Model}\\(Figure \ref{fig:globe_arch})
        };

    \node[box, 
    fill=RoyalPurple!15, text width=\dimexpr0.35\textwidth\relax, 
    align=left, anchor=east
    ] (output) at (1.0,0.0){
        {\centering \textbf{Output}\par}
        \begin{itemize}[noitemsep, topsep=0pt, left=1.2em]
            \item Values of the solution fields (scalars and/or vectors, all unitless) at the query points
        \end{itemize}
    };

    \draw[arrow] (input.south) -- (solver.north);
    \draw[arrow] (coords.east) -- (solver.west);
    \draw[arrow] (solver.east) -- (output.west);
\end{tikzpicture}

%% file: checklist.tex
\section*{NeurIPS Paper Checklist}

\begin{enumerate}

\item {\bf Claims}
    \item[] Question: Do the main claims made in the abstract and introduction accurately reflect the paper's contributions and scope?
    \item[] Answer: \answerYes{}
    \item[] Justification: The abstract and \cref{sec:necessary-attributes,sec:model} state the contributions; the headline accuracy claims are quantified in \cref{tab:airfrans-full,tab:airfrans-scarce,tab:drivaer}, and the architectural-attribute claims are formally proved in \cref{sec:mathematical-proofs}.

\item {\bf Limitations}
    \item[] Question: Does the paper discuss the limitations of the work performed by the authors?
    \item[] Answer: \answerYes{}
    \item[] Justification: A dedicated \emph{Limitations} paragraph in \cref{sec:conclusions} discusses scope of evaluation (2D AirFRANS, plus a 3D DrivAerML validation study in which \globe{} trails a matched Transolver baseline), single-run point estimates, the absence of integrated force and moment metrics on AirFRANS, far-field decay envelope assumptions, $\mathcal{O}(\nsrc^2 + \nsrc \ntrg)$ complexity scaling without hierarchical acceleration, and the expected weakness of the boundary-element inductive bias in strongly nonlinear regimes.

\item {\bf Theory assumptions and proofs}
    \item[] Question: For each theoretical result, does the paper provide the full set of assumptions and a complete (and correct) proof?
    \item[] Answer: \answerYes{}
    \item[] Justification: \cref{sec:mathematical-proofs} provides formal proofs of translation-, rotation-, and parity-equivariance, plus discretization- and units-invariance, with assumptions stated alongside each proof; supporting derivations are in \cref{sec:kernel-appendix}.

    \item {\bf Experimental result reproducibility}
    \item[] Question: Does the paper fully disclose all the information needed to reproduce the main experimental results of the paper to the extent that it affects the main claims and/or conclusions of the paper (regardless of whether the code and data are provided or not)?
    \item[] Answer: \answerYes{}
    \item[] Justification: \cref{sec:preprocessing-details,sec:hyperparameters} (\cref{tab:hyperparameters}) document the AirFRANS preprocessing, hyperparameters, and training protocol, and \cref{sec:drivaer-appendix} documents the DrivAerML models, baseline, and evaluation; both datasets \citep{bonnetAirfRANSHighFidelity2023,ashtonDrivAerMLHighFidelityComputational2024} are public.

\item {\bf Open access to data and code}
    \item[] Question: Does the paper provide open access to the data and code, with sufficient instructions to faithfully reproduce the main experimental results, as described in supplemental material?
    \item[] Answer: \answerYes{}
    \item[] Justification: The AirFRANS dataset \citep{bonnetAirfRANSHighFidelity2023} is publicly available; the GLOBE reference implementation is released open-source as part of NVIDIA PhysicsNeMo (\url{https://github.com/NVIDIA/physicsnemo}, under \texttt{examples/cfd/external\_aerodynamics/globe/}), along with environment specifications and example training scripts.

\item {\bf Experimental setting/details}
    \item[] Question: Does the paper specify all the training and test details (e.g., data splits, hyperparameters, how they were chosen, type of optimizer) necessary to understand the results?
    \item[] Answer: \answerYes{}
    \item[] Justification: AirFRANS splits (\cref{sec:task-airfrans}), hyperparameters (\cref{tab:hyperparameters}), and training details (Muon + RAdamW optimizer, Huber loss, float32 training; \cref{sec:hyperparameters}) are specified alongside the DrivAerML splits, models, and training protocol (\cref{sec:task-drivaer,sec:drivaer-appendix}).

\item {\bf Experiment statistical significance}
    \item[] Question: Does the paper report error bars suitably and correctly defined or other appropriate information about the statistical significance of the experiments?
    \item[] Answer: \answerNo{}
    \item[] Justification: \cref{tab:airfrans-full,tab:airfrans-scarce,tab:airfrans-reynolds,tab:airfrans-aoa} report single-run point estimates per regime, matching prior AirFRANS literature (\cref{tab:drivaer} adds three-seed ensembles on DrivAerML). Multi-seed error bars across all four AirFRANS regimes exceeded our compute budget, so we label AirFRANS results as single-run point estimates and draw no conclusions from small differences.

\item {\bf Experiments compute resources}
    \item[] Question: For each experiment, does the paper provide sufficient information on the computer resources (type of compute workers, memory, time of execution) needed to reproduce the experiments?
    \item[] Answer: \answerYes{}
    \item[] Justification: \cref{sec:results-cost} reports, for the pre-release implementation, training throughput on 40--128 NVIDIA GB200 GPUs (0.341 GPU-seconds per training sample), per-sample inference time (2.326\,s on a single GB200), and chunked-inference memory ($\sim$700\,MB at chunk size 128), plus epoch and inference times of the public implementation on GB300 GPUs; \cref{sec:drivaer-appendix} reports the DrivAerML training hardware and inference times on GB300 GPUs. The reported Full and Scarce models were trained on 32--80 NVIDIA H100 GPUs (\cref{sec:hyperparameters}).
    
\item {\bf Code of ethics}
    \item[] Question: Does the research conducted in the paper conform, in every respect, with the NeurIPS Code of Ethics \url{https://neurips.cc/public/EthicsGuidelines}?
    \item[] Answer: \answerYes{}
    \item[] Justification: The work develops PDE surrogate models for engineering simulation using only public synthetic CFD data; it does not involve human subjects or sensitive personal data. Like other aerodynamic analysis tools, the surrogate is dual-use, as noted in the \emph{Broader Impacts} section; our intended use is open civil and scientific applications.

\item {\bf Broader impacts}
    \item[] Question: Does the paper discuss both potential positive societal impacts and negative societal impacts of the work performed?
    \item[] Answer: \answerYes{}
    \item[] Justification: The \emph{Broader Impacts} section discusses positive impacts (once validated on integrated design quantities, faster engineering design cycles in aerospace, automotive, and energy systems and reduced computational energy associated with traditional CFD) and notes that aerodynamic surrogates are inherently dual-use; our intended use is open civil and scientific applications.
    
\item {\bf Safeguards}
    \item[] Question: Does the paper describe safeguards that have been put in place for responsible release of data or models that have a high risk for misuse (e.g., pre-trained language models, image generators, or scraped datasets)?
    \item[] Answer: \answerNA{}
    \item[] Justification: The released artifact is a numerical surrogate for aerodynamic flowfields trained on public synthetic CFD data; it is neither a generative model with content-misuse risk nor a scraped dataset, so dedicated safeguards are not applicable.

\item {\bf Licenses for existing assets}
    \item[] Question: Are the creators or original owners of assets (e.g., code, data, models), used in the paper, properly credited and are the license and terms of use explicitly mentioned and properly respected?
    \item[] Answer: \answerYes{}
    \item[] Justification: The AirFRANS \citep{bonnetAirfRANSHighFidelity2023} and DrivAerML \citep{ashtonDrivAerMLHighFidelityComputational2024} datasets are cited at point of use; each is used in compliance with its license.

\item {\bf New assets}
    \item[] Question: Are new assets introduced in the paper well documented and is the documentation provided alongside the assets?
    \item[] Answer: \answerYes{}
    \item[] Justification: The new asset introduced is the GLOBE reference implementation; its architecture is documented in \cref{sec:model}, hyperparameters in \cref{tab:hyperparameters}, training protocol in \cref{sec:hyperparameters}, and preprocessing equations in \cref{sec:preprocessing-details}. The code is released open-source in NVIDIA PhysicsNeMo (\url{https://github.com/NVIDIA/physicsnemo}, under \texttt{examples/cfd/external\_aerodynamics/globe/}), with a README and example scripts.

\item {\bf Crowdsourcing and research with human subjects}
    \item[] Question: For crowdsourcing experiments and research with human subjects, does the paper include the full text of instructions given to participants and screenshots, if applicable, as well as details about compensation (if any)? 
    \item[] Answer: \answerNA{}
    \item[] Justification: The work uses only public synthetic CFD data; no crowdsourcing or human-subjects research was conducted.

\item {\bf Institutional review board (IRB) approvals or equivalent for research with human subjects}
    \item[] Question: Does the paper describe potential risks incurred by study participants, whether such risks were disclosed to the subjects, and whether Institutional Review Board (IRB) approvals (or an equivalent approval/review based on the requirements of your country or institution) were obtained?
    \item[] Answer: \answerNA{}
    \item[] Justification: As noted above, no human subjects were involved in this work, so IRB review was not applicable.

\item {\bf Declaration of LLM usage}
    \item[] Question: Does the paper describe the usage of LLMs if it is an important, original, or non-standard component of the core methods in this research? Note that if the LLM is used only for writing, editing, or formatting purposes and does \emph{not} impact the core methodology, scientific rigor, or originality of the research, declaration is not required.
    \item[] Answer: \answerNA{}
    \item[] Justification: LLMs are not a component of the core methods, models, or analyses presented in this paper; the GLOBE architecture is a kernel-based neural surrogate for PDEs.

\end{enumerate}